\documentclass{article}

\usepackage[utf8]{inputenc}
\usepackage{newunicodechar}
\newunicodechar{−}{\textminus}

 \usepackage[preprint]{neurips_2026}

\usepackage[utf8]{inputenc} 
\usepackage[T1]{fontenc}    
\usepackage{hyperref}       
\usepackage{url}            
\usepackage{booktabs}       
\usepackage{amsfonts}       
\usepackage{nicefrac}       
\usepackage{microtype}      
\usepackage{xcolor}         

\PassOptionsToPackage{numbers}{natbib}
\usepackage{graphicx,subcaption}
\usepackage{enumitem}
\usepackage{algorithm}
\usepackage{algorithmic}
\usepackage{booktabs}  
\usepackage{ragged2e}
\usepackage{wrapfig}
\usepackage{arydshln}
\usepackage{setspace}
\usepackage{adjustbox} 
\usepackage{framed}
\usepackage{tikz}
\usepackage{bbm}
\usepackage{tabularx}
\usepackage{amssymb}
\usepackage{mathtools}
\usepackage{amsthm}
\usepackage{xcolor}
\usetikzlibrary{matrix,calc}
\usepackage{multirow}
\usepackage{arydshln}
\usepackage{subcaption}
\usepackage{rotating} 
\usepackage{booktabs}
\usepackage{makecell}
\usepackage{xcolor}
\usepackage{graphicx}

\usepackage{pdflscape}
\usepackage{longtable}
\usepackage{array}
\usepackage{booktabs}
\usepackage{ragged2e}
\newcolumntype{P}[1]{>{\RaggedRight\arraybackslash}p{#1}}

\usepackage[table]{xcolor}
\usepackage{soul}
\usetikzlibrary{tikzmark,calc,arrows.meta}
\usetikzlibrary{positioning,calc,fit,backgrounds}
\definecolor{Ucolor}{RGB}{232,242,255}  
\definecolor{Rcolor}{RGB}{210,230,215}
\definecolor{Scolor}{RGB}{255,238,232}
\definecolor{pidHLblue}{RGB}{198,220,255}
\definecolor{pidHLgreen}{RGB}{206,242,206}
\definecolor{pidHLpink}{RGB}{255,210,225}
\definecolor{pidHLorange}{RGB}{255,225,190}

\DeclareRobustCommand{\fillcircle}{\tikz[baseline=-0.5ex]\draw[fill=black] (0,0) circle (0.55ex);}
\DeclareRobustCommand{\opencircle}{\tikz[baseline=-0.5ex]\draw (0,0) circle (0.55ex);}

\theoremstyle{remark}

\usepackage[most]{tcolorbox}
\usepackage{calc}

\definecolor{titlebg}{HTML}{E6E5EF}    
\definecolor{linkblue}{HTML}{116E8A}

\newcommand{\mitlogoheight}{0.85cm}
\newcommand{\kuleuvenlogoheight}{1.33cm}
\newcommand{\logorowheight}{1.33cm}

\makeatletter
\newcommand{\fairauthors}[1]{\gdef\@fairauthors{#1}}
\newcommand{\fairaffiliations}[1]{\gdef\@fairaffiliations{#1}}
\newcommand{\fairabstract}[1]{\gdef\@fairabstract{#1}}
\newcommand{\faircorrespondence}[1]{\gdef\@faircorrespondence{#1}}
\gdef\@fairauthors{}\gdef\@fairaffiliations{}
\gdef\@fairabstract{}\gdef\@faircorrespondence{}

\newcommand{\synibmaketitle}{%
  \thispagestyle{empty}%
  \begingroup
  \tcbset{enhanced,frame hidden,
          left=0.7cm,right=0.7cm,top=0.32cm,bottom=0.28cm,
          arc=5pt,colback=titlebg,
          before skip=0pt,after skip=0.3cm,
          grow to left by=1.5pt,grow to right by=1.5pt}
  \begin{tcolorbox}
    \setlength{\parindent}{0pt}%
    \raggedright
    {\Large\bfseries \@title\par}
    \vskip 0.16cm
    {\small\bfseries \@fairauthors\par}
    \vskip 0.06cm
    {\small \@fairaffiliations\par}
    \vskip 0.16cm
    {\normalsize \@fairabstract\par}
    \vskip 0.16cm
    \noindent
    \begin{minipage}[c]{0.48\linewidth}
      \raggedright
      \ifx\@faircorrespondence\@empty\else
        {\small\textbf{Correspondence:} \@faircorrespondence}%
      \fi
    \end{minipage}%
    \hfill
    \begin{minipage}[c]{0.50\linewidth}
      \raggedleft
      \mbox{%
        \parbox[c][\logorowheight][c]{\widthof{\includegraphics[height=\kuleuvenlogoheight]{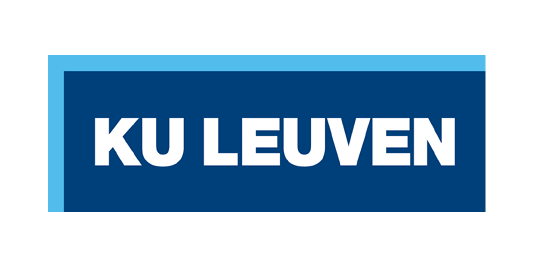}}}{%
          \centering\includegraphics[height=\kuleuvenlogoheight]{Figures/logos/KU-Leuven-logo.png}%
        }%
        \hspace{0.4cm}%
        \parbox[c][\logorowheight][c]{\widthof{\includegraphics[height=\mitlogoheight]{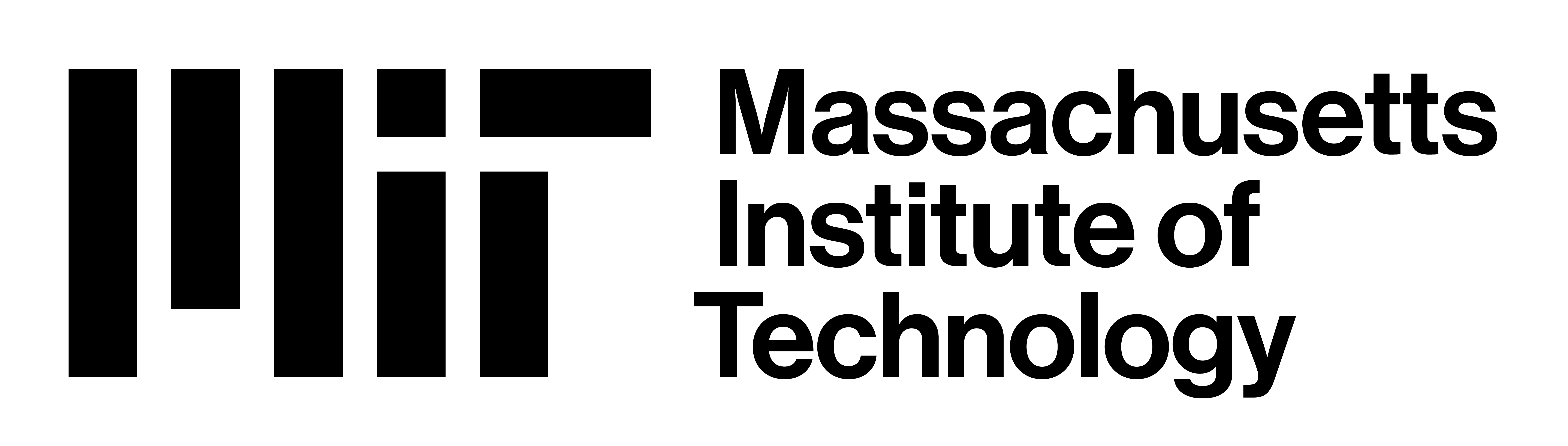}}}{%
          \centering\includegraphics[height=\mitlogoheight]{Figures/logos/mit_lockup_std-three-line_rgb_black.png}%
        }%
      }%
    \end{minipage}%
  \end{tcolorbox}
  \endgroup
}
\makeatother

\title{NeuroAtlas: Benchmarking Foundation Models for Clinical EEG and Brain-Computer Interfaces}

\author{}

\fairauthors{%
  Konstantinos Kontras$^{1,2,*}$,
  Trui Osselaer$^{1,*}$,
  Stylianos G. Mouslech$^{1,*}$,
  Angeliki-Ilektra Karaiskou$^{1,*}$,
  Guido Gagliardi$^{1,*}$,
  Thomas Strypsteen$^{1,*}$,
  Mohammad Hossein Badiei$^{1}$,
  Anku Rani$^{2}$,
  Maarten Vanmarcke$^{1}$,
  Miguel Bhagubai$^{1}$,
  Chanakya Ekbote$^{2}$,
  Jaedong Hwang$^{2}$,
  Christos Chatzichristos$^{1}$,
  Paul Pu Liang$^{2}$,
  Maarten De Vos$^{1}$%
}
\fairaffiliations{%
  $^{1}$KU Leuven,\quad
  $^{2}$MIT, \quad
  $^{*}$Equal contribution%
}
\faircorrespondence{\href{mailto:konstantinos.kontras@kuleuven.be}{\texttt{konstantinos.kontras@kuleuven.be}}}

\fairabstract{%
Foundation models (FMs) promise to extract unified representations that generalize across downstream tasks. They have emerged across fields including electroencephalography (EEG), but it is less clear how effective they are in this particular field. Published evaluations differ in datasets, in the EEG-specific preprocessing that might influence reported results, and in the reported metrics frequently obscuring the clinical relevance in EEG. We introduce NeuroAtlas, the largest EEG benchmark to date: 42 datasets and $\sim$260k hours covering clinical EEG (epilepsy, sleep medicine, brain age estimation) and brain-computer interfaces, and include multiple datasets per task along with bespoke clinical evaluation metrics. Besides evaluating EEG-FMs with respect to supervised baselines, we present results from generic time-series FMs. We report three findings. First, EEG-specific FMs do not consistently outperform time-series FMs which have neither EEG-focused architectures nor been pretrained on EEG. Second, standard machine learning metrics are insufficient to assess clinical utility: Thus, we thoroughly evaluate more appropriate measures such as the quality of event-level decision-making, hypnogram-derived features, and the brain-age gap in the domains of epilepsy, sleep and brain age respectively. Third, model rankings and performance can vary substantially within domains. We conclude that pretrained models perform largely on par, with only narrow advantages for a few, and that current FMs do not yet deliver on the promise of an out-of-the-box unified EEG model. NeuroAtlas exposes this gap and provides the datasets and metrics for the next generation of unified EEG-FMs.%
}

\begin{document}

\synibmaketitle

\section{Introduction}
\label{sec:intro}
Electroencephalography (EEG) provides a non-invasive and affordable assessment of brain function, used across a variety of clinical domains including epilepsy, sleep medicine, intensive care monitoring, and for steering brain-computer interfaces (BCIs). Each application is served today by a separate pipeline, typically tuned to a specific cohort, site, and device. Foundation models (FMs) aim to tackle this fragmentation. Recently, several EEG-FMs have been proposed with the promise to derive cross-domain generalizable embeddings from a single backbone by training on massive and diverse EEG datasets. In other fields there have been notable successes in off-the-shelf performance ~\citep{radford2021learning, caron2021emerging}, raising the question: How close are current EEG-FMs to fulfilling their potential?

Existing evaluations give an incomplete answer. Most benchmarks reduce each domain to a single dataset, a single metric, and a single family of models~\citep{xiong_eeg-fm-bench_2026, kastrati_eeg-bench_2025}, leaving three central aspects of EEG-FM behavior untested. \emph{First, intra-domain generalization.} Performance varies substantially across datasets within a single domain due to differences in acquisition device, electrode layout, referencing scheme, artifact profile, and population phenotype, and a credible benchmark should include multiple datasets per domain that span this variation. \emph{Second, metrics that reflect clinical relevance.} Aggregate metrics such as AUROC or accuracy can hide failure modes that determine real-world usability, so each domain demands bespoke, clinically grounded evaluation. \emph{Third, comparison against model alternatives.} Existing benchmarks typically compare EEG-FMs only against each other, leaving the value of EEG-specific pretraining unverifiable against time-series FMs (TS-FMs) and models pretrained with supervision on other EEG datasets (referred to throughout as \textit{Supervised-Pre} models). We introduce NeuroAtlas, advancing EEG-FM evaluation on three fronts:
\begin{figure}
    \centering
    \includegraphics[width=1\linewidth]{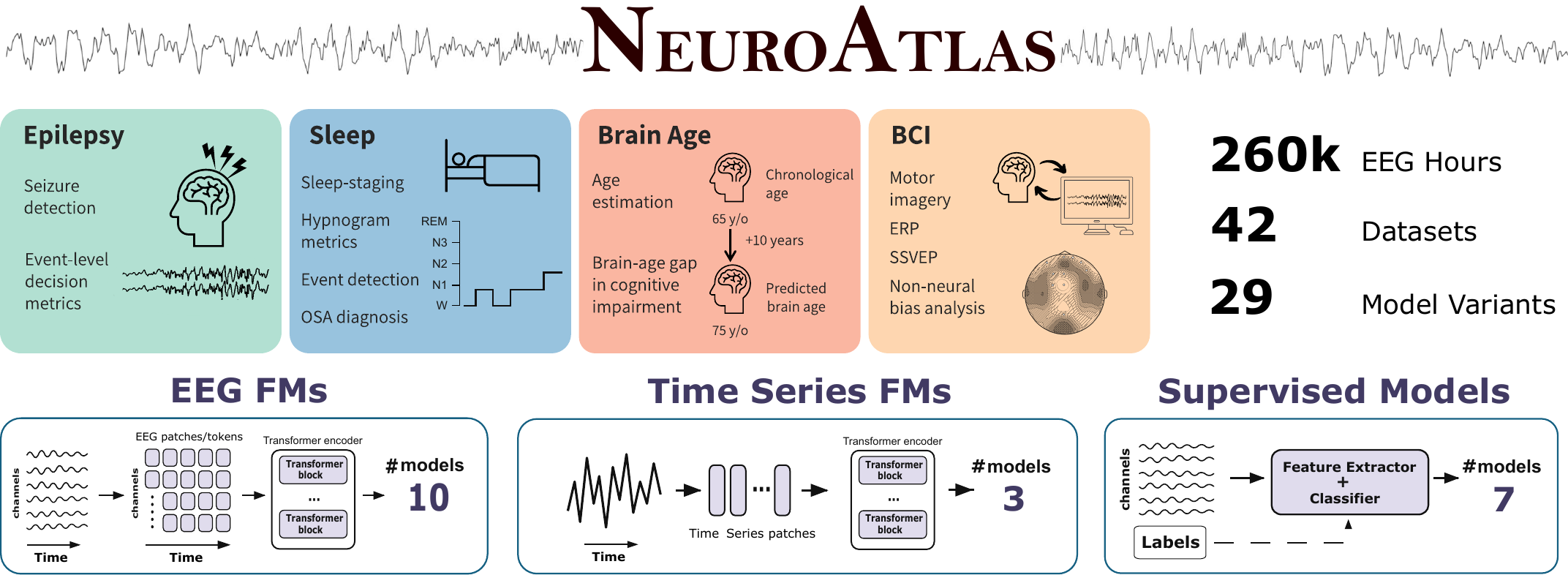}
    \caption{\textbf{NeuroAtlas overview.} NeuroAtlas spans four EEG domains (epilepsy, sleep, brain age, BCI) 42 datasets and ${\sim}260$k hours of EEG, and benchmarks three model families on each. Per-domain task list and dataset coverage, with total EEG hours per domain (epilepsy: $58$k h; sleep: $201$k h, shared with brain age; BCI: $170$ h). Each domain is evaluated with both standard and clinically grounded tasks: seizure detection with event-level decision metrics (epilepsy); sleep staging, hypnogram-derived features, event detection, and OSA diagnosis (sleep); chronological age estimation and the brain-age gap as a marker of cognitive impairment (brain age); and motor imagery, ERP, SSVEP, and non-neural confound analysis (BCI). Three model families are compared under a unified frozen-encoder probe: 10 EEG foundation models, 3 time-series foundation models, and 7 task-specific supervised baselines.}
\label{fig:neuroatlas_overview}
\end{figure}

\begin{enumerate}[leftmargin=*,itemsep=2pt,topsep=2pt]

    \item \textbf{Comprehensive data coverage.} NeuroAtlas comprises 42 datasets and ${\sim}260$k hours of EEG across seizure detection, sleep staging, BCI, and brain-age prediction (a new benchmark domain), with multiple datasets per domain spanning varied acquisition protocols, populations, and sites to expose intra-domain variability.

    \item \textbf{Comprehensive model coverage.} We evaluate 20 distinct models (29 size variants in total) spanning recent EEG foundation models, general-purpose time-series foundation models, and specialized supervised-pretrained baselines.

    \item \textbf{Neurophysiologically and clinically grounded evaluation.} Each domain is evaluated both with standard metrics and bespoke domain-specific measures that reflect real-world clinical utility: event-level sensitivity at fixed false-alarm rates for seizure detection, hypnogram-derived macrostructure for sleep, per-subject reliability for brain age, and adjustment for non-neural confounds in BCIs.
\end{enumerate}

Three findings emerge consistently across NeuroAtlas. (i) Model rankings and absolute performance shift substantially across datasets within the same domain, making single-dataset comparisons unreliable predictors of broader behavior. (ii) EEG-FMs do not consistently outperform generic TS-FMs, raising the question of how much of the reported gain stems from EEG-specific architectural choices or pretraining on neurophysiological data. (iii) Aggregate metrics translate only partially into clinical usability, with gaps revealed by hypnogram structure, event-level decision quality, and individual-level reliability. We release NeuroAtlas in full, including dataset access scripts, preprocessing pipelines, model configurations, and evaluation code, to support reproducible progress.

\section{Related Work}\label{sec:related}

\subsection{EEG Benchmarks and Evaluation Protocols}

A number of recent benchmarks have delivered tremendous efforts to evaluate EEG-FMs across multiple domains: AdaBrain-Bench \citep{wu2025adabrain}, (13 datasets, 4 EEG-FMs) , EEG-Bench \citep{kastrati_eeg-bench_2025} (14 datasets, 3 EEG-FMs), EEG-FM-Benchmark \citep{liu2026eeg} (13 datasets, 12 EEG-FMs), EEG-FM-Bench \citep{xiong_eeg-fm-bench_2026} (14
datasets, 7 EEG-FMs, 2 general-purpose time-series FMs), ST-EEGFormer \citep{yang2026eeg} (10 datasets, 5 EEG-FMs) and Brain4FMs
\citep{shen_brain4fms_2026} (18 datasets, 15 brain
FMs). A recurring limitation among many of these benchmarks is that, while they aim for a wide array of tasks, the performance within distinct clinical domains is typically represented by a single cohort or dataset. 
In addition, as observed by Kuruppu et al.~\citep{kuruppu2026eeg}, the evaluation protocols employed by the EEG-FM papers themselves make direct comparison impossible. With the exception of TUAB and TUEV, there is a significant heterogeneity in the selected downstream tasks across these works. Plus, only a limited number of these works extensively investigated generalization through out-of-distribution evaluation without fine-tuning, using, e.g., linear probing or few-shot evaluation. Furthermore, the EEG-FM benchmarks typically report a fixed set of evaluation metrics (e.g., balanced accuracy and AUROC for classification tasks), disregarding the clinical meaning of different metrics across domains.

\subsection{Foundation Models}

EEG-FMs such as BIOT \citep{yang2023biot}, LaBraM
\citep{jiang2024large}, EEGPT \citep{wang2024eegpt}, CBraMod
\citep{wang2024cbramod}, NeuroLM \citep{jiang2024neurolm}, and REVE
\citep{ouahidi2025reve} aim to learn generalizable representations or embeddings from large-scale, heterogeneous, unlabeled EEG data. They are typically pretrained through self-supervised learning, with the pretext tasks mainly consisting of masked signal reconstruction in the temporal and/or spectral domain, or contrastive learning. On the other hand, general-purpose
TS-FMs such as Chronos~\citep{ansari2024chronos}, MOMENT~\citep{goswami2024moment}, and Moirai~\citep{woo2024unified} are pretrained on diverse temporal
corpora and target forecasting and related tasks. They use no EEG-specific architecture or pretraining data, but recent evidence shows EEG-FMs might be competitive on specific clinical tasks: TimesNet~\citep{wu2022timesnet} on sleep staging in the FoME~\citep{shi2024fome} evaluation, MOMENT on
seizure detection in the BrainWave~\citep{yuan2024brainwave} evaluation, and PatchTST~\citep{nie2022time} on
forecasting and imputation in the Brant evaluation \citep{kuruppu2026eeg}. This raises the question of how much of
the reported gain on EEG-FM benchmarks comes from EEG-specific
design rather than from generic large-scale temporal pretraining, a
question NeuroAtlas is designed to answer.


\section{Benchmark Pipeline}
\label{sec:benchmark}

NeuroAtlas asks whether the latent representations learned by pretrained models, regardless of the pretraining strategy, encode the information required for clinically meaningful tasks across diverse EEG domains, in a sufficiently disentangled form to be recovered by a simple downstream readout. To answer this directly, we evaluate every model under a unified protocol: The pretrained backbone is treated as a fixed feature extractor, and a linear probe is fit per fold of each downstream task, with classifier and evaluation held identical across model families. This isolates the quality of the pretrained representation itself from confounds such as fine-tuning, architecture-specific tuning, or task-specific heads.

\paragraph{Models and datasets} \label{subsec:model} NeuroAtlas evaluates three groups of models: EEG-FMs, TS-FMs and Supervised-Pre EEG models. This setup enables comparison between  EEG self-supervised pretraining, generic temporal representation learning, and task-specific supervised training. A full description of the models is found in Appendix~\ref{Appendix:Model-Description--Included-Models}. On top of these pretrained models, we evaluate CBraMod~\cite{wang2024cbramod} with randomly initialized weights as a unified random-init reference, isolating the contribution of pretraining from architecture and probe. Furthermore, the benchmark spans 42 datasets distributed over the domains of epileptic seizure detection, sleep medicine, brain age prediction and BCI, an overview of which can be found in Appendix~\ref{Appendix:Dataset-Description}.

\paragraph{Preprocessing}\label{subsec:preprocessing} Establishing a fully harmonized preprocessing pipeline in EEG is a challenging task. Not only have the different application domains typically operated under different electrode configurations, channel referencing, sampling rates and frequency filtering, but even the models themselves have been pretrained under different specific preprocessing conditions. In NeuroAtlas, we apply minimal shared preprocessing by only applying a highpass filter at 0.5Hz, a notch filter at the relevant power line noise to mitigate power line interference and windowing to a window size determined by the downstream task. The remainder of the preprocessing pipeline is model-specific: resampling, referencing, channel matching and scaling are all determined by the pretraining conditions specified by each model. This way, we ensure the downstream data matches the in-distribution data of each model as closely as possible. Dataset-model channel matching information and further details on preprocessing can be found in Appendix~\ref{Appendix:Model-Description--Included-Models}.

\paragraph{Evaluation}\label{subsec: evaluation} We evaluate EEG-FMs, TS-FMs, and Supervised-Pre models based on their ability to encode generalizable, task-relevant features both within and across domains. To this end, we consistently apply a linear probing approach on the extracted, frozen embeddings across all the domains, without any fine-tuning of the encoding backbones (neither FMs nor Supervised-Pre models). 
In the rare cases that a downstream dataset overlaps with a dataset used for pretraining, this is clearly indicated in the associated results section. Furthermore, every evaluation is carried out in a purely cross-subject manner, without any fine-tuning on specific subject characteristics. The measured performance metrics are unique to each domain to ensure their clinical relevance to the specific applications. Finally, we investigate the performance gap between domain-specific probes trained in a leave-one-dataset (LODO) paradigm (using no training data from the held-out downstream dataset under evaluation) and dataset-specific probes in order to assess the zero-shot potential of these models.

\section{Experiments}\label{sec:exp}

\subsection{Epilepsy}

Epileptic seizure detection is among the most frequently reported clinical tasks in the EEG-FM literature, yet evaluations share two limitations that obscure what is being measured. (i) Typically, the task is performed by segmenting each recording into non-overlapping 10-second windows \cite{jiang2024neurolm}; the model produces a binary classification per window (seizure vs. non-seizure), and performance is summarized by aggregating across windows. Window-level AUROC measured in cases of extreme class imbalance can remain high even while the model is generating a significant amount of false alarms, which is unacceptable under the strict false-alarm constraints of clinical seizure detection ~\citep{dan2024szcore}. (ii) The value of pretraining is realized when learned representations transfer to new tasks with minimal adaptation; current EEG-FM evaluations, however, do not isolate the contribution of pretraining from that of the downstream task-specific training, leaving the quality of the representation itself difficult to assess. NeuroAtlas addresses both by evaluating across seven datasets ($\sim$1{,}410 patients, $\sim$58,374 hours, $\sim$9{,}019 annotated seizures).

\paragraph{Evaluation.}
We report event-level sensitivity over $0.1$--$100$\,FA/h as the area under the mean (5-fold) sensitivity vs.\ $\log_{10}(\text{FA/h})$ curve~\citep{bhagubai2024towards} and for completeness the AUROC and balanced accuracy in App.~\ref{app:auroc_metrics}. For every model, we perform the same steps in terms of preprocessing, channel selection, and window cropping as their original pretraining or their indicated downstream task usage.


\begin{figure}
    \centering
    \includegraphics[width=\linewidth]{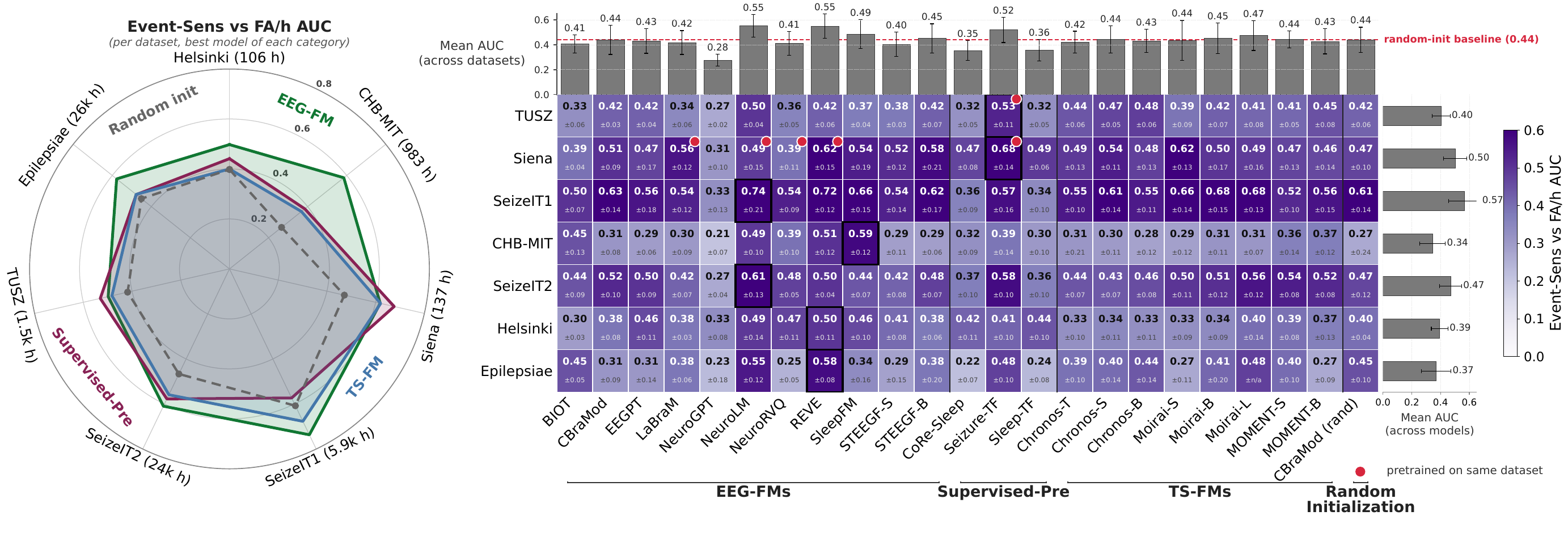}
\caption{\textbf{Epilepsy: no model or family is consistently best across datasets, and most models perform on par with random initialization.} \textbf{(a)} Family-best Event-Sens@FA AUC across seven continuously-labeled datasets (labels include total EEG hours), as the area under the mean (5-fold) sensitivity vs. $\log_{10}$(FA/h) curve over $0.1$--$100$ FA/h under any-overlap event scoring~\cite{dan2024szcore}. The leading EEG-FM differs across datasets and no family consistently dominates. \textbf{(b)} Per-(model,~dataset) AUC for 23 models: EEG-FMs, Supervised-Pre, TS-FMs, CBraMod random-init. Cells show mean AUC (bold) and 5-fold std; black borders mark the dataset-best model; red dots flag pretraining/evaluation overlap. Grey bars: per-model means across datasets (top, red dashed: random-init baseline at $0.44$) and per-dataset means across models (right). Most per-model means lie within ${\approx}0.05$ AUC of the $0.44$ baseline; only NeuroLM and REVE exceed it at approx. 1 std (both $0.55$), and the dataset-best alternates between NeuroLM, REVE and SleepFM. }
\label{fig:epilepsy}
\end{figure}

\paragraph{Results.}
Event-level performance varies strongly across datasets (Figure~\ref{fig:epilepsy}): the dataset-best Event-Sens@FA AUC ranges from ${\approx}0.50$ on Helsinki to ${\approx}0.74$ on SeizeIT1, the leading model alternates between NeuroLM, REVE, and SleepFM, and within-dataset spreads of ${\approx}0.2$--$0.4$ AUC make these rank changes consequential. All three pretrained families exceed the CBraMod random-init reference on every dataset, but per-model the margin lies within one fold-standard-deviation of random init on the majority of datasets. A one-sided binomial top-3 win test (Appendix~\ref{sec:epilepsy-stat-testing}, Table~\ref{tab:top3_win_test}) identifies only NeuroLM ($6/7$ datasets, $p < 10^{-3}$) and REVE ($5/7$ overall, $6/7$ within EEG-FMs; both $p \leq 0.002$) as exceeding chance; the remaining EEG-FMs do not, even within their own family. Pretraining therefore captures something beyond architecture and probe, but the gain is concentrated in two models and is small enough (\~1 std) that similar performance on a new dataset is not yet warranted. Three further analyses refine these conclusions. First, metric choice matters: AUROC and Event-Sens@FA produce different rankings on every dataset (Spearman $\rho \in [0.61, 0.81]$; Appendix~\ref{app:metric-rank-disagreement}, Figure~\ref{fig:metric-rank-disagreement}), most visibly on TUSZ where NeuroLM is rank-1 by AUROC but is displaced by Seizure-TF on Event-Sens@FA. Second, ILAE-2017-type-stratified~\cite{avakyan2017ilae} probability distributions (App.~\ref{sec:seizure-type-violins}, Figure~\ref{fig:seizure-type-violins}) expose dataset-dependent failure modes: The best EEG-FM concentrates near $1$ on focal, generalized, and focal-to-bilateral events on TUSZ and SeizeIT2 but shows long tails on Epilepsiae focal events. Third, on three recording-level datasets (Bonn, TUAB, NMT; App.~\ref{appendix:non_window}- Figure~\ref{fig:non_window_dumbbell}), REVE leads TUAB and NMT, and TS-FMs reach near-parity with EEG-FMs on NMT. Fourth, when the probe is fitted without any labels from the target dataset, every model loses a quarter to nearly half of its event-level performance, and none reaches the within-dataset random-init reference (App. \ref{app:epilepsy_lodo}, Table \ref{tab:lodo-epilepsy}).

\subsection{Sleep-staging}

Current EEG-FM sleep-staging benchmarks remain limited in both scale and clinical relevance. First, evaluations are typically conducted on small amounts of data in terms of both hours and number of patients \cite{kastrati_eeg-bench_2025, shen_brain4fms_2026, xiong_eeg-fm-bench_2026, yao_foundation_2025}. The previous most extensive benchmark included 558 patients from 3 datasets totaling 7.5k hours \cite{wu2025adabrain}, making it difficult to assess generalized performance and to disentangle transferable representations from dataset-specific effects. Second, in clinical practice, sleep is interpreted through the hypnogram, the ordered sequence of stages (wake, REM, N1–N3) staged according to the American Academy of Sleep Medicine (AASM) criteria, from which clinical macrostructural features like REM latency and total sleep time (TST) are derived \cite{aasm2023}. Furthermore, diagnosis is supported by event annotations such as microarousals, respiratory events, and limb movements. Beyond primary sleep disorders, such markers have also been linked to cognitive impairment and neurodegenerative diseases \cite{christensen_sleep_2016, li_evaluating_2022}. However, EEG-FMs' performance is predominantly reported as epoch-wise agreement with expert labels, which does not directly assess agreement on the sequential organization underlying macrostructural hypnogram features, used in clinical practice \cite{jiang2024neurolm, ouahidi2025reve}. Overall, this highlights the need for more comprehensive evaluation in terms of downstream cohorts and evaluated metrics.
 
\paragraph{Evaluation protocol} We evaluate sleep representations across 15 datasets, representing  $\sim$15.8k patients and  $\sim$201k EEG hours, by splitting recordings in 30-second epochs, and extracting embeddings per epoch. Evaluation is conducted along four complementary axes. First, epoch-wise sleep staging is evaluated by window-level Cohen's $\kappa$. Second, we assess whether predicted hypnograms preserve macrostructural hypnogram features used in downstream clinical tasks (for a full overview see App. \ref{Appendix:Experiments-Sleep}). Third, to assess representation beyond the primary stages labels, we evaluate epoch-level detection of microarousals, respiratory events and periodic limb movements. Finally, we assess recording-level diagnosis of obstructive sleep apnea (OSA), a primary sleep disorder, and cognitive impairment from sleep EEG.

\begin{figure}[t]
    \centering
    \includegraphics[width=\textwidth]{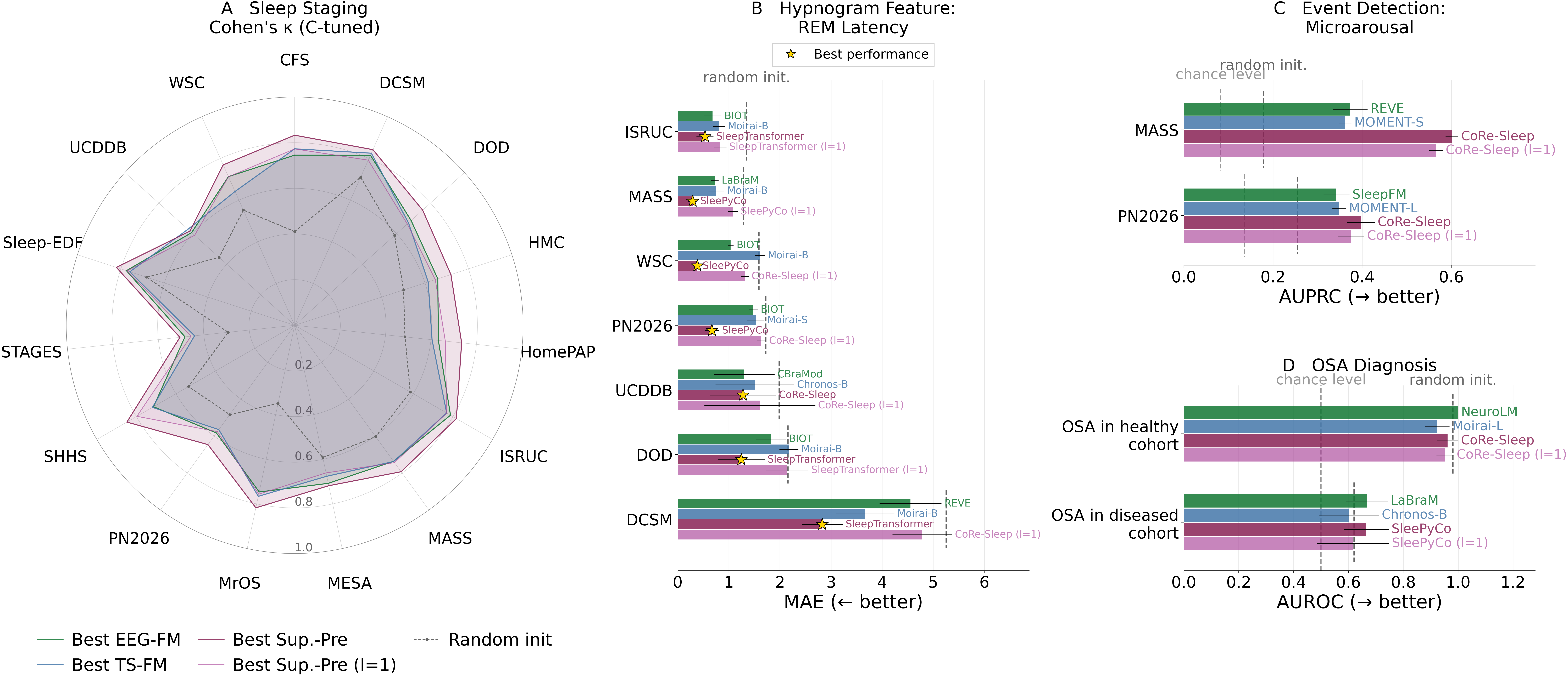}
    \caption{\textbf{Sleep: Performance is strongly dataset-dependent across tasks.} Metrics are shown for the best-performing model per category, i.e., EEG-FMs, TS-FMs, and Supervised-Pre models with and without temporal context (sequence length l=1), and a random-initialized reference. \textbf{(a)} Cohen's $\kappa$ sleep staging agreement with ground-truth labels. Model categories achieve similar performance when excluding the effects of sequence modeling. \textbf{(b)} MAE between ground truth and predicted REM latency. Performance on sleep staging agreement does not always correspond to performance on hypnogram features. \textbf{(c)} AUPRC for microarousal detection, for events with a minimal duration of 3 seconds. Chance baseline is indicated by prevalence. While all model categories perform above baseline, FMs do not reach Supervised-Pre performance. \textbf{(d)} AUROC for diagnosis of OSA compared to healthy patients, and in a cohort with other disorders. Chance baseline is indicated at 0.5. While OSA is detected in a healthy cohort, performance degrades in the presence of other disorders.}
    \label{fig:Sleep}
\end{figure}

\paragraph{Results} 
We first examine epoch-wise sleep-staging agreement (Cohen's $\kappa$) across families and datasets (Figure~\ref{fig:Sleep}A), where three patterns emerge. Performance varies strongly across datasets, with the best per-dataset $\kappa$ ranging from $0.62$ to $0.85$ (per-model results in App.~\ref{Appendix:Results-Sleep}). Across families, Supervised-Pre models achieve the strongest overall performance (binomial test, Table~\ref{tab:SleepStatistics}), but this advantage is largely attributable to their sequence-based architectures, which encode inter-epoch temporal context. An ablation with sequence length 1 (Sup-Pre $(\ell{=}1)$ in Figure~\ref{fig:Sleep}) removes most of this gap, consistent with the well-known role of sequence modeling in sleep staging~\cite{phan_seqsleepnet_2019}, a property not explicitly encoded in current FM architectures. Once sequence context is removed, Supervised-Pre $(\ell{=}1)$ match TS-FMs and EEG-FMs, suggesting that EEG-specific pretraining is less decisive for sleep staging than architectural design factors. Furthermore, replacing the linear probe with a small LSTM to recover this temporal context largely closes the gap between FMs and supervised models (App.~\ref{Appendix:Results-Sleep}). Within EEG-FMs, NeuroLM, REVE, and BIOT consistently rank among the top three across datasets (binomial test, Table~\ref{tab:SleepStatistics}). Lastly, a cross-dataset evaluation shows that FMs do not generalize well to new datasets without requiring target labels (App.~\ref{Appendix:Results-Sleep}).


Beyond epoch-wise staging, we evaluate three clinical axes: hypnogram-derived features, event detection, and recording-level diagnosis. \emph{Hypnogram features} (Figure~\ref{fig:Sleep}B shows REM latency, used for the diagnosis of hypersomnolence~\cite{barateau_narcolepsy_2022}; full feature list in App.~\ref{Appendix:Results-Sleep}) show substantial dataset variability, and crucially do not track Cohen's $\kappa$: As illustrated by DCSM (comparing Figure~\ref{fig:Sleep}A and B), strong staging agreement does not imply accurate feature recovery. Reporting both metrics offers a more clinically grounded view of model performance. For \emph{event detection} (Figure~\ref{fig:Sleep}C, microarousals), all model categories exceed the prevalence baseline in AUPRC, but inter-dataset variability remains pronounced; even without inter-epoch context, Supervised-Pre models outperform FMs, indicating that current FMs' representations encode microarousal events incompletely. \emph{Respiratory and limb-movement detection} are reported in App.~\ref{Appendix:Results-Sleep}. Finally, recording-level \emph{OSA diagnosis} succeeds when patients are compared to healthy controls but degrades when other conditions such as snoring or affective disorders are included; EEG-only embeddings therefore insufficiently encode disorder-specific features, and accurate OSA diagnosis should incorporate respiratory and oxygen-saturation channels. \emph{Cognitive-impairment diagnosis} is reported in App.~\ref{Appendix:Results-Sleep}.


\subsection{Brain Aging Estimation for Neurodegenerative Assessment}

Chronological age influences the spectral and temporal dynamics in EEG, particularly during sleep, where changes in micro- and macrostructure provide a window into neurophysiological aging \cite{SUN_age_changes}.  Predicting age from sleep EEG can therefore reveal whether learned representations capture broad, physiological patterns rather than only task-specific discriminative cues. In this sense, sleep-EEG age prediction provides a continuous regression task tied to known neurophysiological changes across the lifespan \cite{ENGEMANN2022119521}, but it has not been included in most existing EEG-FM benchmarks.
Age prediction itself is primarily useful as a representation-level probe. Its clinical relevance appears more promising through the "Brain Age Gap" modeling framework, where a model learns the expected relationship between EEG and age from healthy individuals. The brain-age gap (BAG) is defined as the difference between predicted and chronological age. Positive values indicate older-appearing EEG activity and is commonly interpreted as accelerated brain aging.  BAG provides a general, label-agnostic marker of brain health and has been associated with outcomes such as cognitive impairment, dementia risk, and mortality \cite{Sun2026SleepEEGBrainAgeDementiaRisk}. This makes brain age a useful benchmark for EEG-FMs: It tests whether representations capture continuous neurophysiological patterns and whether this signal can be translated into a clinically motivated digital biomarker.

\paragraph{Evaluation}
We evaluate brain aging in two complementary ways. First, we test age encoding by predicting chronological age from sleep EEG, reporting Pearson's $r$ for subject-level age ordering and MAE for error in years. Because raw MAE depends on cohort age range, we also report improvement over a train-mean baseline, which predicts the training-set mean age for all subjects. Improvement over this baseline aims to show improvement beyond learning the mean age of the cohort. Second, we assess whether the learned age signal can be used as a clinically meaningful marker. For this, models are trained on healthy subjects to learn a reference pattern of healthy aging, and then applied to an independent holdout set containing age-matched healthy and cognitively impaired (CI) subjects. We use the resulting brain-age gap to test whether the models indicate whether CI subjects show older-appearing EEG than healthy subjects of the same chronological age. We evaluate this both by comparing mean BAG between groups, and by testing whether BAG separates CI from healthy participants . This analysis is applied to all our model categories. More details on the metrics and the models can be found in App. \ref{Appendix:Experiments-Age} 

\begin{figure}[t]                           
    \centering                                      
    \includegraphics[width=1\textwidth]{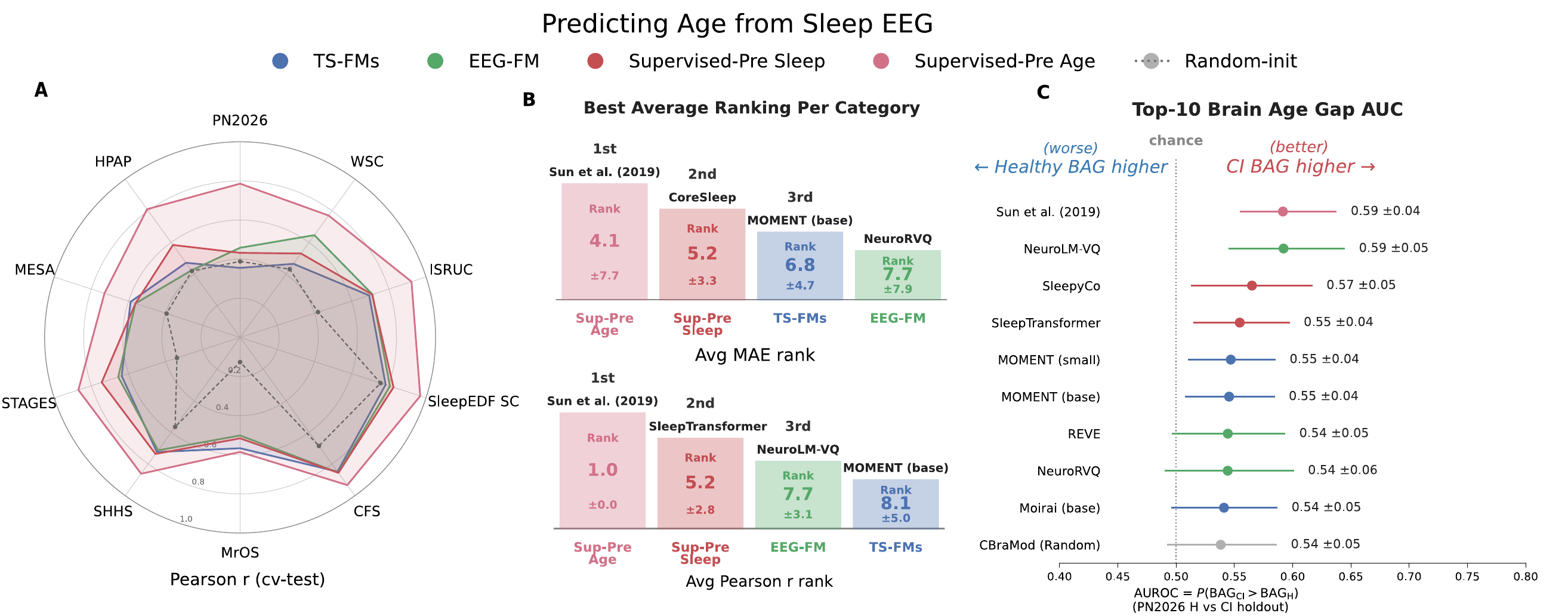}
    \caption{\textbf{Brain Age: Supervised Models outperform EEG-FMs for Brain Age Prediction, but performance does not translate to neurodegenerative assessment biomarker.} \textbf{(a)} Radar plot of performance (Pearson's r) of best model per category of models in different cohorts. Supervised Brain Age prediction outperforms the rest. \textbf{(b)} Best average ranking of the best model of each category for MAE and r metrics. \textbf{(c)} Top-10  AUROC for cognitive-impairment separation, using BAG (predicted minus chronological age). AUROC measures whether cognitively impaired subjects have higher BAG than age-matched healthy subjects. Although several models show weak positive separation, performance remains modest across model categories, indicating that accurate age prediction does not necessarily yield a robust BAG-based biomarker.
}    \label{fig:brain_age_combined}                                     
  \end{figure}

\paragraph{Results} 
Across cohorts, the feature-based brain-age model of Sun et al. \cite{sun_brain_2019} provides the strongest task-specific reference, achieving the largest MAE improvement over the dummy baseline and the highest Pearson correlation in most datasets (Figures~\ref{fig:brain_age_combined}A-B). To quantify whether this apparent consistency exceeds chance, we counted how often each model ranked among the top three within each dataset and compared this count against a chance top-3 rate using a one-sided exact binomial test, reported in App. \ref{Appendix:Results-BrainAge}, Tables \ref{stat_test_MAE}, \ref{stat_test_r}. Across all 26 models, Sun et al. is the only model significant for both MAE improvement and Pearson $r$. Among EEG-FMs specifically, performance is more fragmented. In the EEG-FM-only analysis, NeuroRVQ-EEG shows significant top-3 recurrence for MAE, while EEGPT and NeuroRVQ-EEG have the highest Pearson $r$ recurrence without reaching significance. Thus, while EEG-FMs encode useful age-related information, no single EEG-FM emerges as a stable winner across both metrics and cohorts. The benchmark instead reveals a heterogeneous landscape: Task-specific brain-age supervision is the most consistent reference, supervised sleep-staging models remain competitive, and different foundation models perform well in different datasets. Further ablations on the performance is discussed in App. \ref{Appendix:Results-BrainAge}. Finally, cross-dataset evaluation reveals that FMs exhibit limited generalization to unseen datasets in the absence of target-domain labels. This setting is particularly challenging due to domain shifts affecting both the input data distribution and the target-label distribution, as reflected by differences in age distributions across datasets.

\paragraph{FM-derived BAG Shows Limited Clinical Transfer}
We next evaluate BAG on an age-matched PhysioNet 2026 healthy/cognitive impairment (CI) holdout. BAG-based separation remains weak overall. The supervised Sun et al. model shows the clearest group-level shift ($\Delta\mathrm{BAG}=+1.99$ years; Figure S\ref{fig:dbag_per_subject_top10}) and reaches AUROC $=0.592$ (Figure \ref{fig:brain_age_combined}C). NeuroLM-VQ is the only EEG-FM, matching this AUROC with a similar positive BAG shift ($+1.72$ years), but both remain far from robust for healthy and CI separation. Sleep-staging models and smaller MOMENT variants show weak positive separation, whereas most remaining models are close to chance. Thus, BAG transfer to healthy-CI discrimination remains limited. The unclear transfer from models fitting age to BAG remain consistent with literature, where the relationship between fitting the age distribution and the residual error of prediction as a digital biomarker remains debatable \cite{jirsaraie2023multimodalbrainage}.

\subsection{Brain Computer Interfaces (BCI)}
\label{sec: BCI}

BCI paradigms are well represented in EEG-FM benchmarks~\citep{yang2023biot, cui2024neuro, jiang2024neurolm, liu2026eeg, yang2026eeg} with performance being reported on multiple datasets. The most extensive benchmark, EEG-FM-Benchmark~\citep{liu2026eeg}, reports performance on 7 BCI datasets spanning Motor Imagery (MI), Event-Related Potentials (ERP) and Steady-State Visually Evoked Potential (SSVEP). This work further expands on these efforts by evaluating EEG-FMs on a diverse set of $17$ datasets in total, adding also two cognitive tasks and one emotion recognition task. Additionally, we draw the attention to the fact that eye movements and gaze orientation can correlate with task conditions, causing models to rely on oculomotor activity instead of the intended neural signal. Low-frequency drifts can further contribute to signal non-stationarity over time~\citep{RAZA2019154} and post-cue saccades might obscure neural activity~\citep{bakas2025latent}. We investigate to what degree peformance is driven by such non-neural confounding factors.

\paragraph{Evaluation protocol}
 The model backbones remain frozen, and evaluation follows a leave-one-subject-out (LOSO) protocol to assess cross-subject generalization. The effect of non-neural confounding is assessed by comparing performance metrics with and without artifact correction preprocessing. We do so by applying a 4Hz highpass filter, a 40Hz lowpass and removing the first second post-cue in each trial, as proposed by \citep{bakas2025latent}.  Because decoding accuracies in finite-sample neural classification settings can substantially exceed theoretical chance levels purely by chance \citep{BetterThanChance}, we additionally report chance-corrected metrics and interpret performance relative to statistically meaningful baselines rather than raw accuracy alone. Further details can be round in App. \ref{Appendix:BCI_Exp}. 

\begin{figure*}[ht]
    \centering                                                              \includegraphics[width=\textwidth]{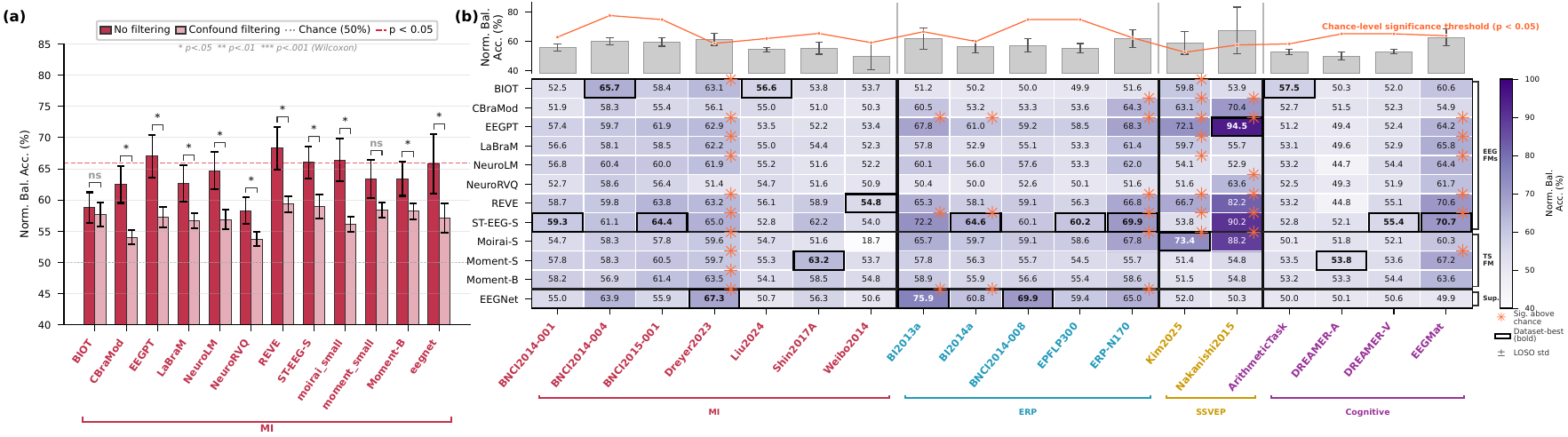}
    \caption{\textbf{BCI: MI performance is strongly driven by confounds} Linear probing performance of frozen EEG and time-series foundation models across 17 BCI datasets spanning four paradigms (MI, ERP, SSVEP, Cognitive) for token flattening. Metrics are reported as normalized balanced accuracy (\%), where 50\% equals chance regardless of the number of classes. The per-dataset significance threshold ($p<0.05$, binomial test \cite{BetterThanChance}) is shown as a dashed red line or orange markers. \textbf{(a)}~Normalized balanced accuracy on Motor Imagery datasets comparing embeddings extracted without confound filtering and with confound filtering bars for 8 EEG-FMs, 3 TS-FMs and EEGNet supervised model. Error bars denote the standard error of the mean across MI datasets (Wilcoxon signed-rank test; $*$\,$p<.05$, $**$\,$p<.01$, $***$\,$p<.001$). Confound filtering significantly affect performance, suggesting that the evaluated models rely primarily on artifacts. \textbf{(b)}~Per-dataset heatmap for 8 EEG-FMs and 3 TS-FM evaluated on confound filtered data. Cell color encodes normalized balanced accuracy (40-100\%; darker~$=$~higher); each cell reports the score (bold for dataset-best) with LOSO standard deviation ($\pm$). Orange stars (\textcolor{orange}{$\ast$}) mark results significantly above chance. Note that due to the varying number of classes and the low sample counts, the significance level is much higher than chance level and varies across datasets.}

  \label{fig:BCI}                                     
  \end{figure*}

 
\paragraph{Results} As shown in Fig.~\ref{fig:BCI}A, our investigation suggests that FMs are still prone to capturing non-neural confounding sources, such as eye gaze. In MI paradigms, not filtering or removing the first post-cue second significantly increases performance above statistically significant chance level for most EEG-FMs, while confound filtering drops the performance below chance level. Similar analyses on the remaining paradigms did not further affect model performance, and are reported in the App. \ref{Appendix:BCI_Res}. For linear probing, we evaluated two strategies for aggregating the extracted embeddings: mean pooling across tokens and flattening the full embedding representation. As flattening generally yielded stronger downstream performance, we report the flattened-embedding results in the main text, while the corresponding mean-pooled results and a detailed comparison of both strategies are provided in App.~\ref{Appendix:BCI_Res}. Furthermore, as shown in Fig.~\ref{fig:BCI}B, although linear probing often exceeds the normalized balanced accuracy chance level of $50\%$, it frequently remains below the statistically significant chance threshold under LOSO evaluation, suggesting that FMs still do not reliably extract generalizable task-relevant information across subjects. In terms of EEG-FM variability, the results demonstrate substantial differences in how well EEG-FMs generalize across BCI paradigms and datasets. While EEGPT and ST-EEG-S achieve the strongest performance on MI tasks, other models such as REVE and NeuroLM show superior performance on ERP and cognitive benchmarks, indicating that different EEG-FMs capture complementary neural representations. Importantly, no single model consistently outperforms others across all paradigms, suggesting that current unsupervised EEG representation learning approaches remain paradigm-dependent rather than universally transferable.

\section{Discussion} \label{sec:discussion}

\textbf{Are EEG-FM models generally consistent within domains?} The first goal of this benchmark was to provide an extensive analysis of cross-dataset performance within the domains of epileptic seizure detection, sleep staging, brain age prediction and BCI. While performances for all models naturally varied significantly between datasets of every domain, we investigated to what degree different domains have a consistent set of best performing EEG-FMs. In epilepsy, sleep and BCIs, Neuro-LM and REVE tend to end up consistently in the top-3 ranked EEG-FMs, while in the novel task (brain age), no clear top competitors emerge.

\textbf{How do EEG-FMs compare to other model families?} The linear probing performances on EEG-FMs and pre-trained supervised models vary significantly between different domains. In seizure detection, EEG-FMs consistently outperform their supervised counterparts with the only exceptions being the datasets that the supervised models were actually pretrained on. In sleep, EEG-FMs perform similarly to pretrained supervised models on a window-per-window level, but fall significantly behind when the supervised models are trained to operate in a sequence-to-sequence manner, which fall outside of the capabilities of linear probing. In brain age prediction, EEG-FMs perform worse on the task of chronological age prediction, yet perform on par when it comes to the clinically more relevant task of estimating the BAG. Finally, EEG-FMs struggle to reach motor imagery performance significantly above chance level. The comparison with general-purpose time-series FMs is more consistent across domains: We do not find evidence that EEG-FMs are able to consistently outperform time-series FMs, despite their apparent advantage in only having to model EEG. We observe the same trends among the model families when evaluated in a leave-one-dataset-out (LODO) framework as well.

\textbf{The importance of clinically relevant metrics!} Across domains, we observe that standard metrics insufficiently address clinical utility. In epilepsy, we observe that relying on AUROC for model selection leads to diverging results from using the more application-oriented EventSens@FA/h (Figure \ref{fig:metric-rank-disagreement}). In sleep, models with similar window-level $\kappa$-score show large differences in REM-latency (Figure \ref{fig:Sleep}B). Similarly in brain age, models with substantially different MAE on age estimation do not necessarily differ in BAG. Finally, in BCI, applying appropriate preprocessing to prevent non-neural confounding often makes the difference between a model performing above significance level or not. 

\textbf{Usage of the benchmark} Since the benchmark is based on public data, it is unable to serve as a public scoreboard as there is no verifiable way to ensure that a given FM has not been pre-trained on any of the downstream datasets included in the benchmark. Within the results reported here, we have simply indicated when this is the case, so their respective performance can be taken with a grain of salt. Furthermore, it is difficult to establish general claims about desirable components of well-performing FMs, since comparing two of them immediately compares a whole range of differing design choices, ranging from pretraining data composition, to channel matching and pretext task. Instead, we present this benchmark as a way for researchers to perform thorough ablation themselves by fixing anything downstream of the encoder and evaluate the impact of their design choices on clinically grounded metrics across domains.


\textbf{Limitations and conclusions} First, we have focused our efforts on a single evaluation protocol in linear probing, explicitly not including other paradigms such as fine-tuning and few-shot learning. Linear probing was chosen because it isolates what the frozen representation itself encodes: a linear readout adds no capacity of its own, so differences between models are attributable to pretraining rather than to how much of the task a downstream head can learn on top. It also keeps the comparison identical across three model families with very different architectures, which fine-tuning would not, and it matches the promise being tested, namely an off-the-shelf backbone reusable across domains without per-task adaptation. The trade-off is that our probe can only recover information that is linearly accessible, and the sleep results make this concrete: replacing the linear probe with a small LSTM largely closes the gap to sequence-based supervised models, so a representation can hold task-relevant structure that a linear readout cannot express. Extending NeuroAtlas to fine-tuning and few-shot regimes would therefore be valuable, since a model's ranking under linear probing need not predict its ranking once adaptation is allowed, and the amount of target data a representation needs to become useful is itself informative about the quality of pretraining.

Secondly, our baseline for randomly initialized FMs was limited to a single model, CBraMod, which performed remarkably decently. While we believe a more dedicated study comparing each pre-trained FM against its own randomly initalized counterpart can be highly illuminating about the benefits of specific components of pretraining, we kept it outside the scope of this study. In addition, we were not able to include promising models such as BrainWave \cite{yuan2024brainwave} due to their implementation not being fully available. Finally, while we show that more clinically relevant,  the domain-specific metrics can significantly disagree with commonly used ML metrics, even these metrics are still insufficient to fully address the question of clinical utility. We hope these observations open up a debate in both clinical and engineering communities which clinically grounded properties \textit{must} be encoded in FM representations. In the end, true clinical utility should be measured by the effect on patient outcomes when these models are involved in the clinical decision-making process.  Despite these limitations, NeuroAtlas significantly advances the state of EEG-FM benchmarking both in scale and rigorous evaluation metrics.

\clearpage
\newpage
\section*{Acknowledgements}

This work was supported by the Flemish Government under the METHUSALEM programme (METH/26/003, "Methusalem–BioMedAI: Explainable and generative AI for accelerating biomedical discoveries: from genome to function") and by the Flemish Government AI Research Program; several of the authors are affiliated with Leuven.AI of KU Leuven Institute for AI, B-3000 Leuven, Belgium. This work was also  supported by the European Union’s Horizon Europe research and innovation programme under the Marie Skłodowska-Curie Actions Doctoral Network InSilicoHealth (Grant Agreement No. 101169278). Further support was provided by KU Leuven through project C3/25/059 ("A Multicenter RCT on the impact of Wearable EEG and AI-Based Seizure Detection in absence seizure management"), and by the Research Foundation – Flanders (FWO) through research projects G046925N ("Task- and device-agnostic Artificial Intelligence (AI) for EEG analysis"), and G0C9623N ("Artificial Intelligence (AI) for data-driven personalised medicine"), as well as FWO PhD fellowships 1166926N, 1107725N, 1S30225N, 1S12523N, 1SB5922N, 1S61822N and 1SH4Z24N. The collaboration was also supported by the MIT-Belgium–KU Leuven Seed Fund. The resources and services used in this work were provided by the VSC (Flemish Supercomputer Center), funded by the Research Foundation - Flanders (FWO) and the Flemish Government. We acknowledge LUMI-BE for awarding this project access to the LUMI supercomputer, owned by the EuroHPC Joint Undertaking, hosted by CSC (Finland) and the LUMI consortium through a LUMI-BE Regular Access call. LUMI-BE is joint effort from BELSPO (federal), SPW Économie, Emploi, Recherche (Wallonia), Department of Economy, Science \& Innovation (Flanders) and Innoviris (Brussels).

\section*{Data Licenses}

NeuroAtlas is built entirely on data collected and shared by others; we thank the investigators, clinical staff and participants of all 42 cohorts.

On epilepsy, CHB-MIT \cite{shoeb2009chbmit} is distributed via PhysioNet under ODC-BY 1.0 and Siena \cite{detti2020siena} under CC BY 4.0; Helsinki \cite{stevenson2019helsinki} via Zenodo (10.5281/zenodo.2547147); Bonn \cite{andrzejak2001indications} by the Department of Epileptology, University of Bonn, and NMT \cite{khan2022nmt} by the National Medical Trust Hospital, Karachi, under CC BY-SA 4.0; and SeizeIT2 \cite{bhagubai2025seizeit2} openly on OpenNeuro (ds005873, BIDS). TUSZ \cite{shah2018tusz} and TUAB \cite{tuab} were obtained from the Temple University Hospital EEG Corpus \cite{obeid2016temple} under its data use agreement; corpus development was supported by the National Institutes of Health (NHGRI award U01HG008468), the National Science Foundation (IIP-1622765, CNS-1305190), DARPA (D13AP00065), and Temple University. Epilepsiae \cite{klatt2012epilepsiae}, SeizeIT1 \cite{sz1}, and SeizeIT2 \cite{bhagubai2025seizeit2} are shared under managed access, and we accessed them under those terms: Epilepsiae can be licensed on application through the Epilepsiae database (https://epilepsiae.uniklinik-freiburg.de), while SeizeIT1-2 were collected at UZ Leuven and partner centres. It can be made available after agreeing on appropriate data-sharing approvals. 

On sleep and brain age, Sleep-EDF Expanded \cite{Kemp2018-bk_sleepedf}, UCDDB \cite{McNicholas2004-dq_ucddb} and HMC \cite{PhysioNet-hmc-sleep-staging-1.1} are distributed via PhysioNet under ODC-BY 1.0, and PN2026 \cite{Westover2023-ca_physionet}, a subset of the Human Sleep Project, under the Challenge terms. ISRUC-Sleep \cite{ISRUC}, the Dreem Open Datasets \cite{PaperDOD, Thorey2025-ke_DOD} and DCSM \cite{Perslev2021-cp_DCSM} were released openly by their respective centres. MASS \cite{mass} is shared through C\'EAMS and the Borealis repository on local ethics approval. Seven further cohorts were accessed through the National Sleep Research Resource (NSRR) \cite{Zhang2018-sw_nssr}, whose required acknowledgements follow.

The Cleveland Family Study (CFS) \cite{Redline1995-yz_cfs} was supported by grants from the National Institutes of Health (HL46380, M01 RR00080-39, T32-HL07567, RO1-46380). The Home Positive Airway Pressure study (HomePAP) \cite{Rosen2012-yy_homepap} was supported by the American Sleep Medicine Foundation 38-PM-07 Grant: Portable Monitoring for the Diagnosis and Management of OSA. The Multi-Ethnic Study of Atherosclerosis (MESA) \cite{Chen2015-gq_mesa} Sleep Ancillary study was funded by NIH-NHLBI Association of Sleep Disorders with Cardiovascular Health Across Ethnic Groups (RO1 HL098433). MESA is supported by NHLBI funded contracts HHSN268201500003I, N01-HC-95159, N01-HC-95160, N01-HC-95161, N01-HC-95162, N01-HC-95163, N01-HC-95164, N01-HC-95165, N01-HC-95166, N01-HC-95167, N01-HC-95168 and N01-HC-95169 from the National Heart, Lung, and Blood Institute, and by cooperative agreements UL1-TR-000040, UL1-TR-001079, and UL1-TR-001420 funded by NCATS. The National Heart, Lung, and Blood Institute provided funding for the ancillary MrOS Sleep Study \cite{Blackwell2011-ca_mros}, ``Outcomes of Sleep Disorders in Older Men,'' under the following grant numbers: R01 HL071194, R01 HL070848, R01 HL070847, R01 HL070842, R01 HL070841, R01 HL070837, R01 HL070838, and R01 HL070839. The Sleep Heart Health Study (SHHS) \cite{Quan1997SleepHeartHealth} was supported by National Heart, Lung, and Blood Institute cooperative agreements U01HL53916 (University of California, Davis), U01HL53931 (New York University), U01HL53934 (University of Minnesota), U01HL53937 and U01HL64360 (Johns Hopkins University), U01HL53938 (University of Arizona), U01HL53940 (University of Washington), U01HL53941 (Boston University), and U01HL63463 (Case Western Reserve University). This research has been conducted using the STAGES -- Stanford Technology, Analytics and Genomics in Sleep Resource funded by the Klarman Family Foundation. The investigators of the STAGES study contributed to the design and implementation of the STAGES cohort and/or provided data and/or collected biospecimens, but did not necessarily participate in the analysis or writing of this report. The full list of STAGES investigators can be found at the project website. This Wisconsin Sleep Cohort (WSC) Study \cite{Young2009WisconsinSleepCohort} was supported by the U.S. National Institutes of Health, National Heart, Lung, and Blood Institute (R01HL62252), National Institute on Aging (R01AG036838, R01AG058680), and the National Center for Research Resources (1UL1RR025011). The National Sleep Research Resource was supported by the U.S. National Institutes of Health, National Heart Lung and Blood Institute (R24 HL114473, 75N92019R002).

Most BCI datasets were accessed through MOABB \cite{moabb2018}, whose identifiers we adopt throughout; the cognitive and affective datasets came directly from their original repositories. Among the motor imagery datasets, BNCI2014-001, BNCI2014-004 \cite{tangermann2012review} and BNCI2015-001 \cite{faller2012autocalibration} are distributed via the BNCI Horizon 2020 database under CC BY-ND 4.0 and CC BY-NC-ND 4.0, and Dreyer2023 \cite{pillette2021experimenters} and Liu2024 \cite{liu2024eeg} under CC BY 4.0; we also use Weibo2014 \cite{yi2014evaluation} and Shin2017A \cite{shin2016open}, from the latter of which we take the EEG modality only. Among the ERP datasets, BI2013a \cite{vaineau2019brain} and BI2014a \cite{korczowski2019brain} are distributed via Zenodo under CC BY 1.0 and CC BY 4.0 and ErpCore2021-N170 \cite{kappenman2021erp} via OSF under CC BY 4.0, BNCI2014-008 \cite{riccio2013attention} via the BNCI Horizon 2020 database under CC BY-NC-ND 4.0, and EPFLP300 \cite{hoffmann2008efficient} is available at http://bci.epfl.ch/p300. The SSVEP datasets Nakanishi2015 \cite{nakanishi2015comparison} and Kim2025BetaRange \cite{kim202540} are available from the SCCN archive at UC San Diego and from figshare (10.6084/m9.figshare.28806815.v2) respectively. Finally, ArithmeticTask \cite{rodriguez2019tracking} is available via OSF under CC0 1.0, while EEGMat \cite{zyma2019electroencephalograms} and DREAMER \cite{dreamer} are covered by the PhysioNet and restricted-access terms stated above. We thank the MOABB maintainers, and the NSRR, C\'EAMS, PhysioNet and OpenNeuro teams, for making these resources available.

{\small
\bibliographystyle{plain}
\bibliography{citations}  

@article{nie2022time,
  title={A time series is worth 64 words: Long-term forecasting with transformers},
  author={Nie, Yuqi and Nguyen, Nam H and Sinthong, Phanwadee and Kalagnanam, Jayant},
  journal={arXiv preprint arXiv:2211.14730},
  year={2022}
}

@article{shi2024fome,
  title={FoME: A foundation model for EEG using adaptive temporal-lateral attention scaling},
  author={Shi, Enze and Zhao, Kui and Yuan, Qilong and Wang, Jiaqi and Hu, Huawen and Yu, Sigang and Zhang, Shu},
  journal={arXiv preprint arXiv:2409.12454},
  year={2024}
}

@article{wu2022timesnet,
  title={Timesnet: Temporal 2d-variation modeling for general time series analysis},
  author={Wu, Haixu and Hu, Tengge and Liu, Yong and Zhou, Hang and Wang, Jianmin and Long, Mingsheng},
  journal={arXiv preprint arXiv:2210.02186},
  year={2022}
}

@inproceedings{goswami2024moment,
  title={MOMENT: A Family of Open Time-series Foundation Models},
  author={Mononito Goswami and Konrad Szafer and Arjun Choudhry and Yifu Cai and Shuo Li and Artur Dubrawski},
  booktitle={International Conference on Machine Learning},
  year={2024}
}

@inproceedings{woo2024unified,
  title={Unified training of universal time series forecasting transformers},
  author={Woo, Gerald and Liu, Chenghao and Kumar, Akshat and Xiong, Caiming and Savarese, Silvio and Sahoo, Doyen},
  booktitle={Forty-first International Conference on Machine Learning},
  year={2024}
}

@article{
ansari2024chronos,
title={Chronos: Learning the Language of Time Series},
author={Abdul Fatir Ansari and Lorenzo Stella and Ali Caner Turkmen and Xiyuan Zhang and Pedro Mercado and Huibin Shen and Oleksandr Shchur and Syama Sundar Rangapuram and Sebastian Pineda Arango and Shubham Kapoor and Jasper Zschiegner and Danielle C. Maddix and Hao Wang and Michael W. Mahoney and Kari Torkkola and Andrew Gordon Wilson and Michael Bohlke-Schneider and Bernie Wang},
journal={Transactions on Machine Learning Research},
issn={2835-8856},
year={2024},
url={https://openreview.net/forum?id=gerNCVqqtR},
}

@article{thapa2024sleepfm,
  title={SleepFM: multi-modal representation learning for sleep across brain activity, ECG and respiratory signals},
  author={Thapa, Rahul and He, Bryan and Kj{\ae}r, Magnus Ruud and Moore, Hyatt and Ganjoo, Gauri and Mignot, Emmanuel and Zou, James},
  journal={arXiv preprint arXiv:2405.17766},
  year={2024}
}

@article{barmpas2025neurorvq,
  title={NeuroRVQ: Multi-Scale EEG Tokenization for Generative Large Brainwave Models},
  author={Barmpas, Konstantinos and Lee, Na and Koliousis, Alexandros and Panagakis, Yannis and Adamos, Dimitrios A and Laskaris, Nikolaos and Zafeiriou, Stefanos},
  journal={arXiv preprint arXiv:2510.13068},
  year={2025}
}

@inproceedings{cui2024neuro,
  title={Neuro-gpt: Towards a foundation model for eeg},
  author={Cui, Wenhui and Jeong, Woojae and Th{\"o}lke, Philipp and Medani, Takfarinas and Jerbi, Karim and Joshi, Anand A and Leahy, Richard M},
  booktitle={2024 IEEE International Symposium on Biomedical Imaging (ISBI)},
  pages={1--5},
  year={2024},
  organization={IEEE}
}

@article{wang2024cbramod,
  title={Cbramod: A criss-cross brain foundation model for eeg decoding},
  author={Wang, Jiquan and Zhao, Sha and Luo, Zhiling and Zhou, Yangxuan and Jiang, Haiteng and Li, Shijian and Li, Tao and Pan, Gang},
  journal={arXiv preprint arXiv:2412.07236},
  year={2024}
}

@article{yang2023biot,
  title={Biot: Biosignal transformer for cross-data learning in the wild},
  author={Yang, Chaoqi and Westover, M and Sun, Jimeng},
  journal={Advances in Neural Information Processing Systems},
  volume={36},
  pages={78240--78260},
  year={2023}
}

@article{sun2019brain,
  title={Brain age from the electroencephalogram of sleep},
  author={Sun, Haoqi and Paixao, Luis and Oliva, Jefferson T and Goparaju, Balaji and Carvalho, Diego Z and van Leeuwen, Kicky G and Akeju, Oluwaseun and Thomas, Robert J and Cash, Sydney S and Bianchi, Matt T and others},
  journal={Neurobiology of aging},
  volume={74},
  pages={112--120},
  year={2019},
  publisher={Elsevier}
}

@article{liu2025mirepnet,
  title={{MIR}ep{N}et: A Pipeline and Pre-trained Model for {EEG}-Based Motor Imagery Classification},
  author={Dingkun Liu and Zhu Chen and Jingwei Luo and Shijie Lian and Yuheng Chen and Shaojie Hou and Xiaolian Zhu and Dongrui Wu},
  journal={Knowledge-Based Systems},
  volume = {343},
  pages = {115966},
  year={2026},
  publisher={Elsevier},
  doi = {https://doi.org/10.1016/j.knosys.2026.115966}
}

@inproceedings{m2023deepsoz,
  title={DeepSOZ: A robust deep model for joint temporal and spatial seizure onset localization from multichannel EEG data},
  author={M. Shama, Deeksha and Jing, Jiasen and Venkataraman, Archana},
  booktitle={International Conference on Medical Image Computing and Computer-Assisted Intervention},
  pages={184--194},
  year={2023},
  organization={Springer}
}

@article{lawhern2018eegnet,
  title={EEGNet: a compact convolutional neural network for EEG-based brain--computer interfaces},
  author={Lawhern, Vernon J and Solon, Amelia J and Waytowich, Nicholas R and Gordon, Stephen M and Hung, Chou P and Lance, Brent J},
  journal={Journal of neural engineering},
  volume={15},
  number={5},
  pages={056013},
  year={2018},
  publisher={iOP Publishing}
}

@article{ouahidi2025reve,
  title={REVE: A Foundation Model for EEG--Adapting to Any Setup with Large-Scale Pretraining on 25,000 Subjects},
  author={Ouahidi, Yassine El and Lys, Jonathan and Th{\"o}lke, Philipp and Farrugia, Nicolas and Pasdeloup, Bastien and Gripon, Vincent and Jerbi, Karim and Lioi, Giulia},
  journal={arXiv preprint arXiv:2510.21585},
  year={2025}
}

@article{wang2024eegpt,
  title={Eegpt: Pretrained transformer for universal and reliable representation of eeg signals},
  author={Wang, Guagnyu and Liu, Wenchao and He, Yuhong and Xu, Cong and Ma, Lin and Li, Haifeng},
  journal={Advances in Neural Information Processing Systems},
  volume={37},
  pages={39249--39280},
  year={2024}
}

@inproceedings{
jiang2024neurolm, 
title={Neuro{LM}: A Universal Multi-task Foundation Model for Bridging the Gap between Language and {EEG} Signals},
author={Wei-Bang Jiang and Yansen Wang and Bao-Liang Lu and Dongsheng Li},
booktitle={The Thirteenth International Conference on Learning Representations},
year={2025},
url={https://openreview.net/forum?id=Io9yFt7XH7}
}

@inproceedings{
jiang2024large,
title={Large Brain Model for Learning Generic Representations with Tremendous {EEG} Data in {BCI}},
author={Wei-Bang Jiang and Li-Ming Zhao and Bao-Liang Lu},
booktitle={The Twelfth International Conference on Learning Representations},
year={2024},
url={https://openreview.net/forum?id=QzTpTRVtrP}
}

@article{kontras2024core,
  title={Core-sleep: A multimodal fusion framework for time series robust to imperfect modalities},
  author={Kontras, Konstantinos and Chatzichristos, Christos and Phan, Huy and Suykens, Johan and De Vos, Maarten},
  journal={IEEE Transactions on Neural Systems and Rehabilitation Engineering},
  volume={32},
  pages={840--849},
  year={2024},
  publisher={IEEE}
}

@article{wu2025adabrain,
  title={Adabrain-bench: Benchmarking brain foundation models for brain-computer interface applications},
  author={Wu, Jiamin and Ren, Zichen and Wang, Junyu and Zhu, Pengyu and Song, Yonghao and Liu, Mianxin and Zheng, Qihao and Bai, Lei and Ouyang, Wanli and Song, Chunfeng},
  journal={arXiv preprint arXiv:2507.09882},
  year={2025}
}

@misc{kastrati_eeg-bench_2025,
	title = {{EEG}-{Bench}: {A} {Benchmark} for {EEG} {Foundation} {Models} in {Clinical} {Applications}},
	copyright = {Creative Commons Attribution 4.0 International},
	shorttitle = {{EEG}-{Bench}},
	url = {https://arxiv.org/abs/2512.08959},
	doi = {10.48550/ARXIV.2512.08959},
	language = {en},
	urldate = {2026-05-04},
	publisher = {arXiv},
	author = {Kastrati, Ard and Bürki, Josua and Lauer, Jonas and Xuan, Cheng and Iaquinto, Raffaele and Wattenhofer, Roger},
	year = {2025},
	note = {Version Number: 1},
}

@inproceedings{yang2026eeg,
  title={Are EEG foundation models worth it? comparative evaluation with traditional decoders in diverse BCI tasks},
  author={Yang, Liuyin and Sun, Qiang and Li, Ang and Van Hulle, Marc M},
  booktitle={The Fourteenth International Conference on Learning Representations},
  year={2026}
}

@article{yao_foundation_2025,
	title = {Foundation models for {EEG} decoding: current progress and prospective research},
	volume = {22},
	issn = {1741-2560, 1741-2552},
	shorttitle = {Foundation models for {EEG} decoding},
	url = {https://iopscience.iop.org/article/10.1088/1741-2552/ae17e9},
	doi = {10.1088/1741-2552/ae17e9},
	language = {en},
	number = {6},
	urldate = {2026-05-04},
	journal = {Journal of Neural Engineering},
	author = {Yao, Yuxuan and Wang, Hongbo and Chen, Li and Peng, Yiheng and Luo, Jingjing},
	month = dec,
	year = {2025},
	pages = {061002},
}

@misc{xiong_eeg-fm-bench_2026,
	title = {{EEG}-{FM}-{Bench}: {A} {Comprehensive} {Benchmark} for the {Systematic} {Evaluation} of {EEG} {Foundation} {Models}},
	shorttitle = {{EEG}-{FM}-{Bench}},
	url = {http://arxiv.org/abs/2508.17742},
	doi = {10.48550/arXiv.2508.17742},
	language = {en},
	urldate = {2026-05-04},
	publisher = {arXiv},
	author = {Xiong, Wei and Li, Jiangtong and Li, Jie and Zhu, Kun and Jiang, Changjun},
	month = feb,
	year = {2026},
	note = {arXiv:2508.17742 [eess]},
}

@misc{shen_brain4fms_2026,
	title = {{Brain4FMs}: {A} {Benchmark} of {Foundation} {Models} for {Electrical} {Brain} {Signal}},
	copyright = {arXiv.org perpetual, non-exclusive license},
	shorttitle = {{Brain4FMs}},
	url = {https://arxiv.org/abs/2602.11558},
	doi = {10.48550/ARXIV.2602.11558},
	language = {en},
	urldate = {2026-05-04},
	publisher = {arXiv},
	author = {Shen, Fanqi and Yang, Enhong and Li, Jiahe and Hong, Junru and Pan, Xiaoran and Yuan, Zhizhang and Li, Meng and Yang, Yang},
	year = {2026},
	note = {Version Number: 1},
}

@article{liu2026eeg,
  title={EEG Foundation Models: Progresses, Benchmarking, and Open Problems},
  author={Liu, Dingkun and Chen, Yuheng and Chen, Zhu and Cui, Zhenyao and Wen, Yaozhi and An, Jiayu and Luo, Jingwei and Wu, Dongrui},
  journal={arXiv preprint arXiv:2601.17883},
  year={2026}
}

@article{kuruppu2026eeg,
  title={EEG foundation models: a critical review of current progress and future directions},
  author={Kuruppu, Gayal and Wagh, Neeraj and Kremen, Vaclav and Varatharajah, Yogatheesan},
  journal={Journal of neural engineering},
  volume={23},
  number={2},
  pages={021001},
  year={2026},
  publisher={IOP Publishing}
}

@misc{neurorvq,
      title={{NeuroRVQ: Multi-Scale EEG Tokenization for Generative Large Brainwave Models}}, 
      author={Konstantinos Barmpas and Na Lee and Alexandros Koliousis and Yannis Panagakis and Dimitrios A. Adamos and Nikolaos Laskaris and Stefanos Zafeiriou},
      year={2025},
      eprint={2510.13068},
      archivePrefix={arXiv},
      primaryClass={cs.LG},
      url={https://arxiv.org/abs/2510.13068}, 
}

@article{moabb2018,
  title={{MOABB: trustworthy algorithm benchmarking for BCIs}},
  author={Jayaram, Vinay and Barachant, Alexandre},
  journal={Journal of neural engineering},
  volume={15},
  number={6},
  pages={066011},
  year={2018},
  publisher={IOP Publishing}
}

@inproceedings{GuePapTan23a,
 author = {Pierre Guetschel and Théodore Papadopoulo and Michael Tangermann},
 booktitle = {Proceedings of the 10th International Brain-Computer Interface Meeting 2023},
 doi = {10.3217/978-3-85125-962-9-82},
 grant = {BLBT,DCC},
 month = {June},
 title = {Neural network transfer learning with fast calibration for mental imagery decoding},
 year = {2023}
}

@article{singh2021mental,
  title={{Mental workload estimation based on physiological features for pilot-UAV teaming applications}},
  author={Singh, Gaganpreet and Chanel, Caroline PC and Roy, Rapha{\"e}lle N},
  journal={Frontiers in Human Neuroscience},
  volume={15},
  pages={692878},
  year={2021},
  publisher={Frontiers Media SA}
}

@ARTICLE{dreamer,
  author={Katsigiannis, Stamos and Ramzan, Naeem},
  journal={IEEE Journal of Biomedical and Health Informatics}, 
  title={{DREAMER: A Database for Emotion Recognition Through EEG and ECG Signals From Wireless Low-cost Off-the-Shelf Devices}}, 
  year={2018},
  volume={22},
  number={1},
  pages={98-107},
  doi={10.1109/JBHI.2017.2688239}}

@article{bakas2025latent,
  title={{Latent alignment in deep learning models for EEG decoding}},
  author={Bakas, Stylianos and Ludwig, Siegfried and Adamos, Dimitrios A and Laskaris, Nikolaos and Panagakis, Yannis and Zafeiriou, Stefanos},
  journal={Journal of neural engineering},
  volume={22},
  number={1},
  pages={016047},
  year={2025},
  publisher={IOP Publishing}
}

@article{zhu2010survey,
  title={{A survey of stimulation methods used in SSVEP-based BCIs}},
  author={Zhu, Danhua and Bieger, Jordi and Garcia Molina, Gary and Aarts, Ronald M},
  journal={Computational intelligence and neuroscience},
  volume={2010},
  number={1},
  pages={702357},
  year={2010},
  publisher={Wiley Online Library}
}

@book{luck2012event,
  editor    = {Kappenman, Emily S. and Luck, Steven J.},
  title     = {The Oxford Handbook of Event-Related Potential Components},
  series    = {Oxford Library of Psychology},
  year      = {2011},
  publisher = {Oxford University Press},
  doi       = {10.1093/oxfordhb/9780195374148.001.0001},
  url       = {https://doi.org/10.1093/oxfordhb/9780195374148.001.0001},
  urldate   = {2026-05-15},
  note      = {Online edition, Oxford Academic, 18 September 2012}
}

@ARTICLE{MIBCI_YuanBe20214,
  author={Yuan, Han and He, Bin},
  journal={IEEE Transactions on Biomedical Engineering}, 
  title={{Brain–Computer Interfaces Using Sensorimotor Rhythms: Current State and Future Perspectives}}, 
  year={2014},
  volume={61},
  number={5},
  pages={1425-1435},
  doi={10.1109/TBME.2014.2312397}}

@article{AHN2015103,
title = {{Performance variation in motor imagery brain–computer interface: A brief review}},
journal = {Journal of Neuroscience Methods},
volume = {243},
pages = {103-110},
year = {2015},
issn = {0165-0270},
doi = {https://doi.org/10.1016/j.jneumeth.2015.01.033},
url = {https://www.sciencedirect.com/science/article/pii/S0165027015000400},
author = {Minkyu Ahn and Sung Chan Jun}
}

@article{BetterThanChance,
title = {Exceeding chance level by chance: The caveat of theoretical chance levels in brain signal classification and statistical assessment of decoding accuracy},
journal = {Journal of Neuroscience Methods},
volume = {250},
pages = {126-136},
year = {2015},
note = {Cutting-edge EEG Methods},
issn = {0165-0270},
doi = {https://doi.org/10.1016/j.jneumeth.2015.01.010},
url = {https://www.sciencedirect.com/science/article/pii/S0165027015000114},
author = {Etienne Combrisson and Karim Jerbi}
}

@article{RAZA2019154,
title = {Covariate shift estimation based adaptive ensemble learning for handling non-stationarity in motor imagery related EEG-based brain-computer interface},
journal = {Neurocomputing},
volume = {343},
pages = {154-166},
year = {2019},
note = {Learning in the Presence of Class Imbalance and Concept Drift},
issn = {0925-2312},
doi = {https://doi.org/10.1016/j.neucom.2018.04.087},
url = {https://www.sciencedirect.com/science/article/pii/S0925231219301560},
author = {Haider Raza and Dheeraj Rathee and Shang-Ming Zhou and Hubert Cecotti and Girijesh Prasad}
}

@article{li_evaluating_2022,
	title = {Evaluating the {Bidirectional} {Causal} {Association} {Between} {Daytime} {Napping} and {Alzheimer}’s {Disease} {Using} {Mendelian} {Randomization}},
	volume = {89},
	issn = {13872877, 18758908},
	url = {https://journals.sagepub.com/doi/full/10.3233/JAD-220497},
	doi = {10.3233/JAD-220497},
	language = {en},
	number = {4},
	urldate = {2025-06-05},
	journal = {Journal of Alzheimer's Disease},
	author = {Li, Sijie and Liu, Bian and Li, Qing-hao and Zhang, Yan and Zhang, Haihua and Gao, Shan and Wang, Longcai and Wang, Tao and Han, Zhifa and Liu, Guiyou and Wang, Kun},
	editor = {Yu, Jin-Tai},
	month = oct,
	year = {2022},
	pages = {1315--1322},
}

@article{christensen_sleep_2016,
	title = {Sleep stability and transitions in patients with idiopathic {REM} sleep behavior disorder and patients with {Parkinson}’s disease},
	volume = {127},
	issn = {13882457},
	url = {https://linkinghub.elsevier.com/retrieve/pii/S1388245715001716},
	doi = {10.1016/j.clinph.2015.03.006},
	language = {en},
	number = {1},
	urldate = {2025-10-16},
	journal = {Clinical Neurophysiology},
	author = {Christensen, Julie Anja Engelhard and Jennum, Poul and Koch, Henriette and Frandsen, Rune and Zoetmulder, Marielle and Arvastson, Lars and Christensen, Søren Rahn and Sorensen, Helge Bjarrup Dissing},
	month = jan,
	year = {2016},
	pages = {537--543},
}

@article{lee_interrater_2022,
	title = {Interrater reliability of sleep stage scoring: a meta-analysis},
	volume = {18},
	issn = {1550-9389, 1550-9397},
	shorttitle = {Interrater reliability of sleep stage scoring},
	url = {http://jcsm.aasm.org/doi/10.5664/jcsm.9538},
	doi = {10.5664/jcsm.9538},
	language = {en},
	number = {1},
	urldate = {2025-06-18},
	journal = {Journal of Clinical Sleep Medicine},
	author = {Lee, Yun Ji and Lee, Jae Yong and Cho, Jae Hoon and Choi, Ji Ho},
	month = jan,
	year = {2022},
	pages = {193--202},
}

@book{aasm2023,
  title   = {The AASM Manual for the Scoring of Sleep and Associated Events: Rules, Terminology and Technical Specifications},
  author  = {Troester, Matthew M. and Quan, Stuart F. and Berry, Richard B. and others},
  year    = {2023},
  edition = {Version 3},
  publisher = {American Academy of Sleep Medicine},
  address = {Darien, IL}
}

@article{phan_seqsleepnet_2019,
	title = {{SeqSleepNet}: {End}-to-{End} {Hierarchical} {Recurrent} {Neural} {Network} for {Sequence}-to-{Sequence} {Automatic} {Sleep} {Staging}},
	volume = {27},
	copyright = {https://ieeexplore.ieee.org/Xplorehelp/downloads/license-information/IEEE.html},
	issn = {1534-4320, 1558-0210},
	shorttitle = {{SeqSleepNet}},
	url = {https://ieeexplore.ieee.org/document/8631195/},
	doi = {10.1109/TNSRE.2019.2896659},
	language = {en},
	number = {3},
	urldate = {2025-06-10},
	journal = {IEEE Transactions on Neural Systems and Rehabilitation Engineering},
	author = {Phan, Huy and Andreotti, Fernando and Cooray, Navin and Chen, Oliver Y. and De Vos, Maarten},
	month = mar,
	year = {2019},
	pages = {400--410},
}

@ARTICLE{Redline1995-yz_cfs,
  title     = "The familial aggregation of obstructive sleep apnea",
  author    = "Redline, S and Tishler, P V and Tosteson, T D and Williamson, J
               and Kump, K and Browner, I and Ferrette, V and Krejci, P",
  journal   = "Am. J. Respir. Crit. Care Med.",
  publisher = "Oxford University Press (OUP)",
  volume    =  151,
  number    = "3 Pt 1",
  pages     = "682--687",
  month     =  mar,
  year      =  1995,
  copyright = "https://academic.oup.com/pages/standard-publication-reuse-rights",
  language  = "en"
}

@ARTICLE{Zhang2018-sw_nssr,
  title     = "The National Sleep Research Resource: towards a sleep data
               commons",
  author    = "Zhang, Guo-Qiang and Cui, Licong and Mueller, Remo and Tao,
               Shiqiang and Kim, Matthew and Rueschman, Michael and Mariani,
               Sara and Mobley, Daniel and Redline, Susan",
  journal   = "J. Am. Med. Inform. Assoc.",
  publisher = "Oxford University Press (OUP)",
  volume    =  25,
  number    =  10,
  pages     = "1351--1358",
  month     =  oct,
  year      =  2018,
  copyright = "http://creativecommons.org/licenses/by-nc-nd/4.0/",
  language  = "en"
}

@MISC{Perslev2021-cp_DCSM,
  title     = "{DCSM} Sleep Staging Dataset",
  author    = "Perslev, Mathias and Darkner, Sune and Kempfner, Lykke and
               Nikolic, Miki and Jennum, Poul J{\o}rgen and Igel, Christian",
  publisher = "University of Copenhagen",
  year      =  2021
}

@MISC{Thorey2025-ke_DOD,
  title     = "Dreem Open Datasets",
  author    = "Thorey, Valentin",
  publisher = "Zenodo",
  year      =  2025
}

@article{PhysioNet-hmc-sleep-staging-1.1,
  author = {Alvarez-Estevez, Diego and Rijsman, Roselyne},
  title = {{Haaglanden Medisch Centrum sleep staging database}},
  journal = {{PhysioNet}},
  year = {2022},
  month = mar,
  note = {Version 1.1},
  doi = {10.13026/t79q-fr32},
  url = {https://doi.org/10.13026/t79q-fr32}
}

@article{ISRUC,
author = {Khalighi, Sirvan and Sousa, Teresa and Santos, José and Nunes, Urbano},
year = {2015},
month = {11},
pages = {},
title = {ISRUC-Sleep: A comprehensive public dataset for sleep researchers},
volume = {124},
journal = {Computer Methods and Programs in Biomedicine},
doi = {10.1016/j.cmpb.2015.10.013}
}

@ARTICLE{Rosen2012-yy_homepap,
  title     = "A multisite randomized trial of portable sleep studies and
               positive airway pressure autotitration versus laboratory-based
               polysomnography for the diagnosis and treatment of obstructive
               sleep apnea: the {HomePAP} study",
  author    = "Rosen, Carol L and Auckley, Dennis and Benca, Ruth and
               Foldvary-Schaefer, Nancy and Iber, Conrad and Kapur, Vishesh and
               Rueschman, Michael and Zee, Phyllis and Redline, Susan",
  journal   = "Sleep",
  publisher = "Oxford University Press (OUP)",
  volume    =  35,
  number    =  6,
  pages     = "757--767",
  month     =  jun,
  year      =  2012,
  language  = "en"
}

@article{mass,
author = {O'Reilly, Christian and Gosselin, Nadia and Carrier, Julie and Nielsen, Tore},
title = {Montreal Archive of Sleep Studies: an open-access resource for instrument benchmarking and exploratory research},
journal = {Journal of Sleep Research},
volume = {23},
number = {6},
pages = {628-635},
doi = {https://doi.org/10.1111/jsr.12169},
url = {https://onlinelibrary.wiley.com/doi/abs/10.1111/jsr.12169},
eprint = {https://onlinelibrary.wiley.com/doi/pdf/10.1111/jsr.12169},
year = {2014}
}

@ARTICLE{PaperDOD,
  author={Guillot, Antoine and Sauvet, Fabien and During, Emmanuel H. and Thorey, Valentin},
  journal={IEEE Transactions on Neural Systems and Rehabilitation Engineering}, 
  title={Dreem Open Datasets: Multi-Scored Sleep Datasets to Compare Human and Automated Sleep Staging}, 
  year={2020},
  volume={28},
  number={9},
  pages={1955-1965},
  doi={10.1109/TNSRE.2020.3011181}}

@ARTICLE{Chen2015-gq_mesa,
  title     = "Racial/ethnic differences in sleep disturbances: The
               {Multi-Ethnic} Study of atherosclerosis ({MESA})",
  author    = "Chen, Xiaoli and Wang, Rui and Zee, Phyllis and Lutsey, Pamela L
               and Javaheri, Sogol and Alc{\'a}ntara, Carmela and Jackson,
               Chandra L and Williams, Michelle A and Redline, Susan",
  journal   = "Sleep",
  publisher = "Oxford University Press (OUP)",
  volume    =  38,
  number    =  6,
  pages     = "877--888",
  month     =  jun,
  year      =  2015,
  language  = "en"
}

@ARTICLE{Blackwell2011-ca_mros,
  title     = "Associations between sleep architecture and sleep-disordered
               breathing and cognition in older community-dwelling men: the
               Osteoporotic Fractures in Men Sleep Study",
  author    = "Blackwell, Terri and Yaffe, Kristine and Ancoli-Israel, Sonia
               and Redline, Susan and Ensrud, Kristine E and Stefanick, Marcia
               L and Laffan, Alison and Stone, Katie L and {Osteoporotic
               Fractures in Men Study Group}",
  journal   = "J. Am. Geriatr. Soc.",
  publisher = "Wiley",
  volume    =  59,
  number    =  12,
  pages     = "2217--2225",
  month     =  dec,
  year      =  2011,
  copyright = "http://onlinelibrary.wiley.com/termsAndConditions\#vor",
  language  = "en"
}

@MISC{Westover2023-ca_physionet,
  title     = "The Human Sleep Project",
  author    = "Westover, M Brandon and Moura Junior, Valdery and Thomas, Robert
               and Cash, Sydney and Nasiri, Samaneh and Sun, Haoqi and Gupta,
               Aditya and Rosand, Jonathan and Ghanta, Manohar and Ganglberger,
               Wolfgang and Katwa, Umakanth and Stone, Katie and Zhang, Zhiyong
               and Ganjoo, Gauri and Nassi PhD Candidate, Thijs E and Wei,
               Ruoqi and Hwang, Dennis and Trotti, Lynn Marie and Parekh, Ankit
               and Meulenbrugge, Erikjan and Mignot, Emmanuel and Au, Rhoda and
               Clifford, Gari and Rapoport, David",
  publisher = "BDSP",
  year      =  2023
}

@MISC{Kemp2018-bk_sleepedf,
  title     = "The {sleep-EDF} database [expanded]",
  author    = "Kemp, Bastiaan and Zwinderman, Aeilko and Tuk, Bert and
               Kamphuisen, Hilbert and Obery{\'e}, Josefien",
  publisher = "physionet.org",
  year      =  2018
}

@article{Quan1997SleepHeartHealth,
  author  = {Quan, S. F. and Howard, B. V. and Iber, C. and Kiley, J. P. and Nieto, F. J. and O'Connor, G. T. and Rapoport, D. M. and Redline, S. and Robbins, J. and Samet, J. M. and Wahl, P. W.},
  title   = {The Sleep Heart Health Study: design, rationale, and methods},
  journal = {Sleep},
  year    = {1997},
  volume  = {20},
  number  = {12},
  pages   = {1077--1085},
  month   = dec,
  pmid    = {9493915}
}

@MISC{McNicholas2004-dq_ucddb,
  title     = "St. Vincent's university hospital / university college Dublin
               sleep apnea database",
  author    = "McNicholas, Walter and Doherty, Liam and Ryan, Silke and Garvey,
               John and Boyle, Patricia and Chua, Eric",
  publisher = "physionet.org",
  year      =  2004
}

@article{Young2009WisconsinSleepCohort,
  author  = {Young, T. and Palta, M. and Dempsey, J. and Peppard, P. E. and Nieto, F. J. and Hla, K. M.},
  title   = {Burden of sleep apnea: rationale, design, and major findings of the Wisconsin Sleep Cohort study},
  journal = {WMJ},
  year    = {2009},
  volume  = {108},
  number  = {5},
  pages   = {246--249},
  month   = aug,
  pmid    = {19743755},
  pmcid   = {PMC2858234}
}

@ARTICLE{heremans_wearable_2024,
  title     = "Wearable sleep recording augmented by artificial intelligence
               for Alzheimer's disease screening",
  author    = "Heremans, Elisabeth R M and Devulder, Astrid and Borz{\'e}e,
               Pascal and Vandenberghe, Rik and De Winter, Fran{\c
               c}ois-Laurent and Vandenbulcke, Mathieu and Van Den Bossche,
               Maarten and Buyse, Bertien and Testelmans, Dries and Van
               Paesschen, Wim and De Vos, Maarten",
  journal   = "NPJ Aging",
  publisher = "Springer Science and Business Media LLC",
  volume    =  11,
  number    =  1,
  pages     = "34",
  month     =  may,
  year      =  2025,
  copyright = "https://creativecommons.org/licenses/by-nc-nd/4.0",
  language  = "en"
}

@article{SUN_age_changes,
title = {The sleep and wake electroencephalogram over the lifespan},
journal = {Neurobiology of Aging},
volume = {124},
pages = {60-70},
year = {2023},
issn = {0197-4580},
doi = {https://doi.org/10.1016/j.neurobiolaging.2023.01.006},
url = {https://www.sciencedirect.com/science/article/pii/S0197458023000076},
author = {Haoqi Sun and Elissa Ye and Luis Paixao and Wolfgang Ganglberger and Catherine J. Chu and Can Zhang and Jonathan Rosand and Emmanuel Mignot and Sydney S. Cash and David Gozal and Robert J. Thomas and M. Brandon Westover}
}

@article{Sun2026SleepEEGBrainAgeDementiaRisk,
  author  = {Sun, Haoqi and Milton, Sarah and Fang, Yi and others},
  title   = {Machine Learning--Based Sleep Electroencephalographic Brain Age Index and Dementia Risk: An Individual Participant Data Meta-Analysis},
  journal = {JAMA Network Open},
  year    = {2026},
  volume  = {9},
  number  = {3},
  pages   = {e261521},
  doi     = {10.1001/jamanetworkopen.2026.1521}
}

@article{sun_brain_2019,
	title = {Brain age from the electroencephalogram of sleep},
	volume = {74},
	issn = {1558-1497},
	doi = {10.1016/j.neurobiolaging.2018.10.016},
	language = {eng},
	journal = {Neurobiology of Aging},
	author = {Sun, Haoqi and Paixao, Luis and Oliva, Jefferson T. and Goparaju, Balaji and Carvalho, Diego Z. and van Leeuwen, Kicky G. and Akeju, Oluwaseun and Thomas, Robert J. and Cash, Sydney S. and Bianchi, Matt T. and Westover, M. Brandon},
	month = feb,
	year = {2019},
	pages = {112--120},
}

@article{ENGEMANN2022119521,
title = {A reusable benchmark of brain-age prediction from M/EEG resting-state signals},
journal = {NeuroImage},
volume = {262},
pages = {119521},
year = {2022},
issn = {1053-8119},
doi = {https://doi.org/10.1016/j.neuroimage.2022.119521},
url = {https://www.sciencedirect.com/science/article/pii/S105381192200636X},
author = {Denis A. Engemann and Apolline Mellot and Richard Höchenberger and Hubert Banville and David Sabbagh and Lukas Gemein and Tonio Ball and Alexandre Gramfort}
}

@article{phan_sleeptransformer_2022,
	title = {{SleepTransformer}: {Automatic} {Sleep} {Staging} with {Interpretability} and {Uncertainty} {Quantification}},
	volume = {69},
	issn = {0018-9294, 1558-2531},
	shorttitle = {{SleepTransformer}},
	url = {http://arxiv.org/abs/2105.11043},
	doi = {10.1109/TBME.2022.3147187},
	language = {en},
	number = {8},
	urldate = {2025-08-04},
	journal = {IEEE Transactions on Biomedical Engineering},
	author = {Phan, Huy and Mikkelsen, Kaare and Chén, Oliver Y. and Koch, Philipp and Mertins, Alfred and Vos, Maarten De},
	month = aug,
	year = {2022},
	note = {arXiv:2105.11043 [cs]},
	pages = {2456--2467},
}

@article{lee_sleepyco_2024,
	title = {{SleePyCo}: {Automatic} sleep scoring with feature pyramid and contrastive learning},
	volume = {240},
	issn = {0957-4174},
	url = {https://www.sciencedirect.com/science/article/pii/S0957417423030531},
	doi = {https://doi.org/10.1016/j.eswa.2023.122551},
	journal = {Expert Systems with Applications},
	author = {Lee, Seongju and Yu, Yeonguk and Back, Seunghyeok and Seo, Hogeon and Lee, Kyoobin},
	year = {2024},
	pages = {122551},
}

@article{avakyan2017ilae,
  title={ILAE Classification of the epilepsies: the 2017 revision and update},
  author={Avakyan, GN and Blinov, DV and Lebedeva, AV and Burd, SG and Avakyan, GG},
  journal={Epilepsy and paroxysmal conditions},
  volume={9},
  number={1},
  pages={6--25},
  year={2017}
}

@inproceedings{radford2021learning,
  title={Learning transferable visual models from natural language supervision},
  author={Radford, Alec and Kim, Jong Wook and Hallacy, Chris and Ramesh, Aditya and Goh, Gabriel and Agarwal, Sandhini and Sastry, Girish and Askell, Amanda and Mishkin, Pamela and Clark, Jack and others},
  booktitle={International conference on machine learning},
  pages={8748--8763},
  year={2021},
  organization={PmLR}
}

@inproceedings{caron2021emerging,
  title={Emerging properties in self-supervised vision transformers},
  author={Caron, Mathilde and Touvron, Hugo and Misra, Ishan and J{\'e}gou, Herv{\'e} and Mairal, Julien and Bojanowski, Piotr and Joulin, Armand},
  booktitle={Proceedings of the IEEE/CVF international conference on computer vision},
  pages={9650--9660},
  year={2021}
}

@ARTICLE{Busch2024-om_cognitive,
  title     = "Automated detection of cognitive impairment in clinical practice",
  author    = "Busch, Robyn M and Hogue, Olivia and Postle, Abagail F and
               Floden, Darlene P",
  journal   = "J. Neurol.",
  publisher = "Springer Science and Business Media LLC",
  volume    =  271,
  number    =  8,
  pages     = "5187--5196",
  month     =  aug,
  year      =  2024,
  copyright = "https://creativecommons.org/licenses/by/4.0",
  language  = "en"
}

@article{barateau_narcolepsy_2022,
	title = {Narcolepsy},
	volume = {31},
	issn = {0962-1105, 1365-2869},
	url = {https://onlinelibrary.wiley.com/doi/10.1111/jsr.13631},
	doi = {10.1111/jsr.13631},
	language = {en},
	number = {4},
	urldate = {2025-06-20},
	journal = {Journal of Sleep Research},
	author = {Barateau, Lucie and Pizza, Fabio and Plazzi, Giuseppe and Dauvilliers, Yves},
	month = aug,
	year = {2022},
	pages = {e13631},
}

@article{andrzejak2001indications,
  title={Indications of nonlinear deterministic and finite-dimensional structures in time series of brain electrical activity: Dependence on recording region and brain state},
  author={Andrzejak, Ralph G and Lehnertz, Klaus and Mormann, Florian and Rieke, Christoph and David, Peter and Elger, Christian E},
  journal={Physical Review E},
  volume={64},
  number={6},
  pages={061907},
  year={2001},
  publisher={APS}
}

@article{khan2022nmt,
  title={The NMT scalp EEG dataset: An open-source annotated dataset of healthy and pathological EEG recordings for predictive modeling},
  author={Khan, Hassan Aqeel and Ul Ain, Rahat and Kamboh, Awais Mehmood and Butt, Hammad Tanveer and Shafait, Saima and Alamgir, Wasim and Stricker, Didier and Shafait, Faisal},
  journal={Frontiers in neuroscience},
  volume={15},
  pages={755817},
  year={2022},
  publisher={Frontiers Media SA}
}

@article{obeid2016temple,
  title={The temple university hospital EEG data corpus},
  author={Obeid, Iyad and Picone, Joseph},
  journal={Frontiers in neuroscience},
  volume={10},
  pages={196},
  year={2016},
  publisher={Frontiers Media SA}
}

@article{bhagubai2024towards,
  title={Towards automated seizure detection with wearable eeg--grand challenge},
  author={Bhagubai, Miguel and Swinnen, Lauren and Cleeren, Evy and Van Paesschen, Wim and De Vos, Maarten and Chatzichristos, Christos},
  journal={IEEE Open Journal of Signal Processing},
  volume={5},
  pages={717--724},
  year={2024},
  publisher={IEEE}
}

@article{jirsaraie2023multimodalbrainage,
  author  = {Jirsaraie, Robert J. and Gorelik, Alexander J. and Gatavins, M. M. and Engemann, Denis A. and Bogdan, Ryan and Barch, Deanna M. and Sotiras, Aristeidis},
  title   = {A systematic review of multimodal brain age studies: Uncovering a divergence between model accuracy and utility},
  journal = {Patterns},
  year    = {2023},
  volume  = {4},
  number  = {4},
  pages   = {100712},
  doi     = {10.1016/j.patter.2023.100712},
  pmid    = {37123443},
  pmcid   = {PMC10140612}
}

@article{dan2024szcore,
author = {Dan, Jonathan and Pale, Una and Amirshahi, Alireza and Cappelletti, William and Ingolfsson, Thorir Mar and Wang, Xiaying and Cossettini, Andrea and Bernini, Adriano and Benini, Luca and Beniczky, Sándor and Atienza, David and Ryvlin, Philippe},
title = {SzCORE: Seizure Community Open-Source Research Evaluation framework for the validation of electroencephalography-based automated seizure detection algorithms},
journal = {Epilepsia},
volume = {66},
number = {S3},
pages = {14-24},
doi = {https://doi.org/10.1111/epi.18113},
url = {https://onlinelibrary.wiley.com/doi/abs/10.1111/epi.18113},
eprint = {https://onlinelibrary.wiley.com/doi/pdf/10.1111/epi.18113},
year = {2024}
}

@article{wu2025large,
  title={Large EEG-U-Transformer for Time-Step Level Detection Without Pre-Training},
  author={Wu, Kerui and Zhao, Ziyue and Yener, B{\"u}lent},
  journal={arXiv preprint arXiv:2504.00336},
  year={2025}
}

@article{bhagubai2025seizeit2,
  title={SeizeIT2: Wearable Dataset Of Patients With Focal Epilepsy},
  author={Bhagubai, Miguel and Chatzichristos, Christos and Swinnen, Lauren and Macea, Jaiver and Zhang, Jingwei and Lagae, Lieven and Jansen, Katrien and Schulze-Bonhage, Andreas and Sales, Francisco and Mahler, Benno and others},
  journal={Scientific Data},
  volume={12},
  number={1},
  pages={1228},
  year={2025},
  publisher={Nature Publishing Group UK London}
}

@Article{sz1,
AUTHOR = {Bhagubai, Miguel and Vandecasteele, Kaat and Swinnen, Lauren and Macea, Jaiver and Chatzichristos, Christos and De Vos, Maarten and Van Paesschen, Wim},
TITLE = {The Power of ECG in Semi-Automated Seizure Detection in Addition to Two-Channel behind-the-Ear EEG},
JOURNAL = {Bioengineering},
VOLUME = {10},
YEAR = {2023},
NUMBER = {4},
ARTICLE-NUMBER = {491},
URL = {https://www.mdpi.com/2306-5354/10/4/491},
PubMedID = {37106678},
ISSN = {2306-5354},
DOI = {10.3390/bioengineering10040491}
}

@article{tangermann2012review,
  title={Review of the BCI competition IV},
  author={Tangermann, Michael and M{\"u}ller, Klaus-Robert and Aertsen, Ad and Birbaumer, Niels and Braun, Christoph and Brunner, Clemens and Leeb, Robert and Mehring, Carsten and Miller, Kai J and M{\"u}ller-Putz, Gernot R and others},
  journal={Frontiers in neuroscience},
  volume={6},
  pages={55},
  year={2012},
  publisher={Frontiers Research Foundation}
}

@article{detti2020siena,
  author = {Detti, Paolo},
  title = {{Siena Scalp EEG Database}},
  journal = {{PhysioNet}},
  year = {2020},
  month = aug,
  note = {Version 1.0.0},
  doi = {10.13026/5d4a-j060},
  url = {https://doi.org/10.13026/5d4a-j060}
}

@article{faller2012autocalibration,
  title={Autocalibration and recurrent adaptation: Towards a plug and play online ERD-BCI},
  author={Faller, Josef and Vidaurre, Carmen and Solis-Escalante, Teodoro and Neuper, Christa and Scherer, Reinhold},
  journal={IEEE Transactions on Neural Systems and Rehabilitation Engineering},
  volume={20},
  number={3},
  pages={313--319},
  year={2012},
  publisher={IEEE}
}

@article{yi2014evaluation,
  title={Evaluation of EEG oscillatory patterns and cognitive process during simple and compound limb motor imagery},
  author={Yi, Weibo and Qiu, Shuang and Wang, Kun and Qi, Hongzhi and Zhang, Lixin and Zhou, Peng and He, Feng and Ming, Dong},
  journal={PloS one},
  volume={9},
  number={12},
  pages={e114853},
  year={2014},
  publisher={Public Library of Science San Francisco, USA}
}

@article{pillette2021experimenters,
  title={Experimenters' influence on mental-imagery based brain-computer interface user training},
  author={Pillette, L{\'e}a and Roc, Aline and N’kaoua, Bernard and Lotte, Fabien},
  journal={International Journal of Human-Computer Studies},
  volume={149},
  pages={102603},
  year={2021},
  publisher={Elsevier}
}

@article{shoeb2009chbmit,
  author = {Guttag, John},
  title = {{CHB-MIT Scalp EEG Database}},
  journal = {{PhysioNet}},
  year = {2010},
  month = jun,
  note = {Version 1.0.0},
  doi = {10.13026/C2K01R},
  url = {https://doi.org/10.13026/C2K01R}
}

@article{stevenson2019helsinki,
  title={A dataset of neonatal EEG recordings with seizure annotations},
  author={Stevenson, Nathan J and Tapani, Karoliina and Lauronen, Leena and Vanhatalo, Sampsa},
  journal={Scientific data},
  volume={6},
  number={1},
  pages={190039},
  year={2019},
  publisher={Nature Publishing Group}
}

@article{shah2018tusz,
  title={The temple university hospital seizure detection corpus},
  author={Shah, Vinit and Von Weltin, Eva and Lopez, Silvia and McHugh, James Riley and Veloso, Lillian and Golmohammadi, Meysam and Obeid, Iyad and Picone, Joseph},
  journal={Frontiers in neuroinformatics},
  volume={12},
  pages={83},
  year={2018},
  publisher={Frontiers Media SA}
}

@article{shin2016open,
  title={Open access dataset for EEG+ NIRS single-trial classification},
  author={Shin, Jaeyoung and von L{\"u}hmann, Alexander and Blankertz, Benjamin and Kim, Do-Won and Jeong, Jichai and Hwang, Han-Jeong and M{\"u}ller, Klaus-Robert},
  journal={IEEE Transactions on Neural Systems and Rehabilitation Engineering},
  volume={25},
  number={10},
  pages={1735--1745},
  year={2016},
  publisher={IEEE}
}

@article{liu2024eeg,
  title={An EEG motor imagery dataset for brain computer interface in acute stroke patients},
  author={Liu, Haijie and Wei, Penghu and Wang, Haochong and Lv, Xiaodong and Duan, Wei and Li, Meijie and Zhao, Yan and Wang, Qingmei and Chen, Xinyuan and Shi, Gaige and others},
  journal={Scientific Data},
  volume={11},
  number={1},
  pages={131},
  year={2024},
  publisher={Nature Publishing Group UK London}
}

@article{vaineau2019brain,
  title={Brain invaders adaptive versus non-adaptive P300 brain-computer interface dataset},
  author={Vaineau, Erwan and Barachant, Alexandre and Andreev, Anton and Rodrigues, Pedro C and Cattan, Gr{\'e}goire and Congedo, Marco},
  journal={arXiv preprint arXiv:1904.09111},
  year={2019}
}

@phdthesis{korczowski2019brain,
  title={Brain Invaders calibration-less P300-based BCI using dry EEG electrodes Dataset (bi2014a)},
  author={Korczowski, Louis and Ostaschenko, Ekaterina and Andreev, Anton and Cattan, Gr{\'e}goire and Rodrigues, Pedro Luiz Coelho and Gautheret, Violette and Congedo, Marco},
  year={2019},
  school={GIPSA-lab}
}

@article{hoffmann2008efficient,
  title={An efficient P300-based brain--computer interface for disabled subjects},
  author={Hoffmann, Ulrich and Vesin, Jean-Marc and Ebrahimi, Touradj and Diserens, Karin},
  journal={Journal of Neuroscience methods},
  volume={167},
  number={1},
  pages={115--125},
  year={2008},
  publisher={Elsevier}
}

@article{riccio2013attention,
  title={Attention and P300-based BCI performance in people with amyotrophic lateral sclerosis},
  author={Riccio, Angela and Simione, Luca and Schettini, Francesca and Pizzimenti, Alessia and Inghilleri, Maurizio and Belardinelli, Marta Olivetti and Mattia, Donatella and Cincotti, Febo},
  journal={Frontiers in human neuroscience},
  volume={7},
  pages={732},
  year={2013},
  publisher={Frontiers Media SA}
}

@article{kappenman2021erp,
  title={ERP CORE: An open resource for human event-related potential research},
  author={Kappenman, Emily S and Farrens, Jaclyn L and Zhang, Wendy and Stewart, Andrew X and Luck, Steven J},
  journal={NeuroImage},
  volume={225},
  pages={117465},
  year={2021},
  publisher={Elsevier}
}

@article{nakanishi2015comparison,
  title={A comparison study of canonical correlation analysis based methods for detecting steady-state visual evoked potentials},
  author={Nakanishi, Masaki and Wang, Yijun and Wang, Yu-Te and Jung, Tzyy-Ping},
  journal={PloS one},
  volume={10},
  number={10},
  pages={e0140703},
  year={2015},
  publisher={Public Library of Science San Francisco, CA USA}
}

@article{kim202540,
  title={A 40-Class SSVEP Speller Dataset: Beta Range Stimulation for Low-Fatigue BCI Applications},
  author={Kim, Heegyu and Won, Kyungho and Ahn, Minkyu and Jun, Sung Chan},
  journal={Scientific Data},
  volume={12},
  number={1},
  pages={1751},
  year={2025},
  publisher={Nature Publishing Group UK London}
}

@article{zyma2019electroencephalograms,
  title={Electroencephalograms during mental arithmetic task performance},
  author={Zyma, Igor and Tukaev, Sergii and Seleznov, Ivan and Kiyono, Ken and Popov, Anton and Chernykh, Mariia and Shpenkov, Oleksii},
  journal={Data},
  volume={4},
  number={1},
  pages={14},
  year={2019},
  publisher={MDPI}
}

@article{rodriguez2019tracking,
  title={Tracking transient changes in the neural frequency architecture: harmonic relationships between theta and alpha peaks facilitate cognitive performance},
  author={Rodriguez-Larios, Julio and Alaerts, Kaat},
  journal={Journal of Neuroscience},
  volume={39},
  number={32},
  pages={6291--6298},
  year={2019},
  publisher={Society for Neuroscience}
}

@book{tuab,
  title={Automated interpretation of abnormal adult electroencephalograms},
  author={de Diego, Silvia L{\'o}pez},
  year={2017},
  publisher={Temple University}
}

@article{klatt2012epilepsiae,
author = {Klatt, Juliane and Feldwisch-Drentrup, Hinnerk and Ihle, Matthias and Navarro, Vincent and Neufang, Markus and Teixeira, Cesar and Adam, Claude and Valderrama, Mario and Alvarado-Rojas, Catalina and Witon, Adrien and Le Van Quyen, Michel and Sales, Francisco and Dourado, Antonio and Timmer, Jens and Schulze-Bonhage, Andreas and Schelter, Bjoern},
title = {The EPILEPSIAE database: An extensive electroencephalography database of epilepsy patients},
journal = {Epilepsia},
volume = {53},
number = {9},
pages = {1669-1676},
doi = {https://doi.org/10.1111/j.1528-1167.2012.03564.x},
url = {https://onlinelibrary.wiley.com/doi/abs/10.1111/j.1528-1167.2012.03564.x},
eprint = {https://onlinelibrary.wiley.com/doi/pdf/10.1111/j.1528-1167.2012.03564.x},
year = {2012}
}

@article{yuan2024brainwave,
  title={Brainwave: A brain signal foundation model for clinical applications},
  author={Yuan, Zhizhang and Shen, Fanqi and Li, Meng and Yu, Yuguo and Tan, Chenhao and Yang, Yang},
  journal={arXiv preprint arXiv:2402.10251},
  year={2024}
}
}

\newpage
\appendix
\vspace{2.5em}
{\Large\bfseries \centering Supplementary Material\par}
\vspace{2.5em}

\section{Model Description}
\label{Appendix:Model-Description}

\subsection{Input-space specifications}
\label{Appendix:Model-Description--Input-Space-Assumptions}
EEG-FM substantially vary in the specific way they handle EEG signals as input. Axes of variation include sampling rate, normalization, re-referencing, flexibility in number of channels or time windows, and a fixed set of allowed channels versus interpolating spherical channel coordinates. For the models evaluated in \textsc{NeuroAtlas}, we describe the input specifications under which they were designed in Table \ref{tab:neuroatlas-included-models}.

\subsection{Models Included in \textsc{NeuroAtlas}}
\label{Appendix:Model-Description--Included-Models}

\textsc{NeuroAtlas} evaluates models from three complementary groups: supervised EEG models, EEG foundation models, and general time-series foundation models. The goal is to compare task-specific supervised representations, EEG-oriented pretrained representations, and generic temporal representations under a unified frozen-encoder evaluation protocol.

\paragraph{EEG Foundation Models}
We evaluate the following EEG foundation models (EEG-FMs): BIOT~\citep{yang2023biot}, CBraMod~\citep{wang2024cbramod}, EEGPT~\citep{wang2024eegpt}, LaBraM~\citep{jiang2024large}, NeuroGPT~\citep{cui2024neuro}, NeuroLM~\citep{jiang2024neurolm}, NeuroRVQ~\citep{barmpas2025neurorvq}, REVE~\citep{ouahidi2025reve}, SleepFM~\citep{thapa2024sleepfm}, ST-EEGFormer-S/B/L~\citep{yang2026eeg}, and MIRepNet~\citep{liu2025mirepnet}. These models are pretrained on EEG or neurophysiological data using objectives such as masked reconstruction, contrastive learning, autoregressive modeling, discrete token prediction, or multimodal alignment. EEG-FMs differ in preprocessing requirements and channel configurations, summarized in Table \ref{tab:neuroatlas-included-models}.

\paragraph{General Timeseries Foundation Models}
We evaluate the following timeseries foundation models (TS-FMs): Chronos-T/S/B/L~\citep{ansari2024chronos}, Moirai-S/B/L~\citep{woo2024unified}, and MOMENT-S/B/L~\citep{goswami2024moment}. These models are pretrained on heterogeneous non-EEG temporal datasets and are designed to capture generic timeseries features. Chronos models univariate time series through tokenized numerical sequences. MOMENT uses patch-based temporal representations with instance normalization and supports several downstream modes such as forecasting, classification, imputation, and representation extraction. Moirai supports multivariate forecasting through any-variate sequence modeling. Unlike EEG-FMs, these models do not explicitly encode electrode topology, montage geometry, or EEG-specific referencing assumptions, detailed specifications can be found in Table~\ref{tab:neuroatlas-included-models}. They therefore provide a domain-agnostic baseline for testing whether generic time-series pretraining can transfer to EEG tasks when used as a frozen feature extractor.

\paragraph{Supervised EEG Models} 
We evaluate supervised pretrained (Supervised-Pre) from all domains: we use epilepsy-trained Seizure-TF~\citep{wu2025large} and DeepSOZ-HEM (an extension of DeepSOZ~\citep{m2023deepsoz}) and sleep-trained SleepTransformer~\citep{phan_sleeptransformer_2022}, SleePyCo~\citep{lee_sleepyco_2024}, and CoRe-Sleep~\citep{kontras2024core}. On brain age, we evaluate the sleep EEG Brain Age feature-based model described in Sun et al. ~\citep{sun2019brain}. On BCI, we evaluate EEGNet~\citep{lawhern2018eegnet}.
These models are pretrained to perform a specific task and their input assumptions are often closely tied to the target task. For example, EEGNet is used in the BCI setting, where channel selection and spatial filtering are central to motor imagery decoding. 
As such, these models provide task- or domain-specific reference baselines for evaluating FMs. Model specifications are described in Table~\ref{tab:neuroatlas-included-models} and further discussed per domain below.

\paragraph{Supervised Epilepsy Models}
Seizure-TF is an encoder-decoder model used for seizure detection, sleep stage classification and pathological detection. The encoder consists of convolutional layers, followed by a scaling-embedding stack of residual CNN layers and transformer layers. The selected checkpoint was trained on the datasets TUSZ~\citep{shah2018tusz} and Siena~\citep{detti2020siena}, using windows of 60 seconds. At inference, a sliding window of 10 seconds is passed through the model (encoder-only) with no overlap (stride of 10 seconds). The output of the scaling-embedding stack comprises the embeddings. 

DeepSOZ-HEM is an extension of DeepSOZ, a Transformer+LSTM model used for seizure detection and seizure-onset-zone (SOZ) localization. The selected checkpoint was trained on the datasets TUSZ~\citep{shah2018tusz} and Siena~\citep{detti2020siena}, using windows of 10 minutes (600 seconds). At inference, a sliding window of 10 seconds is passed through the model with no overlap (stride of 10 seconds). We extract the embeddings from the Transformer global token, averaged over all seconds in the window.

\paragraph{Supervised Sleep Models}
SleepTransformer is a sequence-to-sequence transformer-based sleep staging model using single-channel EEG, trained on SHHS \cite{phan_sleeptransformer_2022}. Each 30-second epoch is converted to a short-time Fourier transform (STFT) representation. At inference, a sliding window of sequence length 21, i.e., containing 21 consecutive epochs, is passed through the model with a stride of 1 epoch. We extract the embedding from the CLS token output of the outer transformer at the center epoch (index 10 of 21).

SleePyCo is a sleep staging model using single-channel EEG, trained using supervised contrastive learning \cite{lee_sleepyco_2024}. We evaluate the published checkpoint trained on SHHS. During inference, a sliding sequence of 10 consecutive epochs is fed  with a stride of 1 epoch. Extracted embeddings are the attention-pooled transformer output, averaged across the three feature-pyramid scales, attributed to the last epoch of the sequence.

CoRe-Sleep is a transformer-based bimodal sleep staging model trained on EEG and EOG, trained on SHHS \cite{kontras2024core}. At inference, only the EEG channel is used, and each epoch is converted using STFT. Embeddings are extracted using non-overlapping sequences of length 21 epochs, as the summed cross-attention fused representation of the outer Transformer outputs, retained for all epochs in the window.

Thus, in contrast to the evaluated FM embeddings, the Supervised-Pre models explicitly encode inter-epoch temporal context across sequences of 30-second epochs.

\paragraph{Supervised Brain Aging Model}
We additionally include the sleep-EEG brain-age model of Sun et al. as a task-specific supervised reference \cite{sun_brain_2019}. In contrast to the frozen FM probes, this is a feature-based model trained to predict chronological age from sleep EEG features and is therefore directly optimized for brain-age estimation. We use it as an upper reference for the brain-age prediction and BAG analyses, allowing us to compare general-purpose EEG and time-series representations against a model designed specifically for sleep-EEG aging.

 \paragraph{Supervised BCI/MI Model} For the supervised EEGNet baseline, we used publicly available pretrained EEGNetv4 checkpoints released on Hugging Face by Guetschel et al.~\citep{GuePapTan23a}, available at \url{https://huggingface.co/PierreGtch/EEGNetv4}. These pretrained models were only available for the Motor Imagery (MI) paradigm, which is why EEGNet is exclusively included in the MI comparison. Following the same protocol used for EEG foundation models and time-series foundation models, we froze the pretrained encoder, extracted latent embeddings, and evaluated them using linear probing under a leave-one-subject-out (LOSO) setting. 



\begingroup
\fontsize{5.9}{6.1}\selectfont
\setlength{\tabcolsep}{1.15pt}
\renewcommand{\arraystretch}{0.82}
\begin{longtable}{P{1.65cm}|P{2.05cm} P{1.60cm} P{2.35cm} P{1.75cm} P{2.20cm} P{2.45cm}}
\caption{ Overview of models included in \textsc{NeuroAtlas} and their input specifications. }
\label{tab:neuroatlas-included-models}
\\
\toprule
\textbf{Model} &
\textbf{Channels} &
\textbf{Sampling rate(s)} &
\textbf{Scaling / Normalization} &
\textbf{Referencing} &
\textbf{Channel ordering} &
\textbf{Strict vs Flexible}
\\
\midrule
\endfirsthead

\toprule
\textbf{Model} &
\textbf{Channels} &
\textbf{Sampling rate(s)} &
\textbf{Scaling / Normalization} &
\textbf{Referencing} &
\textbf{Channel ordering} &
\textbf{Strict vs Flexible}
\\
\midrule
\endhead

\midrule
\multicolumn{7}{r}{Continued on next page}
\\
\endfoot

\bottomrule
\endlastfoot

\textbf{BIOT} 
& Variable; token-based; pretrained on 16--18 bipolar channels, e.g., C3-A2/C4-A1.
& 200 Hz; resampled; FFT-based segment embedding.
& Per-channel / per-sample normalization using the 95\% quantile of absolute amplitude.
& Dataset-dependent; commonly bipolar 10--20 derivations.
& Index-based; order-sensitive; channel identity encoded by learned channel embeddings.
& Flexible; supports variable channels and sequence lengths; missing segments handled at token level.
\\

\textbf{CBraMod} \\
{\scriptsize pretrained; rand}
& 19 EEG; fixed pretraining subset from standard 10--20 channels.
& 200 Hz; 30 s samples; 1 s non-overlapping patches.
& Bandpass 0.3--75 Hz; notch 60 Hz; bad-sample removal; amplitude scaled by 100 $\mu$V to [-1,1].
& Reference-agnostic design.
& Spatial--temporal encoding with asymmetric conditional positional encoding.
& Semi-flexible; fixed pretraining channels with adaptive positional encoding for downstream variation.
\\

\textbf{EEGPT} 
& 58 pretraining channels; name-based embedding space.
& 256 Hz; patch-based input.
& Amplitude normalization to a consistent scale, depending on dataset convention.
& Average reference / global CAR.
& Name-based channel identity; ordering not strictly enforced.
& Flexible; supports subsets of predefined channels.
\\

\textbf{LaBraM} 
& Variable multi-channel EEG; 10--20 system; patch-based segmentation.
& 200 Hz; non-overlapping patch window.
& Bandpass 0.1--75 Hz; notch 50 Hz; amplitude scaled by 0.1 mV to [-1,1].
& Not explicitly fixed; heterogeneous referencing across datasets.
& Spatially aware channel-identity embeddings.
& Flexible; supports heterogeneous EEG montages but requires correct channel-to-embedding mapping.
\\

\textbf{NeuroGPT} 
& 22 EEG channels; fixed extended 10--20 montage.
& 250 Hz; resampled from heterogeneous sources.
& Bandpass 0.5--100 Hz; notch 60 Hz; detrending; z-score per recording.
& Average reference / global CAR.
& Fixed aligned 10--20 ordering.
& Semi-flexible; supports multi-dataset training through strict preprocessing alignment.
\\

\textbf{NeuroLM} 
& Variable multi-channel EEG; typically 4--62 channels; 10--20 system.
& 200 Hz; patch-based tokenizer.
& Bandpass 0.1--75 Hz; notch 50/60 Hz; amplitude normalization; z-score in frequency branch.
& Not explicitly constrained.
& Spatially aware; channel identity tied to electrode position.
& Flexible; supports heterogeneous EEG configurations.
\\

\textbf{NeuroRVQ} 
& Variable multi-dataset EEG; up to 64 channels.
& Unified to 200 Hz.
& Fourier-based representation with log-amplitude scaling.
& Dataset-dependent.
& Learned spatial embedding.
& Flexible; relatively channel-agnostic representation interface.
\\

\textbf{REVE}
& Variable multi-channel EEG; coordinate-based arbitrary layouts.
& 200 Hz; overlapping temporal patches.
& Per-record z-score; clipping at $\pm 15\sigma$; bandpass 0.5--99.5 Hz.
& Recording-dependent; no fixed re-referencing.
& Spatial representation based on 3D electrode coordinates.
& Flexible; supports arbitrary layouts and sequence lengths.
\\

\textbf{SleepFM} 
& Multimodal PSG; EEG, EOG, EMG, ECG, and respiratory channels.
& 256 Hz; resampled.
& Minimal preprocessing; 30 s clips; raw signals mostly preserved.
& Dataset-dependent PSG derivations.
& Modality-wise temporal encoding with temporal fusion.
& Flexible; handles heterogeneous modalities and cross-modal representation tasks.
\\

\textbf{ST-EEGFormer}\\
{\scriptsize S/B/L}
& 142 unique EEG channels during pretraining; downstream channels are dataset-dependent.
& 128 Hz for pretraining; downstream data resampled to model-native rate.
& Bandpass 0.1--64 Hz; notch filtering; per-channel zero-mean/unit-variance standardization.
& Not explicitly fixed.
& Spatial--temporal tokenization with learned spatial and temporal positional embeddings.
& Flexible; designed for heterogeneous EEG datasets and channel configurations.
\\

\textbf{MIRepNet} 
& MI-specific channel template over FC, C, CP, and T regions; implementation uses 45 template channels.
& 250 Hz; resampled.
& Bandpass 8--30 Hz; Euclidean Alignment / whitening.
& Not explicitly specified.
& Template-based spatial alignment using inverse-distance interpolation.
& Flexible; maps heterogeneous MI layouts to a unified motor-imagery template.
\\

\textbf{CoRe-Sleep} 
& Multimodal EEG + EOG; e.g., C4-A1 and L-R.
& 100 Hz; resampled from EEG and EOG recordings.
& Time-frequency representation; EEG 0.3--40 Hz; EOG 0.3--23 Hz; STFT over 30 s segments.
& Dataset-dependent derivations.
& Multi-stream temporal encoding with cross-attention fusion.
& Flexible; robust to missing or noisy modalities.
\\

\textbf{SleePyCo} 
& Single-channel EEG; e.g., Fpz-Cz, C4-A1, C3-A2.
& 100 Hz; 30 s epochs.
& Minimal preprocessing except downsampling; raw signal input; internal BatchNorm.
& Dataset-dependent sleep EEG derivations.
& Single-channel temporal ordering.
& Flexible across single-channel sleep datasets; not designed for arbitrary multi-channel montages.
\\

\textbf{DeepSOZ-HEM} 
& 19 EEG channels; fixed 10--20 configuration.
& 256 Hz input; resampled to 200 Hz.
& Bandpass 1.6--30 Hz; clipping at 2$\sigma$.
& Dataset-dependent / not specified.
& Fixed channel positional encoding.
& Strict; requires fixed channel configuration.
\\

\textbf{Seizure-TF} 
& RNS-derived A+B pattern features from two RNS channels.
& Daily time series; no raw EEG sampling rate.
& Per-patient z-score normalization of A+B patterns.
& Not applicable; structured RNS biomarkers rather than raw scalp EEG.
& Temporal patch ordering with learnable positional encoding.
& Flexible for structured RNS features; not designed for raw scalp EEG layouts.
\\

\textbf{SleepTransformer} 
& C4-A1 for SHHS, Fpz-Cz for SleepEDF; code also supports EEG/EOG/EMG.
& 100 Hz.
& STFT time-frequency input; log-amplitude spectrum; zero-mean/unit-variance normalization.
& Dataset-defined derivations.
& Temporal ordering with positional encoding.
& Flexible in modality input; mainly evaluated on single-channel sleep EEG.
\\

\textbf{EEGNet} 
& Multi-channel EEG; channel count depends on dataset.
& Typically resampled according to task setting.
& No fixed preprocessing; BatchNorm used internally.
& Dataset-dependent.
& Spatial structure learned implicitly through convolution.
& Flexible; can adapt to different channel counts and paradigms.
\\

\textbf{Brain Age (Sun et al.} 
& 6-channel EEG montage, e.g., F3, F4, C3, C4, O1, O2; also 2-channel variant.
& 200 Hz.
& Sleep-EEG feature normalization; commonly includes robust per-channel scaling and z-scored features.
& Contralateral mastoid reference, e.g., F3-M2.
& Fixed PSG montage.
& Strict; channel-specific model.
\\

\textbf{Chronos}\\
{\scriptsize T/S/B/L} 
& Univariate time series; no channel structure.
& Dataset-dependent; original sampling generally preserved.
& Mean scaling followed by value quantization into discrete tokens.
& Not applicable.
& Temporal sequence ordering.
& Flexible for arbitrary-length univariate sequences; not topology-aware.
\\

\textbf{Moirai} \\
{\scriptsize S/B/L}
& Variable multivariate series; arbitrary numbers of variables.
& Variable; supports multi-frequency time series.
& Instance normalization across heterogeneous datasets.
& Not applicable.
& Flattened temporal-variate ordering with any-variate attention.
& Flexible; universal forecasting model across domains, frequencies, and variable dimensions.
\\

\textbf{MOMENT} \\
{\scriptsize S/B/L}
& Univariate input; multivariate handled through channel-wise processing.
& Variable; fixed input length after resampling or padding.
& Reversible Instance Normalization.
& Not explicitly constrained.
& Patch-based temporal tokenization.
& Flexible; supports forecasting, classification, imputation, anomaly detection, and representation extraction.
\\

\end{longtable}
\endgroup

\section{Dataset Description}
\label{Appendix:Dataset-Description}

\subsection{Epilepsy}

The seven continuously-labeled cohorts span neonatal (Helsinki~\citep{stevenson2019helsinki}), pediatric (CHB-MIT~\citep{shoeb2009chbmit}), adult seizure monitoring (Siena~\citep{detti2020siena}, SeizeIT1~\citep{sz1}), and large-scale clinical archives like TUSZ~\citep{shah2018tusz}, Epilepsiae~\citep{klatt2012epilepsiae} and SeizeIT2~\citep{bhagubai2025seizeit2}. Three additional cohorts (TUAB~\citep{tuab}, NMT~\citep{khan2022nmt}, Bonn~\citep{andrzejak2001indications}) provide recording-level abnormality labels rather than continuous seizure annotations and contribute $\sim$2{,}500 abnormal recordings to the recording-level evaluation only; of these, only Bonn's 200 ictal clips carry explicit seizure annotation, while TUAB and NMT label broader pathology including epileptiform activity, slowing, and artifacts.
We evaluate on splits that are patient-level throughout.

\subsection{Sleep}
The 15 cohorts evaluated for sleep span lab and clinical polysomnography (PSG) recordings (Cleveland Family Study (CFS) \cite{Redline1995-yz_cfs, Zhang2018-sw_nssr}, DCSM Sleep Staging Dataset \cite{Perslev2021-cp_DCSM}, Dreem Open Datasets (DOD) \cite{Thorey2025-ke_DOD, PaperDOD}, Haaglanden Medisch Centrum (HMC) sleep staging database \cite{PhysioNet-hmc-sleep-staging-1.1}, HomePAP \cite{Rosen2012-yy_homepap, Zhang2018-sw_nssr}, ISRUC-Sleep \cite{ISRUC}, Montréal Archive of Sleep Studies (MASS) \cite{mass}, George B. Moody PhysioNet Challenge 2026 (PN2026), a subset of the Human Sleep Project \cite{Westover2023-ca_physionet}, Sleep-EDF Database Expanded \cite{Kemp2018-bk_sleepedf}, Stanford Technology Analytics and Genomics in Sleep (STAGES) \cite{Zhang2018-sw_nssr}, St. Vincent's University Hospital / University College Dublin Sleep Apnea Database (UCDDB), \cite{McNicholas2004-dq_ucddb}, Wisconsin Sleep Cohort
(WSC) \cite{Zhang2018-sw_nssr, Young2009WisconsinSleepCohort}), 
and at-home PSG recordings (Multi-Ethnic Study of Atherosclerosis (MESA) \cite{Chen2015-gq_mesa, Zhang2018-sw_nssr}, Osteoporotic Fractures in Men Sleep Study (MrOS) \cite{Blackwell2011-ca_mros, Zhang2018-sw_nssr}, Sleep Heart Health Study (SHHS) \cite{Quan1997SleepHeartHealth, Zhang2018-sw_nssr}, Sleep-EDF). 
The datasets span multiple clinical populations, including children (CFS), and people with cognitive impairment (CI) (PhysioNet 2026) and obstructive sleep apnea (OSA) (DOD, UCDDB). Furthermore, MASS SS01 is annotated for microarousals, PhysioNet 2026 contains labels for microarousals, respiratory events and limb movements, and UCDDB is annotated for respiratory events.

\subsection{Brain Aging and Neurodegenerative Diseases}

The brain age domain is evaluated on 10 cohorts of PSG recordings which include age labels, spanning $\sim$15{,}000 patients and $\sim$193k hours (CFS, HomePAP, ISRUC, MESA, MrOS, PhysioNet 2026, SHHS, SleepEDF Cassette subset (SC), STAGES, and WSC). Furthermore, PhysioNet 2026 includes patients with cognitive impairment as well as healthy controls.

\subsection{BCI}

The benchmark spans a diverse set of BCI paradigms, including \textbf{motor imagery (MI)} (BNCI2014\_001 \citep{tangermann2012review}, BNCI2014\_004 \citep{tangermann2012review}, BNCI2015\_001 \citep{faller2012autocalibration}, Weibo2014 \citep{yi2014evaluation}, Dreyer2023 \citep{pillette2021experimenters}, Shin2017A \citep{shin2016open}, Liu2024 \citep{liu2024eeg}), \textbf{ERP} (BI2013a \citep{vaineau2019brain}, BI2014a \citep{korczowski2019brain}, EPFLP300 \citep{hoffmann2008efficient}, BNCI2014\_008 \citep{riccio2013attention}, ErpCore2021\_N170 \citep{kappenman2021erp}), and \textbf{SSVEP} (Nakanishi2015 \citep{nakanishi2015comparison}, Kim2025BetaRange \citep{kim202540}), complemented by \textbf{cognitive and affective datasets }(DREAMER valence/arousal \cite{dreamer}, EEGMat \cite{zyma2019electroencephalograms}, ArithmeticTask \cite{rodriguez2019tracking}). In total, the benchmark comprises $509$ participants and exhibits substantial heterogeneity in acquisition, including variation in channel configurations, sampling rates, and recording designs (event-related trials vs.\ continuous sessions). Paradigm structure varies accordingly, with MI datasets providing multi-second trials with moderate sample sizes per subject, ERP (P300 and N170) datasets yielding high-density event-locked epochs, and SSVEP datasets consisting of short ($\approx$2-6\,s) frequency-tagged trials. Combined with continuous recordings, this amounts to approximately 159 hours of EEG. 

\section{Methodology} \label{Appendix:Experiments}

\subsection{Epilepsy}
\label{app:epilepsy}

To evaluate seizure detection, we split EEG recordings in windows of 10 seconds without overlap, and extract embeddings per window for EEG-FMs, TS-FMs, and Supervised-Pre models as frozen-backbone encoders. We assess embeddings using class-balanced logistic regression trained using LBFGS. The L2 regularization strength $C$ is selected by grid search
over $\{10^{-3}, 10^{-2}, 10^{-1}, 1, 10, 10^{2}\}$ on a held-out
validation fold, with selection criterion AUPRC for binary seizure
detection and macro-F1 for multi-class seizure-type classification.
The probe is refit at the selected $C$ on the full training fold before
test evaluation. Folds are defined at the patient-level.

\paragraph{Event-Sens@FA AUC.}
The probe outputs a seizure probability per 10\,s window. For a decision threshold $\tau$, contiguous positive windows are merged into predicted events; a reference seizure counts as detected if any predicted event overlaps it, and a predicted event overlapping no reference seizure counts as a false alarm (any-overlap scoring following SzCORE~\cite{dan2024szcore}). Writing $N$ for the number of reference seizures, $\mathrm{FA}(\tau)$ for the number of false alarms and $T$ for the total
recording duration in hours, this gives 
\begin{equation} 
S(\tau) = \frac{1}{N}\sum_{i=1}^{N} \mathbbm{1}\!\left[\text{seizure } i \text{ overlaps some predicted event at } \tau\right], \qquad
F(\tau) = \frac{\mathrm{FA}(\tau)}{T}.
\end{equation}
Sweeping $\tau$ traces a sensitivity-vs-false-alarm curve, which we reparametrise on a $\log_{10}$ false-alarm axis: for fold $k$, let $s_k(f)$ denote sensitivity at $F(\tau)=10^{f}$, obtained by interpolating the fold's operating points. Averaging across the $K=5$ subject-disjoint folds at matched $f$ and integrating over the clinically relevant range $f \in [-1, 2]$, i.e.\ $0.1$--$100$\,FA/h, yields
\begin{equation}
\bar{s}(f) = \frac{1}{K}\sum_{k=1}^{K} s_k(f),
\qquad
\text{Event-Sens@FA AUC} = \frac{1}{3}\int_{-1}^{2} \bar{s}(f)\,\mathrm{d}f .
\end{equation}
The $\log_{10}$ axis makes each of the three decades contribute equally rather than letting the high-alarm end dominate, and the factor $1/3$ normalises by the width of the interval so the metric is bounded in $[0,1]$. It can therefore be read as the mean event-level sensitivity a system would achieve across log-spaced clinical alarm budgets: $1.0$ means every annotated seizure is recovered even at $0.1$\,FA/h, while $0.5$ means half the seizures are detected on average across the range. Unlike window-level AUROC, the metric is computed on merged events rather than isolated windows and is bounded by the alarm budget, so a model that separates individual windows well can still score poorly if its positives are fragmented or scattered through background.



\subsection{Sleep} \label{Appendix:Experiments-Sleep}

To evaluate sleep, EEG-FMs, TS-FMs, and Supervised-Pre models are consistently evaluated as embedding extractors, by extracting embeddings per 30-second window. For epoch-level tasks, a linear probe is applied to individual epoch-level embeddings; for recording-level diagnosis, embeddings are average-pooled across each recording before classification. Event annotations were converted to binary epoch labels, which however does not assess precise event onset, offset, or duration. We fit a linear probe using LBFGS. For event-detection and diagnosis tasks, a balanced cross-entropy loss is used to account for class imbalance, with default L2 regularization with C = 1.0, while unweighted loss is applied for the sleep staging task, where the regularization strength is selected by grid search over $\{10^{-3}, 10^{-2}, 10^{-1}, 1, 10, 10^{2}\}$ on a held-out validation fold, with Cohen's $\kappa$ as selection criterion. Event detection is evaluated only on datasets that provide expert annotations for the corresponding events. These technician-scored annotations are converted to binary epoch labels using task-specific duration thresholds derived from AASM criteria \cite{aasm2023}. Prior to training, embeddings are standardized to ensure zero mean and unit variance. Folds are defined at the patient level.

\paragraph{Hypnogram Features}

To further assess clinical relevance of EEG-FMs, we analyze whether models can adequately reconstruct macrostructural hypnogram features used in downstream clinical tasks. The evaluated hypnogram features are shown in Table \ref{tab:hypnogram-features}.

  \begin{table}[htbp]
  \centering
  \caption{Hypnogram feature definitions, adapted from \cite{heremans_wearable_2024}. All durations are in hours. The sleep period spans from the first non-Wake epoch to the last non-Wake epoch. Epoch
   length is 30\,s. }
  \label{tab:hypnogram-features}
  \small
  \begin{tabular}{@{}lll@{}}
  \toprule
  \textbf{Feature} & \textbf{Unit} & \textbf{Definition} \\
  \midrule
  \multicolumn{3}{@{}l}{\textit{Duration \& efficiency}} \\
  TST            & h       & Total sleep time (all non-Wake epochs) \\
  TimeW          & h       & Total time in Wake \\
  TimeN1         & h       & Total time in N1 \\
  TimeN2         & h       & Total time in N2 \\
  TimeN3         & h       & Total time in N3 \\
  TimeR          & h       & Total time in REM \\
  \midrule
  \multicolumn{3}{@{}l}{\textit{Sleep-stage proportions}} \\
  RelN1          & --      & $\text{TimeN1} / \text{TST}$ \\
  RelN2          & --      & $\text{TimeN2} / \text{TST}$ \\
  RelN3          & --      & $\text{TimeN3} / \text{TST}$ \\
  RelR           & --      & $\text{TimeR} / \text{TST}$ \\
  \midrule
  \multicolumn{3}{@{}l}{\textit{Transitions}} \\
  Awakenings     & count   & Transitions into Wake from any other stage \\
  SlStCh         & count   & Total epoch-to-epoch stage changes \\
  ChToR          & count   & Transitions into REM from any other stage \\
  ChToN3         & count   & Transitions into N3 from any other stage \\
  \midrule
  \multicolumn{3}{@{}l}{\textit{Bout analysis}} \\
  TimeBetweenW   & h       & Mean inter-bout interval between Wake bouts \\
  TimeBetweenR   & h       & Mean inter-bout interval between REM bouts \\
  TimeBetweenN3  & h       & Mean inter-bout interval between N3 bouts \\
  WBoutDur       & h       & Mean duration of contiguous Wake bouts \\
  RBoutDur       & h       & Mean duration of contiguous REM bouts \\
  N3BoutDur      & h       & Mean duration of contiguous N3 bouts \\
  \midrule
  \multicolumn{3}{@{}l}{\textit{Latencies}} \\
  WASO           & h       & Wake after sleep onset (Wake time within sleep period) \\
  REMLatency     & h       & Time from sleep onset to first REM epoch \\
  \midrule
  \multicolumn{3}{@{}l}{\textit{normalized transition rates}} \\
  RelAwakenings  & h$^{-1}$  & $\text{Awakenings} / \text{TST}$ \\
  RelSlStCh      & h$^{-1}$  & $\text{SlStCh} / \text{TST}$ \\
  RelChToR       & h$^{-1}$  & $\text{ChToR} / \text{TST}$ \\
  RelChToN3      & h$^{-1}$  & $\text{ChToN3} / \text{TST}$ \\
  RelWakeTime    & --      & $\text{WASO} / \text{TST}$ \\
  \midrule
  \multicolumn{3}{@{}l}{\textit{Temporal distribution within sleep period}} \\
  RelOccN1       & --      & Mean normalized position of N1 epochs ($0$--$1$) \\
  RelOccN2       & --      & Mean normalized position of N2 epochs ($0$--$1$) \\
  RelOccN3       & --      & Mean normalized position of N3 epochs ($0$--$1$) \\
  RelOccR        & --      & Mean normalized position of REM epochs ($0$--$1$) \\
  \bottomrule
  \end{tabular}
  \end{table}
\subsection{Brain Age Prediction} \label{Appendix:Experiments-Age}

\paragraph{Brain-Age Regression}

We mapped 30-second sleep epochs to embedding vectors and trained linear ridge-regression probes on the resulting cached features, without updating the encoder. The primary analysis used subject-level mean pooling, where epoch embeddings from each recording were averaged into a single feature vector before regression.

Ridge regression was used as the downstream probe because it provides a linear result while controlling for overfitting through the regularization parameter $\alpha$. For each dataset, we used a fixed five-fold subject-level cross-validation protocol shared across all models. Folds were stratified by age bins  years. Within each outer training fold, $\alpha$ was selected by nested cross-validation over:
\[
\alpha \in \{0.001,\ldots,100\},
\]
using validation MAE. The selected model was then refit on the full outer training set and evaluated on the held-out test fold. For multi-visit cohorts, folds ensured that recordings from the same individual never appeared in both training and test sets.

\paragraph{Brain-age Gap and Clinical Evaluation}

For each held-out subject, we computed the brain-age gap (BAG) as: 
\[
\mathrm{BAG}(x) = \hat{y}(x) - y(x),
\]
where $\hat{y}(x)$ is the predicted age from sleep EEG recording $x$ and $y(x)$ is the chronological age. Since the age-prediction model was trained on the remaining healthy subjects in the training split, BAG reflects deviation from a normative healthy aging trajectory. Positive values indicate an older-appearing brain.

We report two complementary BAG-based statistics. The mean BAG difference:
\[
\Delta \mathrm{BAG}
=
\overline{\mathrm{BAG}}_{\mathrm{CI}}
-
\overline{\mathrm{BAG}}_{\mathrm{Healthy}},
\]
which quantifies group-level shifts in apparent brain age. Its 95\% confidence interval was computed using the independent-samples standard error.

We additionally report  AUROC to assess rank-based Healthy--Cognitive Impairement separation:
\[
\mathrm{AUROC}
=
P\left(\mathrm{BAG}_{\mathrm{CI}} > \mathrm{BAG}_{\mathrm{Healthy}}\right),
\]
Values above 0.5 indicate separation in the expected accelerated-ageing direction, whereas values below 0.5 indicate separation opposite to this direction. 
For AUROC, 95\% confidence intervals were estimated using 1,000 matched-pair bootstrap resamples. Each resample selected age-matched case--control pairs with replacement and included both members of each selected pair, preserving the matching structure. 
\subsection{BCI}\label{Appendix:BCI_Exp}

\paragraph{Heterogeneity of BCI tasks in EEG signal characteristics.}
A key challenge in evaluating EEG foundation models is the large variability across BCI paradigms. Our benchmark includes four paradigms that differ in neural activity, timing, number of classes, and recording hardware.
All foundation models are kept entirely frozen and evaluated via a
logistic regression probe on the extracted embeddings. Raw EEG is
converted to microvolts ($\mu$V) and preprocessed (notch and bandpass
filtering, optional common average re-referencing, resampling); each
model applies its own internal amplitude scaling on the $\mu$V input.
We use leave-one-subject-out (LOSO) cross-validation: In each fold one
subject is held out for testing and the remaining subjects form the
training set. Embeddings are z-score standardized per fold (training
statistics applied to the test set). The probe is an
$\ell_2$-regularized logistic regression
(\texttt{sklearn.linear\_model.LogisticRegression}, $C{=}1.0$, L-BFGS
solver, up to 1\,000 iterations), with balanced class weights for
binary tasks.

\textbf{Motor imagery (MI)} relies on event-related modulations of
sensorimotor rhythms in the $\alpha$~(8-13 Hz) and
$\beta$~(13-30 Hz) bands~\citep{MIBCI_YuanBe20214}. Trials typically
span 3-6 s of imagery within 6-10 s epochs, and discriminative
information is distributed over time, requiring longer integration
windows. Performance is particularly sensitive to inter-subject
variability, as individual differences in brain anatomy, motor strategy,
and neurophysiological patterns can lead to pronounced distribution
shifts between participants~\citep{AHN2015103}. Within MI alone, the
benchmark already exhibits marked heterogeneity: The number of classes
ranges from $C{=}2$ (e.g.\ BNCI2014-004, Dreyer2023, Liu2024,
Shin2017A) through $C{=}4$ (BNCI2014-001) to $C{=}7$ (Weibo2014);
channel counts vary from 3 (BNCI2014-004) to 60 (Weibo2014); sampling
rates span 200-512$\mathrm{Hz}$; and cohort sizes range from 9 to 87 subjects.

\textbf{Event-related potential (ERP)} paradigms (e.g.\ P300, N170)
involve transient neural responses within 170-600 ms
post-stimulus. Trials are considerably shorter (0.5-1 s), but decoding
typically aggregates repetitions over longer sequences, with low
signal-to-noise ratios demanding precise temporal
alignment~\citep{luck2012event}. All five ERP datasets in our benchmark
are binary ($C{=}2$), yet they vary widely in hardware and temporal
resolution: Channel counts range from 8 (BNCI2014-008) to 32
(EPFLP300), and sampling rates from 256 Hz to 2048 Hz, reflecting the
diversity of clinical and research-grade setups. These differences are
particularly challenging for a frozen encoder, which must produce
informative embeddings without adjusting to the specific temporal
structure or hardware characteristics of each dataset.

\textbf{Steady-state visual evoked potentials (SSVEP)} exploit
frequency-locked cortical responses to periodic visual stimulation,
achieving high information transfer rates but remaining sensitive to
attention, fatigue, and harmonic interference~\citep{zhu2010survey}.
This paradigm introduces the most extreme class counts in our
benchmark: $C{=}12$ (Nakanishi2015) and $C{=}40$ (Kim2025, Liu2020B),
bringing the theoretical chance level down to as low as 2.5\%.
Channel counts also span an order of magnitude, from 8 to 64.
Distinguishing among 40 frequency targets from frozen embeddings,
without any spectral fine-tuning, represents one of the most demanding
evaluations in the benchmark.

\textbf{Cognitive and affective state} decoding introduces yet another
source of variability. Trial windows range from short segments
(e.g.,\ 2 s for
workload~\citep{singh2021mental}) to longer epochs lasting tens of
seconds (e.g.,\ the DREAMER affective
dataset~\citep{dreamer}). Class counts vary from
$C{=}2$ (DREAMER, EEGMat) to $C{=}3$ (ArithmeticTask), and the
hardware ranges from consumer-grade devices (Emotiv EPOC at 128 Hz,
14~channels) to research-grade systems (500 Hz, 23~channels). Given
that subject variability in cognitive and affective tasks can be
attributed to differences in brain anatomy, personality, and cognitive
abilities among individuals, the
absence of any subject-specific calibration or model adaptation renders
these tasks particularly challenging.

Taken together, these paradigms span class counts from 2 to 40, channel
counts from 3 to 64, sampling rates from 128 Hz to 2048 Hz, and
cohort sizes from 8 to 100 subjects, all evaluated on entirely frozen
representations with no fine-tuning whatsoever. This heterogeneity
motivates both the use of \emph{normalized balanced accuracy}
(Eq.~\ref{eq:norm_bal_acc}), which places all tasks on a common scale
where 50\% equals chance regardless of~$C$, and the per-dataset
\emph{binomial significance threshold}
(Eq.~\ref{eq:binomial_cdf}), which accounts for the finite number of
evaluation samples. Together, these two normalizations ensure that
cross-paradigm comparisons are neither biased by differing chance levels
nor inflated by small sample sizes, providing a rigorous and unified
framework for assessing the quality of pretrained EEG representations.

\subsubsection{Metrics}
\paragraph{Normalized balanced accuracy}
We evaluate all models using \emph{balanced accuracy}, defined as the
arithmetic mean of per-class recall values:
\begin{equation}
  \mathrm{BA} \;=\; \frac{1}{C}\sum_{c=1}^{C} \frac{\mathrm{TP}_c}{\mathrm{TP}_c + \mathrm{FN}_c}\,,
  \label{eq:balanced_accuracy}
\end{equation}
where $C$ denotes the number of classes and $\mathrm{TP}_c$,
$\mathrm{FN}_c$ are the true positives and false negatives for class~$c$.
Balanced accuracy is preferred over standard accuracy because it is
insensitive to class-frequency imbalance and reduces to the theoretical
chance level $1/C$ for a random classifier regardless of the class
distribution.
However, because the benchmark spans paradigms with widely different
numbers of classes -- ranging from $C{=}2$ (e.g.\ binary motor imagery) to
$C{=}40$ (e.g.\ 40-target SSVEP) -- raw balanced accuracies are not
directly comparable: A score of 60\% is far above chance for a
40-class problem but only marginally so for a binary one.
To enable fair comparison both within and across paradigms, we apply an
affine normalization that maps chance-level performance to 50\% and
perfect classification to 100\%:
\begin{equation}
  \widetilde{\mathrm{BA}} \;=\;
    \frac{\mathrm{BA} - \tfrac{1}{C}}{1 - \tfrac{1}{C}}
    \;\times\; 0.5 \;+\; 0.5\,.
  \label{eq:norm_bal_acc}
\end{equation}
Under this transformation every dataset shares a common chance baseline of
50\% and a common ceiling of 100\%, so that scores can be meaningfully
averaged and compared across tasks with different numbers of classes.

\paragraph{Statistical significance of decoding accuracy}
The theoretical chance level $100/C$\,\% is strictly valid only in the
limit of infinitely many test samples.
As demonstrated by~\citep{BetterThanChance}, classifiers applied
to small datasets, a common scenario in brain-computer interface (BCI)
research, can yield empirical accuracies well above the theoretical
chance level purely by statistical fluctuation.
For example, with only 40 observations a two-class decoder can reach 75\%
accuracy on random data at a rate exceeding $p{=}0.001$.
To guard against such inflated chance-level estimates, we derive a
sample-size--dependent significance threshold using the binomial
cumulative distribution.
Under the assumption that classification errors are binomially
distributed, the probability of correctly classifying at least $z$ out of
$n$ samples by chance with $C$ classes is
\begin{equation}
  P(z) \;=\; \sum_{i=z}^{n}\binom{n}{i}
    \left(\frac{1}{C}\right)^{\!i}
    \left(\frac{C-1}{C}\right)^{\!n-i}.
  \label{eq:binomial_cdf}
\end{equation}
For each dataset we compute the minimum balanced accuracy $\mathrm{BA}^{*}$
such that $P\!\bigl(\lceil \mathrm{BA}^{*}\!\cdot n\rceil\bigr) < \alpha$
with $\alpha{=}0.05$, using the inverse survival function of the binomial
distribution.
This threshold is then mapped to the normalized scale via
Eq.~\eqref{eq:norm_bal_acc}, yielding a per-dataset significance boundary
$\widetilde{\mathrm{BA}}^{*}$ that is directly comparable across all
paradigms.
In all reported figures, results that do not exceed the corresponding
normalized significance threshold are marked accordingly, ensuring that
only decodings that are statistically better than chance, accounting for
both the number of classes and the finite sample size, are interpreted as
meaningful.


\subsection{Statistical tests} 
\label{Appendix:Statistics}
To identify models that reliably rank among the best per domain and across the domains, we apply a binomial win test. For each of the datasets, we rank all models by their mean performance on the main domain-task. Datasets with missing model performances are excluded. We then count, for each model, the number of datasets on which it finishes in the top 3. Under the null hypothesis that rankings are exchangeable across models, the probability of any given model finishing in the top 3 on a single dataset is $p_0 = 3/k$, where $k$ is the number of models compared. We test whether the observed top-3 count exceeds this chance level using a one-sided exact binomial test at $\alpha = 0.05$.
\section{Results} \label{Appendix:Results}

\subsection{Epilepsy}
\subsubsection{Per-type score distributions across model families}
\label{sec:seizure-type-violins}

\begin{figure}[t]
    \centering
    \includegraphics[width=0.9\linewidth]{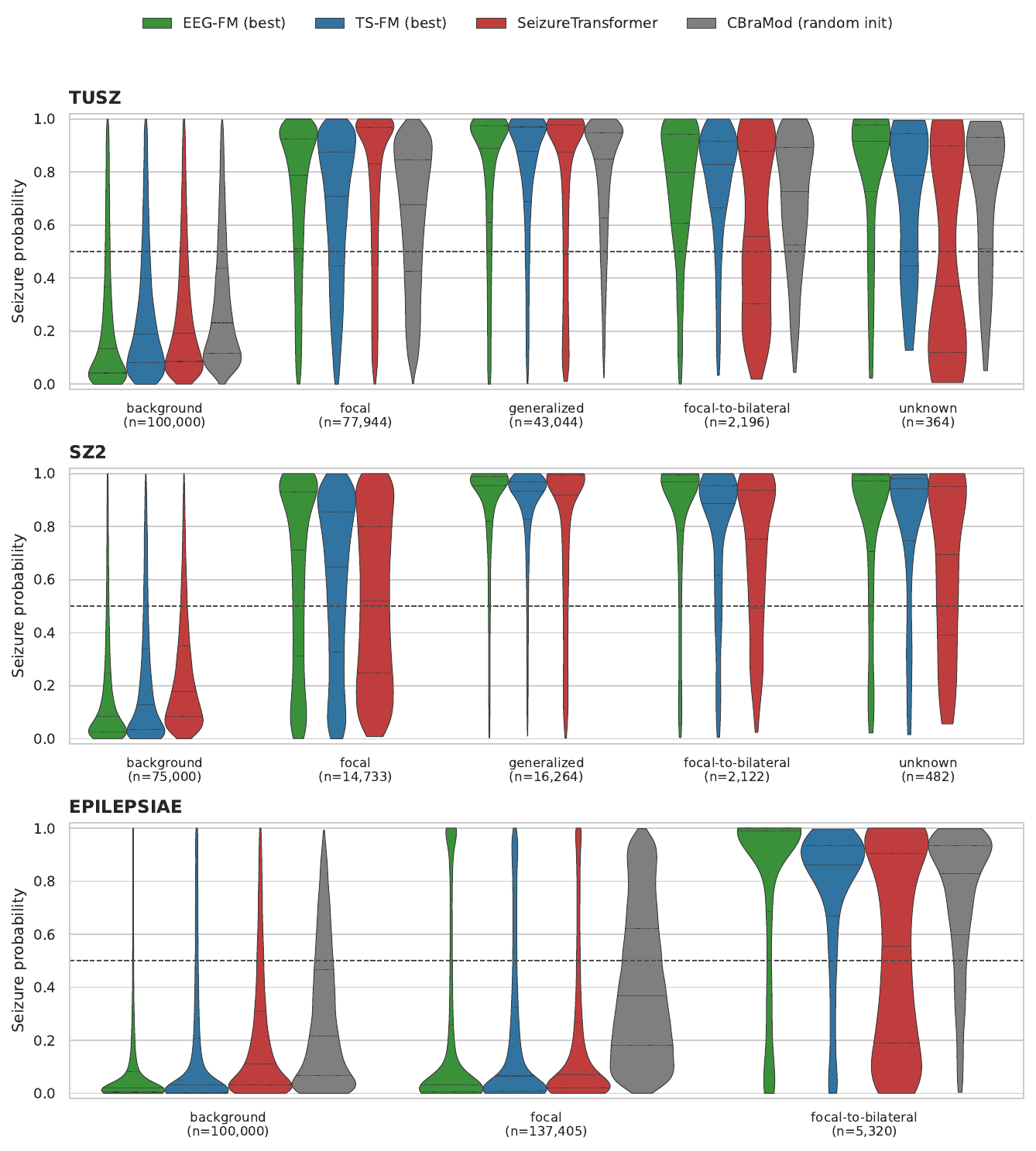}
    \caption{Per-window seizure-probability distributions, stratified by ILAE-2017 main type, on TUSZ (top), SeizeIt2 (middle) and Epilepsiae (bottom). Within each panel, four models are compared: the best EEG foundation model per cohort (\textit{green}; NeuroLM-VQ for TUSZ and SeizeIt2, REVE for Epilepsiae), the best time-series foundation model (\textit{blue}; Moirai-base), the supervised SeizureTransformer (\textit{red}), and a CBraMod random-init baseline (\textit{grey}). Each violin shows the distribution of $\hat{P}(\text{seizure}\mid\text{window})$ over all 10\,s test windows of the indicated type pooled across the five subject-disjoint folds; dashed horizontal lines mark $\hat{P}{=}0.5$. Sample counts per group are noted beneath the x-axis (background windows are subsampled to $25{,}000$ per cell for legibility; all seizure-type windows are retained). Generalized and unknown groups are omitted from the Epilepsiae panel because its annotation schema (1981 ILAE) categorizes every event as focal-onset. The CBraMod random-init baseline is absent from the SeizeIt2 panel because its embeddings were cached at a different window stride and cannot be byte-exactly aligned with the canonical pipeline.}
    \label{fig:seizure-type-violins}
\end{figure}

Scalar metrics summarize a detector's predictions but do not reveal how its confidence distributes across seizure subtypes and background, structure that matters when assessing deployment behavior on rare or clinically critical seizure families. To expose this structure, we plot the per-window seizure probability emitted by classifier, stratified by the ILAE-2017~\cite{avakyan2017ilae} main type of the underlying event. Figure~\ref{fig:seizure-type-violins} contrasts four representative models on TUSZ, SeizeIt2 and Epilepsiae: The best EEG foundation model per dataset (NeuroLM-VQ for TUSZ and SeizeIt2, REVE for Epilepsiae), the best time-series foundation model (Moirai-base), the supervised seizure specialist SeizureTransformer, and a CBraMod random-init baseline that isolates the contribution of pretraining from the architecture and the linear head.

Three patterns stand out. First, the random-init CBraMod baseline emits high seizure probabilities across every type group, including background. The contrast of random initialization with the EEG foundation model, whose distribution concentrates near zero on background, identifies pretraining as the dominant contributor. Second, EEG-FM dominance is dataset-dependent: on TUSZ and SeizeIt2 the green violins on focal, generalized and focal-to-bilateral seizures sit tightly near one, but on Epilepsiae the same architecture exhibits long downward tails and a visibly bimodal distribution on focal events. It is therefore evident that on different datasets the performance on each kind of seizure differs. This analysis suggests that the long pre-surgical Epilepsiae traces, with their subtler focal-onset events, remain difficult for every model in the panel. Third, the supervised seizure specialist does not consistently outperform the foundation models: on TUSZ it exhibits a pronounced false-alarm shoulder on background that the EEG-FM does not.

\subsubsection{Window-level AUROC and event-level sensitivity yield different model rankings}
\label{app:metric-rank-disagreement}
We report two scalar metrics on every seizure-detection cohort, and we argue they are not interchangeable. The first is the mean five-fold window-level AUROC of the linear probe, the metric conventionally reported in the EEG machine-learning literature. The second is the event-based AUC of the median sensitivity-vs-false-alarm-rate curve, integrated against $\log_{10}(\text{FA/h})$ over $[0.1, 100]$ FA/h (``Event-Sens@FA AUC'', bounded in $[0, 1]$). The two quantities measure substantively different things. AUROC summarizes whether each ten-second window is, in isolation, distinguishable from the background distribution. Event-Sens@FA AUC aggregates the same per-window scores into clinically meaningful seizure events and quantifies how many of those events a deployed system would recover at clinically tolerable alarm budgets. Because seizure-detection systems are evaluated and deployed at the event level under explicit false-alarm constraints, Event-Sens@FA AUC is the metric of clinical relevance. We therefore designate it as our primary metric and report AUROC alongside it to expose the consequences of the prevailing convention.

\begin{figure}[t]
    \centering
    \includegraphics[width=0.95\linewidth]{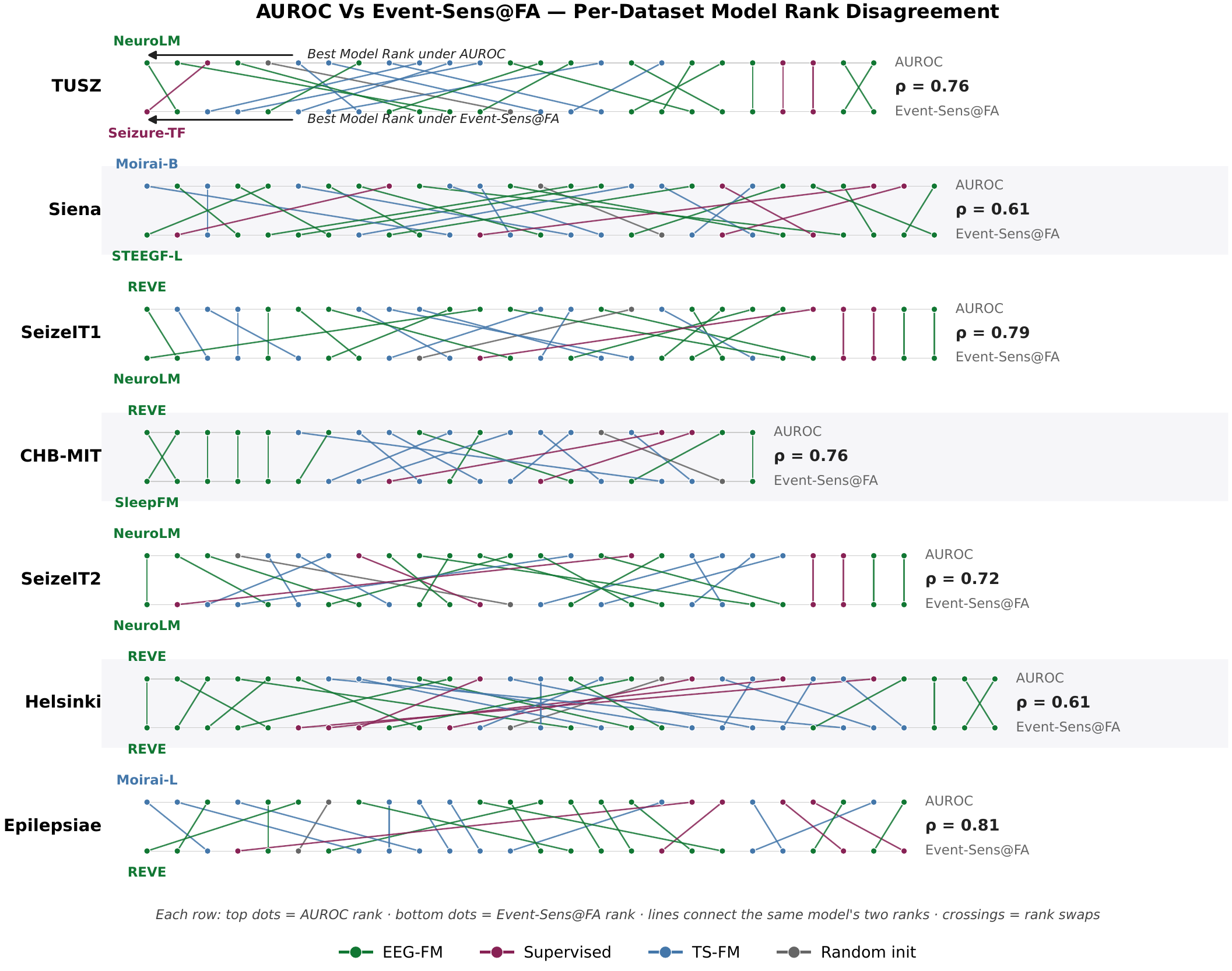}
    \caption{AUROC and event-level Sens@FA yield systematically different model rankings on every seizure-detection cohort. Each row corresponds to one cohort. Within a row, the upper rail of dots is the per-model rank by five-fold-mean window-level AUROC and the lower rail is the per-model rank by the event-based sensitivity-vs-FA/h AUC integrated over $[0.1, 100]$ FA/h; rank 1 (best) is leftmost on both rails. Each line connects the same model's two ranks, coloured by family (green: EEG foundation models; red: supervised pretraining; blue: time-series foundation models; grey: random initialisation). Every line crossing is a pair of models whose relative order flips between the two metrics. Spearman's $\rho$ between the two rankings (right of each row) ranges from 0.61 (Helsinki, Siena) to 0.81 (Epilepsiae) and is below 1 on every cohort, with mean $\bar{\rho} \approx 0.72$ across the seven cohorts. On TUSZ (top row, annotated), NeuroLM is rank-1 by AUROC but is overtaken by the supervised Seizure-TF model on Event-Sens@FA, illustrating that the conventional window-level metric selects a different model than the clinically relevant event-level one. 181 (cohort, model) pairs across the seven seizure-detection cohorts are shown.}
    \label{fig:metric-rank-disagreement}
\end{figure}

Figure~\ref{fig:metric-rank-disagreement} establishes that the choice of metric is consequential. For each of the seven seizure cohorts in our benchmark we rank every probed model on AUROC and on Event-Sens@FA AUC separately, then draw each model as a single line whose top endpoint is its AUROC rank and whose bottom endpoint is its Event-Sens@FA rank. Two lines crossing indicates that the corresponding pair of models swaps relative order between the two metrics. Spearman's $\rho$ printed at the right of each row summarizes the agreement (1.0 denotes identical rankings).

Across the seven cohorts, $\rho$ ranges from 0.61 (Helsinki, Siena) to 0.81 (Epilepsiae), with mean $\bar{\rho} \approx 0.72$. The two metrics never produce identical orderings, and on Helsinki and Siena their agreement is weak enough that they carry largely independent information about which model performs best. The disagreement is most pronounced on TUSZ, where the AUROC-best model, the EEG foundation model NeuroLM, is displaced at the top of the Event-Sens@FA ranking by the fully-supervised seizure detector Seizure-TF. A representation that achieves the highest per-window discriminability does not, in general, produce the cleanest event-level alarm stream. Model selection driven by AUROC, the dominant convention in the broader EEG ML literature, therefore identifies a different model than selection driven by the clinically relevant event-level metric. This motivates our decision to foreground Event-Sens@FA AUC throughout the main results.

\subsubsection{Non-window EEG benchmarks: datasets and foundation-model rankings}
\label{appendix:non_window}

Three datasets in our suite, Bonn, TUAB, and NMT, carry trial- or recording-level binary labels rather than continuous seizure annotations. They therefore do not admit window-level event metrics (sensitivity at fixed false-alarm rate, etc.) and are evaluated separately from the seizure-detection cohorts in the main text.

The Bonn dataset~\citep{andrzejak2001indications} comprises five sets (A--E) of 100 single-channel surface and intracranial EEG segments, each 23.6\,s at 173.6\,Hz, originally released as a small seizure/non-seizure benchmark. We use the canonical \emph{S vs.\ rest} binary split, evaluating set E (ictal) against the four non-ictal sets A--D; after windowing at 10\,s, the cohort contains 1{,}000 segments (800 non-S, 200 S; 1:4 imbalance), evaluated under stratified 5-fold cross-validation at the segment level. Bonn provides no per-trial subject identifier, so subject-grouped cross-validation is not applicable. The Temple University Hospital Abnormal EEG Corpus (TUAB)~\citep{obeid2016temple} contains 2{,}993 clinical EEG sessions from 2{,}329 subjects (train: 2{,}717; official eval: 276), labeled normal or abnormal by board-certified neurologists, with the cohort approximately balanced (1{,}521 normal, 1{,}472 abnormal); we use the official train/eval split, stratifying the train subjects into 5 folds and reporting on the held-out eval set. The NMT EEG Scalp Abnormal dataset~\citep{khan2022nmt} contains 2{,}417 recordings collected at NMT Karachi, labeled normal or abnormal (train: 2{,}232; eval: 185), with severe class imbalance (2{,}002 normal vs.\ 415 abnormal, approximately 83/17\,\%) that makes threshold-dependent metrics more sensitive than on TUAB.

\begin{figure}[t]
  \centering
  \includegraphics[width=0.95\linewidth]{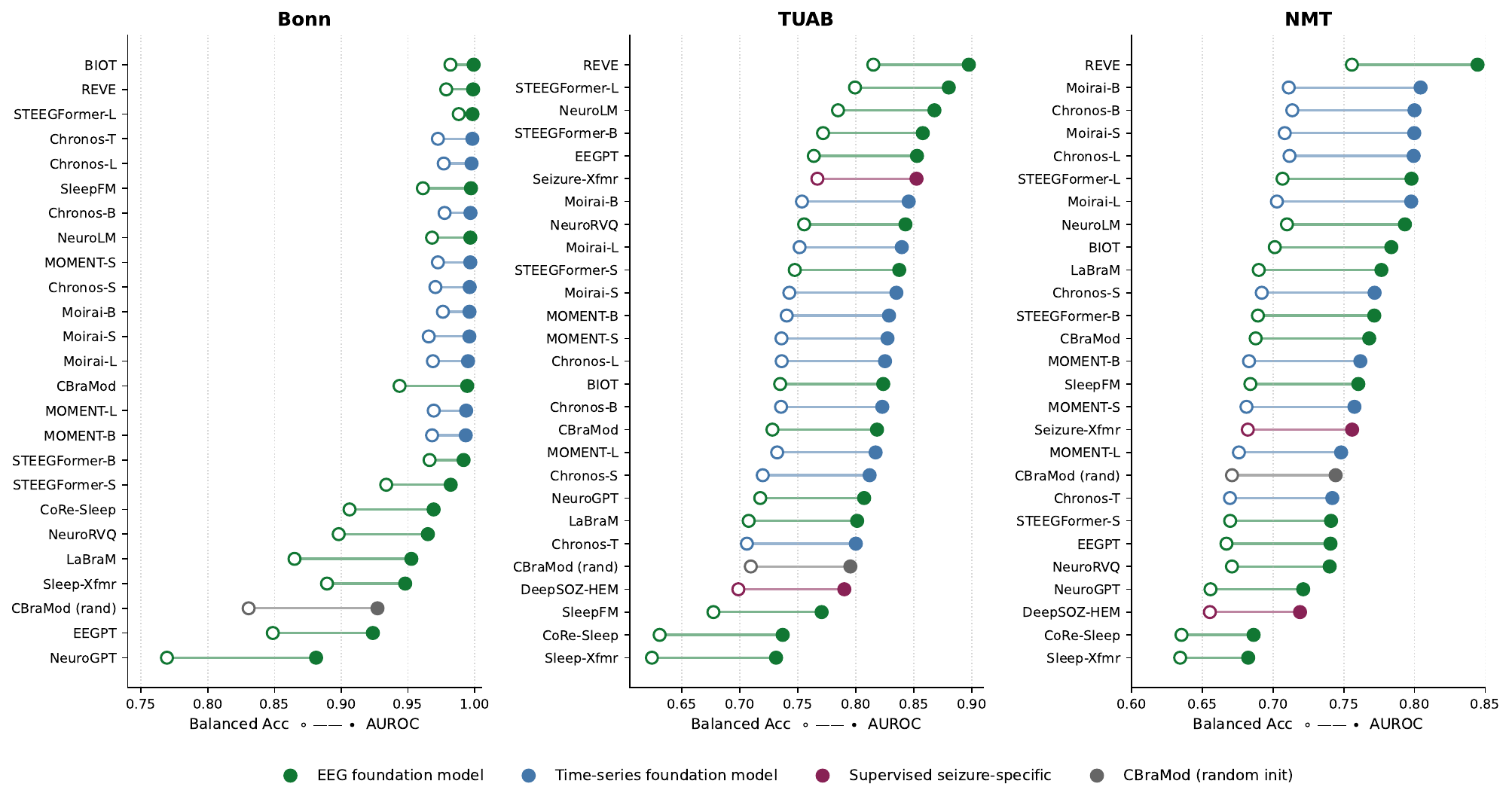}
  \caption{Foundation-model rankings on the three non-window EEG benchmarks (Bonn, TUAB, NMT). For each model and dataset, we plot AUROC (\fillcircle) and balanced accuracy (\opencircle); the connector indicates the threshold-calibration gap (typically 0.05--0.11). Models are sorted within each panel by AUROC and colored by family: dark-green for general EEG foundation models, blue for time-series foundation models, burgundy for supervised seizure-specific baselines (SeizureTransformer, DeepSOZ-HEM), and grey for the CBraMod random-init ablation. All values are means over 5 stratified folds. REVE leads on all three datasets, Bonn is saturated, and NMT is the dataset on which TS-FMs come closest to parity with EEG-FMs.}
  \label{fig:non_window_dumbbell}
\end{figure}

For every (model, dataset) pair, we extract the embedding (10\,s windows, 10\,s stride; window-level features pooled to recording level on TUAB and NMT) and fit a 5-fold sklearn logistic-regression probe with $C \in \{10^{-3}, \ldots, 10^{2}\}$ tuned on the validation fold. Figure~\ref{fig:non_window_dumbbell} pairs the threshold-free AUROC (\fillcircle) with the threshold-dependent balanced accuracy (\opencircle) for every model on each dataset; the connecting segment indexes the calibration gap. Models are colored 

Bonn saturates: eleven models reach AUROC $\geq 0.99$ (BIOT, REVE, STEEGFormer-L, all four Chronos sizes, NeuroLM, SleepFM, MOMENT-S, and CBraMod), confirming that the S-versus-rest split is recoverable from almost any representation. We therefore treat Bonn as a sanity check rather than a discriminating benchmark; the only meaningful cross-model signal lies in the lower tail, where NeuroGPT (0.881), EEGPT (0.924), and the CBraMod random-init ablation (0.927) trail the saturation ceiling. TUAB orders the EEG foundation models more cleanly: REVE leads at 0.898 AUROC, followed by STEEGFormer-L (0.880), NeuroLM (0.868), STEEGFormer-B (0.858), and EEGPT (0.853); the supervised SeizureTransformer is sixth (0.852), competitive but not specialized for this task, while DeepSOZ-HEM places mid-low (0.790), suggesting that a TUSZ-fitted seizure head transfers only partially to abnormality detection. The three sleep-trained EEG models (SleepFM 0.771, CoRe-Sleep 0.737, Sleep-TF 0.731) sit at the bottom of the panel, indicating limited transfer from sleep-stage pretraining to clinical abnormality. On the imbalanced NMT abnormality task, REVE again leads (0.845), but the next four positions are essentially tied between time-series and EEG foundation models (Moirai-B 0.804, Chronos-B 0.800, Moirai-S 0.800, Chronos-L 0.799), making NMT the dataset on which TS-FMs come closest to parity with EEG-FMs; DeepSOZ-HEM (0.719) and the SHHS-pretrained sleep transformers (CoRe-Sleep 0.686, Sleep-TF 0.682) again sit at the bottom.

Two cross-cutting patterns emerge. First, calibration gaps are systematic rather than anomalous: across every (model, dataset) pair, the AUROC-to-balanced-accuracy gap (segment length in Figure~\ref{fig:non_window_dumbbell}) lies between 0.05 and 0.11, the expected offset between threshold-free discrimination and the probe's default 0.5 decision boundary, with no model showing a calibration anomaly that would disqualify a strong AUROC on balanced-accuracy grounds. Second, pretraining provides a small but consistent boost: the CBraMod random-init points fall short of the pretrained CBraMod points by 0.067 AUROC on Bonn (0.994 $\rightarrow$ 0.927), 0.023 on TUAB (0.818 $\rightarrow$ 0.795), and 0.024 on NMT (0.768 $\rightarrow$ 0.744). The Bonn drop is largest in absolute terms but both endpoints remain near the saturation ceiling; on the more discriminating TUAB and NMT, pretraining buys roughly 2--3 percentage points of AUROC.

\subsubsection{Statistical testing across epilepsy dataset}
\label{sec:epilepsy-stat-testing}

Ranking first on a single dataset does not establish broad competence: A model can lead one benchmark through alignment with a particular recording style or through an artefact of a single split. We therefore evaluate cross-dataset consistency directly. For each model we count the dataset on which it ranks in the top three by Event-Sens@FA AUC, and apply an exact one-sided binomial test against the chance rate $p_0 = 3/k$, where $k$ is the size of the comparator pool. Results are reported for two pools on the 7 epileptic datasets. The first comprises all 23 models and tests competitiveness against the full benchmark, which includes supervised seizure detectors, time-series foundation models, and a random-initialized baseline. The second restricts to the 11 EEG foundation models and tests competitiveness within the family.

Table~\ref{tab:top3_win_test} reports the outcome. Two models reach significance in both pools at $\alpha = 0.05$. NeuroLM ranks top-3 on 6 of 7 datasets ($p < 10^{-3}$ in the full pool, $p = 0.002$ within EEG-FMs), and REVE on 5 of 7 in the full pool and 6 of 7 within EEG-FMs (both $p \leq 0.002$). No other model exceeds the chance rate at $\alpha = 0.05$ in either pool. The supervised seizure specialist Seizure-TF is the closest non-significant case at 3 of 7 wins ($p = 0.052$), indicating that task-specific supervision is strong on a subset of datasets but does not generalize reliably across recording styles. No time-series foundation model reaches significance, and the random-initialized CBraMod baseline ranks top-3 on no datasets, consistent with the score-distribution analysis in Section~\ref{sec:seizure-type-violins}. Among the eleven EEG foundation models, nine fail to exceed the chance rate even within the EEG-FM-only pool. The claim that ``EEG foundation models work for seizure detection'' is therefore supported, on this benchmark, by two specific models (NeuroLM and REVE) rather than by the family as a whole.

\begin{table}[t]
\centering
\small
\caption{\textbf{Binomial top-3 win test across the seven seizure-detection datasets.} For each model, the exact one-sided binomial test asks whether the number of datasets on which the model ranks in the top 3 (by Event-Sens vs FA/h AUC, integrated over $[0.1,100]$ FA/h) exceeds the chance rate $p_0 = 3/k$. Two pools are tested independently: all $k=23$ probed models, and the $k=11$ EEG foundation models in isolation. Bold rows are significant at $\alpha=0.05$.}
\label{tab:top3_win_test}
\begin{tabular}{l l c c c c}
\toprule
 & & \multicolumn{2}{c}{All models ($k=23$)} & \multicolumn{2}{c}{EEG-FM ($k=11$)} \\
\cmidrule(lr){3-4}\cmidrule(lr){5-6}
Model & Family & Wins & $p$ & Wins & $p$ \\
\midrule
\textbf{NeuroLM} & \textbf{EEG-FM} & \textbf{6/7} & \textbf{$<\!10^{-3}$} & \textbf{6/7} & \textbf{0.002} \\
\textbf{REVE} & \textbf{EEG-FM} & \textbf{5/7} & \textbf{$<\!10^{-3}$} & \textbf{6/7} & \textbf{0.002} \\
Seizure-TF & Supervised-Pre & 3/7 & 0.052 & -- & -- \\
Moirai-L & TS-FM & 2/7 & 0.229 & -- & -- \\
Chronos-B & TS-FM & 1/7 & 0.624 & -- & -- \\
Moirai-B & TS-FM & 1/7 & 0.624 & -- & -- \\
Moirai-S & TS-FM & 1/7 & 0.624 & -- & -- \\
NeuroRVQ & EEG-FM & 1/7 & 0.624 & 1/7 & 0.892 \\
SleepFM & EEG-FM & 1/7 & 0.624 & 2/7 & 0.610 \\
BIOT & EEG-FM & 0/7 & 1.000 & 1/7 & 0.892 \\
CBraMod & EEG-FM & 0/7 & 1.000 & 1/7 & 0.892 \\
CBraMod (rand) & Random-Init & 0/7 & 1.000 & -- & -- \\
Chronos-S & TS-FM & 0/7 & 1.000 & -- & -- \\
Chronos-T & TS-FM & 0/7 & 1.000 & -- & -- \\
CoRe-Sleep & Supervised-Pre & 0/7 & 1.000 & -- & -- \\
EEGPT & EEG-FM & 0/7 & 1.000 & 1/7 & 0.892 \\
LaBraM & EEG-FM & 0/7 & 1.000 & 1/7 & 0.892 \\
MOMENT-B & TS-FM & 0/7 & 1.000 & -- & -- \\
MOMENT-S & TS-FM & 0/7 & 1.000 & -- & -- \\
NeuroGPT & EEG-FM & 0/7 & 1.000 & 0/7 & 1.000 \\
Sleep-TF & Supervised-Pre & 0/7 & 1.000 & -- & -- \\
STEEGFormer-B & EEG-FM & 0/7 & 1.000 & 2/7 & 0.610 \\
STEEGFormer-S & EEG-FM & 0/7 & 1.000 & 0/7 & 1.000 \\
\bottomrule
\end{tabular}
\end{table}
\subsection{AUROC, balanced accuracy, and event-level metrics per model}
\label{app:auroc_metrics}

To complement the per-dataset view in the main text, this section collapses the seven seizure-detection datasets into a single per-model summary across the three metrics. The pooled view makes it easier to see which models are competitive on average, how the random-init baseline is positioned once dataset-specific noise is averaged out, and how rankings shift between window-level (AUROC, balanced accuracy) and event-level (Event-Sens@FA AUC) metrics.

Table~\ref{tab:cross_dataset_metrics_summary} reports across-dataset means and standard deviations for each model on the three metrics. Three observations stand out. First, only NeuroLM ($0.553$) and REVE ($0.549$) clearly exceed the random-init Event-Sens@FA AUC of $0.440$; SleepFM ($0.487$) and STEEGFormer-B ($0.451$) sit just above, while the remaining seven EEG-FMs sit at or below random init, with NeuroGPT ($0.277$) far below. Pretrained CBraMod records the same $0.440$ as CBraMod-rand, indicating that for some models pretraining does not deliver an event-level gain. Second, metric choice changes the ranking: REVE leads on AUROC ($0.839$) followed by Moirai-L ($0.830$) and NeuroLM ($0.821$), whereas the event-level ranking is NeuroLM, REVE, then Seizure-TF, with Moirai-L falling to fourth (consistent with Appendix~\ref{app:metric-rank-disagreement}). Third, families do not separate cleanly: Seizure-TF reaches $0.519$ on Event-Sens@FA partly through pretraining overlap with TUSZ and Siena, and Moirai-L outperforms $9$ of $11$ EEG-FMs on Event-Sens@FA despite no EEG-specific pretraining, so EEG-specific pretraining alone does not guarantee strong event-level performance.

\begin{table}[h]
\centering
\small
\caption{\textbf{Per-model summary across the seven seizure-detection cohorts.} Each cell is the mean $\pm$ across-cohort standard deviation (7 datasets: TUSZ, Siena, SeizeIT1, CHB-MIT, SeizeIT2, Helsinki, Epilepsiae). Window-level AUROC and balanced accuracy come from the linear probe (5-fold subject-disjoint mean per cohort). The event-level metric is the area under the mean event sensitivity vs.\ $\log_{10}(\text{FA/h})$ curve, integrated over the clinically relevant $[0.1, 100]$ FA/h range under any-overlap event scoring. Models are grouped by family and sorted within family by Event-Sens vs FA/h AUC. \textit{All 23 models in the canonical $7\times 23$ panel are reported.}}
\label{tab:cross_dataset_metrics_summary}
\begin{tabular}{l l c c c}
\toprule
Model & Family & AUROC & Balanced Acc.\ & Event-Sens vs FA/h AUC \\
\midrule
NeuroLM & EEG-FM & 0.821 $\pm$ 0.079 & 0.640 $\pm$ 0.050 & 0.553 $\pm$ 0.093 \\
REVE & EEG-FM & 0.839 $\pm$ 0.086 & 0.664 $\pm$ 0.069 & 0.549 $\pm$ 0.098 \\
SleepFM & EEG-FM & 0.789 $\pm$ 0.099 & 0.603 $\pm$ 0.052 & 0.487 $\pm$ 0.117 \\
STEEGFormer-B & EEG-FM & 0.776 $\pm$ 0.077 & 0.598 $\pm$ 0.061 & 0.451 $\pm$ 0.118 \\
CBraMod & EEG-FM & 0.787 $\pm$ 0.110 & 0.611 $\pm$ 0.070 & 0.440 $\pm$ 0.118 \\
EEGPT & EEG-FM & 0.752 $\pm$ 0.097 & 0.607 $\pm$ 0.081 & 0.431 $\pm$ 0.099 \\
LaBraM & EEG-FM & 0.764 $\pm$ 0.079 & 0.582 $\pm$ 0.062 & 0.418 $\pm$ 0.095 \\
NeuroRVQ & EEG-FM & 0.756 $\pm$ 0.111 & 0.596 $\pm$ 0.066 & 0.410 $\pm$ 0.095 \\
BIOT & EEG-FM & 0.756 $\pm$ 0.082 & 0.591 $\pm$ 0.054 & 0.408 $\pm$ 0.074 \\
STEEGFormer-S & EEG-FM & 0.746 $\pm$ 0.073 & 0.577 $\pm$ 0.051 & 0.405 $\pm$ 0.099 \\
NeuroGPT & EEG-FM & 0.618 $\pm$ 0.113 & 0.555 $\pm$ 0.050 & 0.277 $\pm$ 0.048 \\
\midrule
Seizure-TF & Supervised-Pre & 0.763 $\pm$ 0.111 & 0.613 $\pm$ 0.074 & 0.519 $\pm$ 0.101 \\
Sleep-TF & Supervised-Pre & 0.667 $\pm$ 0.066 & 0.554 $\pm$ 0.050 & 0.357 $\pm$ 0.086 \\
CoRe-Sleep & Supervised-Pre & 0.668 $\pm$ 0.068 & 0.556 $\pm$ 0.050 & 0.354 $\pm$ 0.080 \\
\midrule
Moirai-L & TS-FM & 0.830 $\pm$ 0.067 & 0.616 $\pm$ 0.053 & 0.474 $\pm$ 0.119 \\
Moirai-B & TS-FM & 0.807 $\pm$ 0.093 & 0.625 $\pm$ 0.053 & 0.453 $\pm$ 0.123 \\
Chronos-S & TS-FM & 0.755 $\pm$ 0.078 & 0.590 $\pm$ 0.062 & 0.443 $\pm$ 0.110 \\
MOMENT-S & TS-FM & 0.754 $\pm$ 0.069 & 0.590 $\pm$ 0.053 & 0.443 $\pm$ 0.068 \\
Moirai-S & TS-FM & 0.785 $\pm$ 0.138 & 0.601 $\pm$ 0.076 & 0.436 $\pm$ 0.160 \\
Chronos-B & TS-FM & 0.760 $\pm$ 0.083 & 0.583 $\pm$ 0.056 & 0.431 $\pm$ 0.094 \\
MOMENT-B & TS-FM & 0.756 $\pm$ 0.101 & 0.586 $\pm$ 0.065 & 0.428 $\pm$ 0.101 \\
Chronos-T & TS-FM & 0.743 $\pm$ 0.097 & 0.575 $\pm$ 0.058 & 0.422 $\pm$ 0.087 \\
\midrule
CBraMod (rand) & Random-Init & 0.773 $\pm$ 0.086 & 0.590 $\pm$ 0.064 & 0.440 $\pm$ 0.102 \\
\bottomrule
\end{tabular}
\end{table}

\subsection{Leave-One-Dataset-Out}
\label{app:epilepsy_lodo}
Whether a pretrained representation supports seizure detection on a cohort with no labels is a different question from how it performs once target labels are available. We therefore repeat the epilepsy evaluation in a leave-one-dataset-out (LODO) setting: for each held-out dataset, we fit a single linear probe on the six remaining datasets pooled and evaluate it on the held-out cohort, with no target data used at any stage. Table 5 reports six representative models: the two EEG-FMs that pass the binomial top-3 test (NeuroLM, REVE), the supervised seizure specialist Seizure-TF, two TS-FMs (Chronos-B, Moirai-L), and the CBraMod random-init reference, against their within-dataset scores.

Transfer is costly, and no model escapes it. Mean Event-Sens@FA AUC falls from 0.431–0.553 within-dataset to 0.275–0.362 under LODO, retaining 55–76\% of within-dataset performance, with only 3 of 42 (model, dataset) cells improving. Every LODO mean lies below the 0.440 within-dataset random-init reference: a probe fitted without target labels does not reach what a randomly initialized encoder reaches with them. Rankings also reshuffle, with REVE retaining the least (55\%) and dropping below random-init CBraMod despite matching NeuroLM within-dataset. No family transfers best in general: the per-cohort best alternates between NeuroLM, Chronos-B, and Seizure-TF, the latter only on the two datasets it was itself pretrained on. Cohort composition dominates the failure modes, with the neonatal Helsinki recordings hardest for every model (all $\leq0.247$) and Epilepsiae the only cohort where LODO can match the within-dataset probe. The epilepsy domain therefore mirrors the sleep and brain-age LODO results: the narrow within-dataset advantages of individual EEG-FMs do not survive the removal of target labels.

\begin{table}[h]
  \centering\small\setlength{\tabcolsep}{4pt}
  \renewcommand{\cellalign}{cc}
  \caption{\textbf{Leave-one-dataset-out seizure detection, with the within-dataset reference.} Event-Sens@FA AUC: area under event-level sensitivity vs.\ $\log_{10}$ FA/h over 0.1--100\,FA/h, any-overlap scoring, 10\,s windows. In every cell the upper figure is LODO, a linear probe fitted on the six remaining datasets pooled and evaluated on the held-out dataset, and the grey figure below is the \emph{within-dataset} result for the same model and dataset (mean $\pm$ SD over 5 folds). LODO is a single deterministic fit, so it has no within-cell error bar. Bold marks the best LODO score per held-out dataset. Transfer costs 24--45\% of within-dataset performance, and only three of 42 cells improve under LODO (NeuroLM and CBraMod on Epilepsiae, Chronos-B on CHB-MIT).}
  \label{tab:lodo-epilepsy}
  \begin{tabular}{lcccccc}
    \toprule
    Held-out & \makecell{NeuroLM\\{\scriptsize EEG-FM}} & \makecell{Seizure-TF\\{\scriptsize Sup.-Pre}} & \makecell{CBraMod\\{\scriptsize Rand-init}} & \makecell{REVE\\{\scriptsize EEG-FM}} & \makecell{Chronos-B\\{\scriptsize TS-FM}} & \makecell{Moirai-L\\{\scriptsize TS-FM}} \\
    \midrule
    CHB-MIT & \makecell{0.242\\{\scriptsize\color{gray}0.494\,$\pm$0.103}} & \makecell{0.263\\{\scriptsize\color{gray}0.385\,$\pm$0.141}} & \makecell{0.235\\{\scriptsize\color{gray}0.267\,$\pm$0.238}} & \makecell{0.129\\{\scriptsize\color{gray}0.506\,$\pm$0.119}} & \makecell{\textbf{0.288}\\{\scriptsize\color{gray}0.283\,$\pm$0.120}} & \makecell{0.203\\{\scriptsize\color{gray}0.307\,$\pm$0.073}} \\
    \addlinespace[1pt]
    Siena & \makecell{0.269\\{\scriptsize\color{gray}0.487\,$\pm$0.148}} & \makecell{\textbf{0.389}\\{\scriptsize\color{gray}0.676\,$\pm$0.142}} & \makecell{0.322\\{\scriptsize\color{gray}0.472\,$\pm$0.098}} & \makecell{0.182\\{\scriptsize\color{gray}0.619\,$\pm$0.151}} & \makecell{0.174\\{\scriptsize\color{gray}0.480\,$\pm$0.127}} & \makecell{0.109\\{\scriptsize\color{gray}0.486\,$\pm$0.159}} \\
    \addlinespace[1pt]
    TUSZ & \makecell{0.254\\{\scriptsize\color{gray}0.498\,$\pm$0.043}} & \makecell{\textbf{0.382}\\{\scriptsize\color{gray}0.530\,$\pm$0.111}} & \makecell{0.249\\{\scriptsize\color{gray}0.418\,$\pm$0.058}} & \makecell{0.249\\{\scriptsize\color{gray}0.423\,$\pm$0.056}} & \makecell{0.290\\{\scriptsize\color{gray}0.483\,$\pm$0.059}} & \makecell{0.223\\{\scriptsize\color{gray}0.412\,$\pm$0.079}} \\
    \addlinespace[1pt]
    SeizeIT1 & \makecell{\textbf{0.622}\\{\scriptsize\color{gray}0.737\,$\pm$0.208}} & \makecell{0.409\\{\scriptsize\color{gray}0.573\,$\pm$0.155}} & \makecell{0.507\\{\scriptsize\color{gray}0.608\,$\pm$0.135}} & \makecell{0.610\\{\scriptsize\color{gray}0.720\,$\pm$0.119}} & \makecell{0.348\\{\scriptsize\color{gray}0.548\,$\pm$0.112}} & \makecell{0.479\\{\scriptsize\color{gray}0.677\,$\pm$0.133}} \\
    \addlinespace[1pt]
    SeizeIT2 & \makecell{\textbf{0.485}\\{\scriptsize\color{gray}0.610\,$\pm$0.127}} & \makecell{0.402\\{\scriptsize\color{gray}0.577\,$\pm$0.103}} & \makecell{0.322\\{\scriptsize\color{gray}0.467\,$\pm$0.120}} & \makecell{0.220\\{\scriptsize\color{gray}0.502\,$\pm$0.038}} & \makecell{0.345\\{\scriptsize\color{gray}0.456\,$\pm$0.079}} & \makecell{0.234\\{\scriptsize\color{gray}0.559\,$\pm$0.115}} \\
    \addlinespace[1pt]
    Helsinki & \makecell{0.078\\{\scriptsize\color{gray}0.493\,$\pm$0.137}} & \makecell{0.240\\{\scriptsize\color{gray}0.415\,$\pm$0.103}} & \makecell{0.188\\{\scriptsize\color{gray}0.398\,$\pm$0.042}} & \makecell{0.225\\{\scriptsize\color{gray}0.498\,$\pm$0.113}} & \makecell{\textbf{0.247}\\{\scriptsize\color{gray}0.326\,$\pm$0.113}} & \makecell{0.229\\{\scriptsize\color{gray}0.400\,$\pm$0.145}} \\
    \addlinespace[1pt]
    Epilepsiae & \makecell{\textbf{0.581}\\{\scriptsize\color{gray}0.551\,$\pm$0.124}} & \makecell{0.311\\{\scriptsize\color{gray}0.477\,$\pm$0.098}} & \makecell{0.517\\{\scriptsize\color{gray}0.451\,$\pm$0.096}} & \makecell{0.500\\{\scriptsize\color{gray}0.578\,$\pm$0.082}} & \makecell{0.371\\{\scriptsize\color{gray}0.438\,$\pm$0.139}} & \makecell{0.446\\{\scriptsize\color{gray}0.478\,$\pm$--}} \\
    \addlinespace[1pt]
    \midrule
    Mean & \makecell{\textbf{0.362}\\{\scriptsize\color{gray}0.553}} & \makecell{0.342\\{\scriptsize\color{gray}0.519}} & \makecell{0.334\\{\scriptsize\color{gray}0.440}} & \makecell{0.302\\{\scriptsize\color{gray}0.549}} & \makecell{0.295\\{\scriptsize\color{gray}0.431}} & \makecell{0.275\\{\scriptsize\color{gray}0.474}} \\
    \bottomrule
  \end{tabular}
\end{table}

\subsection{Sleep} \label{Appendix:Results-Sleep}

\subsubsection{Sleep staging}
\begin{figure}
    \centering
    \includegraphics[width=\textwidth]{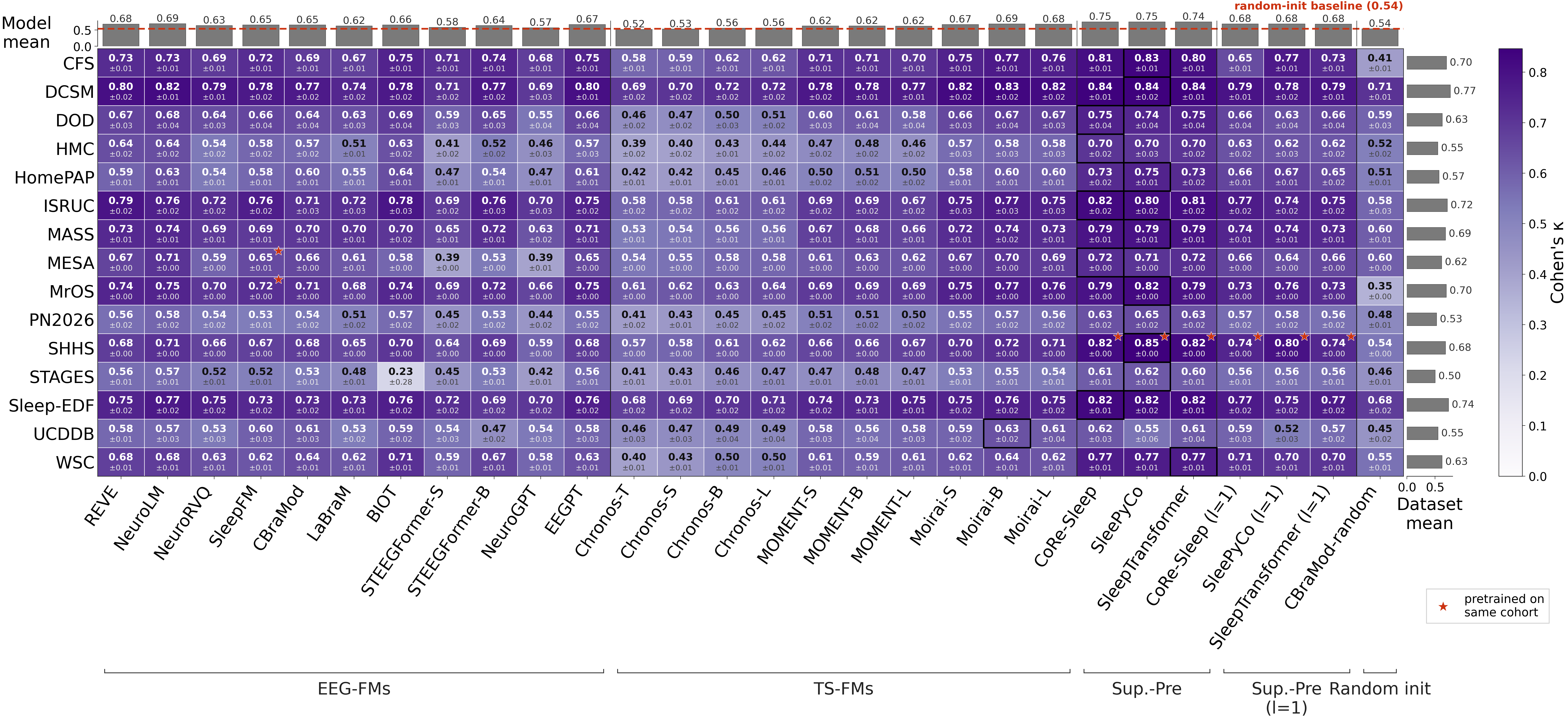}
    \caption{\textbf{Sleep staging performance is strongly dataset-dependent.} Cohen's $\kappa$ sleep staging agreement with ground-truth labels for all datasets and models per model category, i.e., EEG-FMs, TS-FMs, and Supervised-Pre models, with and without temporal context (sequence length l=1), and a random-initialized reference. Cohen's $\kappa$ (above) and standard deviation over folds (below) are shown for each. Black borders mark the cohort best-performing model. Red stars flag pretraining/evaluation overlap. Grey bars show mean performance over dataset and models respectively, with the red dashed line indicating mean random-initialized baseline. Missing results are indicated with a cross.}
    \label{fig:SleepOverview}
\end{figure}

\begin{figure}
    \centering
    \includegraphics[width=\textwidth]{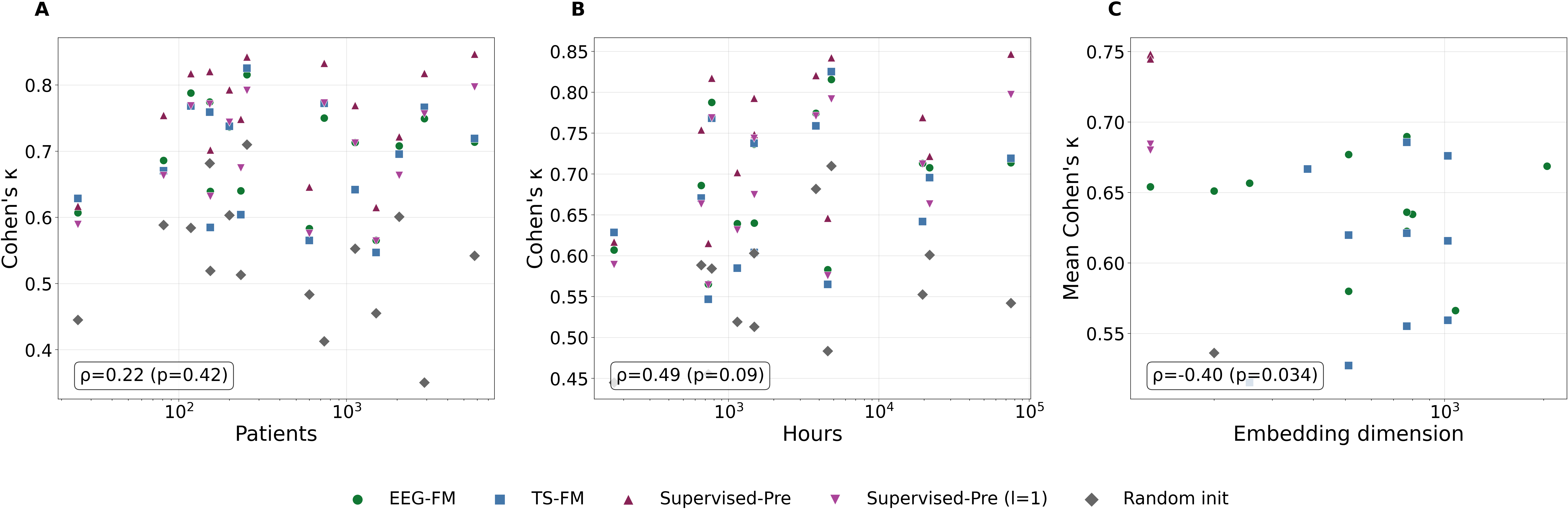}
    \caption{\textbf{Performance is not related to dataset size or embedding dimension.} \textbf{(a), (b)} Cohen's $\kappa$ sleep staging agreement for best performing model per model category, i.e., EEG-FMs, TS-FMs, and Supervised-Pre models, with and without temporal context (sequence length l=1), and a random-initialized reference, is shown. \textbf{(a)} Dataset size in number of patients. \textbf{(b)} Dataset size in number of hours. \textbf{(c)} Mean Cohen's $\kappa$ over datasets and embedding dimension.  }
    \label{fig:DatasetCharacteristics}
\end{figure}

Sleep staging performance per model and dataset is presented in Figure \ref{fig:SleepOverview}. 
Despite substantial variability across datasets, NeuroLM, REVE, and BIOT consistently rank among the top-3-performing EEG-FMs. While all models evaluate sleep staging as a downstream task, only REVE
incorporates sleep-specific data during pretraining; however, this dataset represents only 34 out of the
24,274 patients included across studies \cite{ouahidi2025reve, jiang2024neurolm}. Together with TS-FMs achieving performance comparable to EEG-FMs without relying on EEG-specific
pretraining, this further supports the claim that such pretraining may be less influential than other design factors.
Within TS-FMs, increasing the number of parameters does not consistently lead to improved performance within a given model family. Furthermore, no clear relationship is observed between performance and either dataset size or embedding dimensionality (Figure \ref{fig:DatasetCharacteristics}). On average, all models, except for the TS-FM Chronos, perform above the random-initialized baseline.

\begin{table}
\centering
\small
\caption{\textbf{Binomial top-3 win test across fifteen sleep-staging cohorts.} For each model, the exact one-sided binomial test asks whether the number of cohorts on which the model ranks in the top~3 (by mean Cohen's $\kappa$) exceeds the chance rate $p_0 = 3/k$. Three pools are tested independently: all $k=28$ probed models, $k=25$ models excluding the Supervised-Pre models using sequence modeling, and the $k=11$ EEG-FMs in isolation. Bold cells indicate significance at $\alpha=0.05$.}
\label{tab:SleepStatistics}
\resizebox{\linewidth}{!}{
\begin{tabular}{l l c c c c c c}
\toprule
 & & \multicolumn{2}{c}{All models ($k\!=\!28$)} & \multicolumn{2}{c}{No Sup.-Pre ($k\!=\!25$)} & \multicolumn{2}{c}{EEG-FM only ($k\!=\!11$)} \\
\cmidrule(lr){3-4}\cmidrule(lr){5-6}\cmidrule(lr){7-8}
Model & Family & Wins & $p$ & Wins & $p$ & Wins & $p$ \\
\multicolumn{8}{l}{\footnotesize $N=15$ cohorts; $p_0 = 3/28 = 0.107$ vs.\ $3/25 = 0.120$ vs.\ $3/11 = 0.273$. Bold = significant at $\alpha=0.05$.} \\
\midrule
CoRe-Sleep & Sup.-Pre & \textbf{15/15} & \textbf{$<\!10^{-15}$} & -- & -- & -- & -- \\
SleepTransformer & Sup.-Pre & \textbf{15/15} & \textbf{$<\!10^{-15}$} & -- & -- & -- & -- \\
SleePyCo & Sup.-Pre & \textbf{13/15} & \textbf{$<\!10^{-11}$} & -- & -- & -- & -- \\
NeuroLM & EEG-FM & 1/15 & 0.817 & \textbf{6/15} & \textbf{0.006} & \textbf{13/15} & \textbf{$<\!10^{-6}$} \\
Moirai-B & TS-FM & 0/15 & 1.000 & \textbf{7/15} & \textbf{$<\!10^{-4}$} & -- & -- \\
CoRe-Sleep (l=1) & Sup.-Pre ($\ell$=1) & 0/15 & 1.000 & \textbf{8/15} & \textbf{$<\!10^{-4}$} & -- & -- \\
Moirai-L & TS-FM & 0/15 & 1.000 & 4/15 & 0.096 & -- & -- \\
SleepTransformer (l=1) & Sup.-Pre ($\ell$=1) & 0/15 & 1.000 & 4/15 & 0.096 & -- & -- \\
SleePyCo (l=1) & Sup.-Pre ($\ell$=1) & 0/15 & 1.000 & \textbf{6/15} & \textbf{0.006} & -- & -- \\
REVE & EEG-FM & 0/15 & 1.000 & 3/15 & 0.265 & \textbf{10/15} & \textbf{0.002} \\
Moirai-S & TS-FM & 0/15 & 1.000 & 1/15 & 0.853 & -- & -- \\
BIOT & EEG-FM & 0/15 & 1.000 & 4/15 & 0.096 & \textbf{8/15} & \textbf{0.029} \\
EEGPT & EEG-FM & 0/15 & 1.000 & 0/15 & 1.000 & 6/15 & 0.203 \\
SleepFM & EEG-FM & 0/15 & 1.000 & 1/15 & 0.853 & 2/15 & 0.944 \\
CBraMod & EEG-FM & 1/15 & 0.817 & 1/15 & 0.853 & 2/15 & 0.944 \\
STEEGFormer-B & EEG-FM & 0/15 & 1.000 & 0/15 & 1.000 & 3/15 & 0.820 \\
NeuroRVQ & EEG-FM & 0/15 & 1.000 & 0/15 & 1.000 & 1/15 & 0.992 \\
MOMENT-B & TS-FM & 0/15 & 1.000 & 0/15 & 1.000 & -- & -- \\
MOMENT-S & TS-FM & 0/15 & 1.000 & 0/15 & 1.000 & -- & -- \\
LaBraM & EEG-FM & 0/15 & 1.000 & 0/15 & 1.000 & 0/15 & 1.000 \\
MOMENT-L & TS-FM & 0/15 & 1.000 & 0/15 & 1.000 & -- & -- \\
STEEGFormer-S & EEG-FM & 0/15 & 1.000 & 0/15 & 1.000 & 0/15 & 1.000 \\
NeuroGPT & EEG-FM & 0/15 & 1.000 & 0/15 & 1.000 & 0/15 & 1.000 \\
Chronos-L & TS-FM & 0/15 & 1.000 & 0/15 & 1.000 & -- & -- \\
CBraMod-random & Random-Init & 0/15 & 1.000 & 0/15 & 1.000 & -- & -- \\
Chronos-B & TS-FM & 0/15 & 1.000 & 0/15 & 1.000 & -- & -- \\
Chronos-S & TS-FM & 0/15 & 1.000 & 0/15 & 1.000 & -- & -- \\
Chronos-T & TS-FM & 0/15 & 1.000 & 0/15 & 1.000 & -- & -- \\
\bottomrule
\end{tabular}}
\end{table}

\begin{figure}
    \centering
    \includegraphics[width=\textwidth]{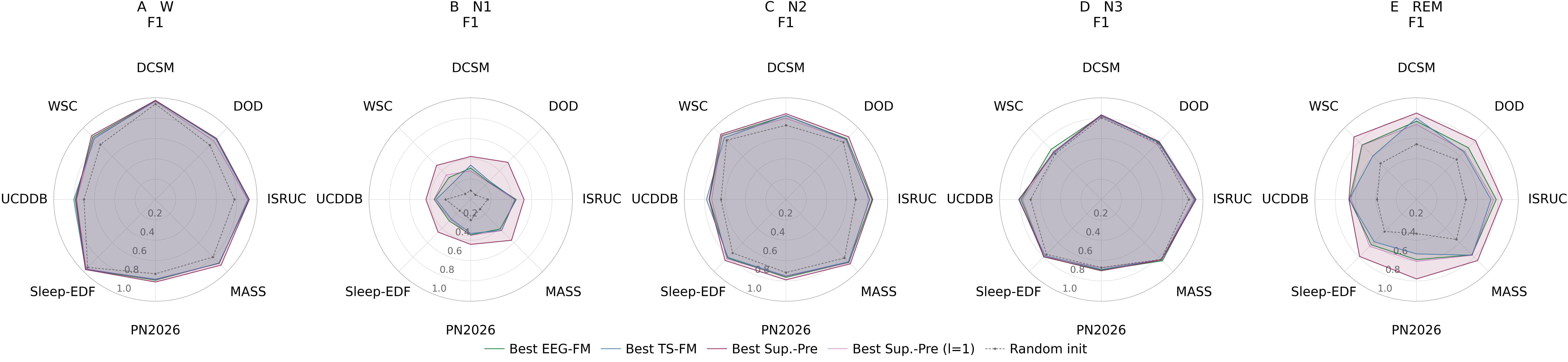}
    \caption{\textbf{Sleep staging performance is lowest for N1 and REM.} F1-score per sleep stage for best performing model per category, i.e., EEG-FMs, TS-FMs, and Supervised-Pre models, with and without temporal context (sequence length l=1), and a random-initialized reference.}
    \label{fig:SleepPerStage}
\end{figure}

Per-stage performance (Figure \ref{fig:SleepPerStage}) indicates that wake is the most reliably classified stage across models, whereas N1 and REM are most frequently misclassified. While reduced performance on N1 is consistent with its low inter-rater reliability among human scorers, REM sleep is generally considered less ambiguous \cite{lee_interrater_2022}. The discrepancy in REM classification performance may be attributed to the reliance on EEG-only inputs, whereas the AASM scoring criteria incorporate additional modalities, such as EOG and EMG, for REM identification \cite{aasm2023}. Moreover, supervised models derive notable benefits from incorporating inter-epoch temporal context, particularly for N1 and REM stages, as AASM guidelines rely on information from adjacent epochs for accurate staging \cite{aasm2023}.

\begin{figure}
    \centering
    \includegraphics[width=\textwidth]{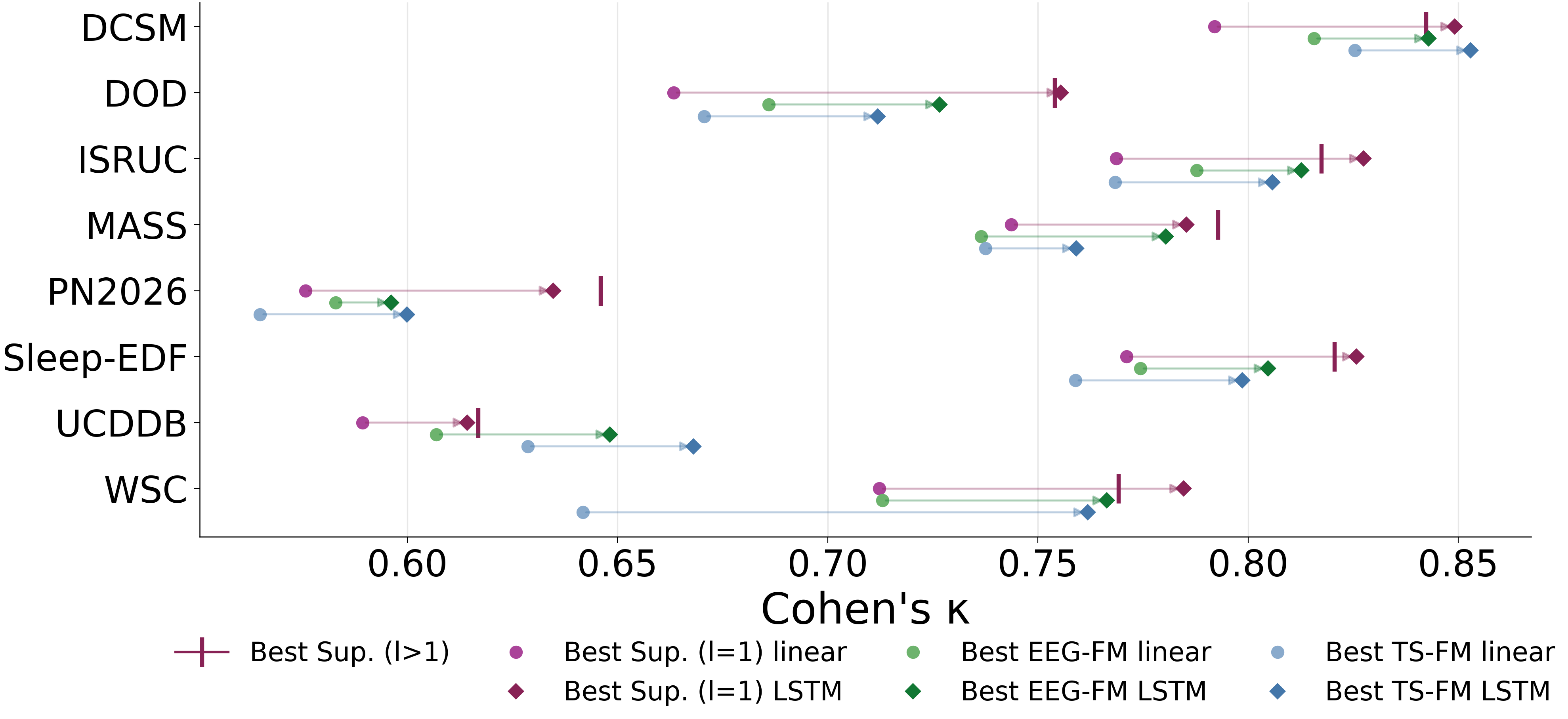}
    \caption{\textbf{LSTM probing provides temporal context to improve sleep staging.} Cohen's $\kappa$ sleep staging agreement for best performing model per model category, under linear and LSTM probing. Supervised models which encode temporal context via sequence modeling are shown as a baseline.}
    \label{fig:LSTM}
\end{figure}

 To further investigate the importance of sequence modeling, we applied a bidirectional LSTM probe with sequence length 20, giving the FMs access to temporal context. As a control, we compared supervised models with temporal context under a linear probe against supervised models with sequence length l=1 under an LSTM probe (Fig. \ref{fig:LSTM}). Their similar performance indicates that the LSTM probe alone is sufficient to supply the missing temporal context. Under the LSTM probe, the gap between FMs and supervised models is largely closed, with FMs surpassing supervised baselines on some datasets. This confirms the importance of temporal context in sleep staging, while task-specific models retain an overall advantage.

\begin{figure}
    \centering
    \includegraphics[width=\textwidth]{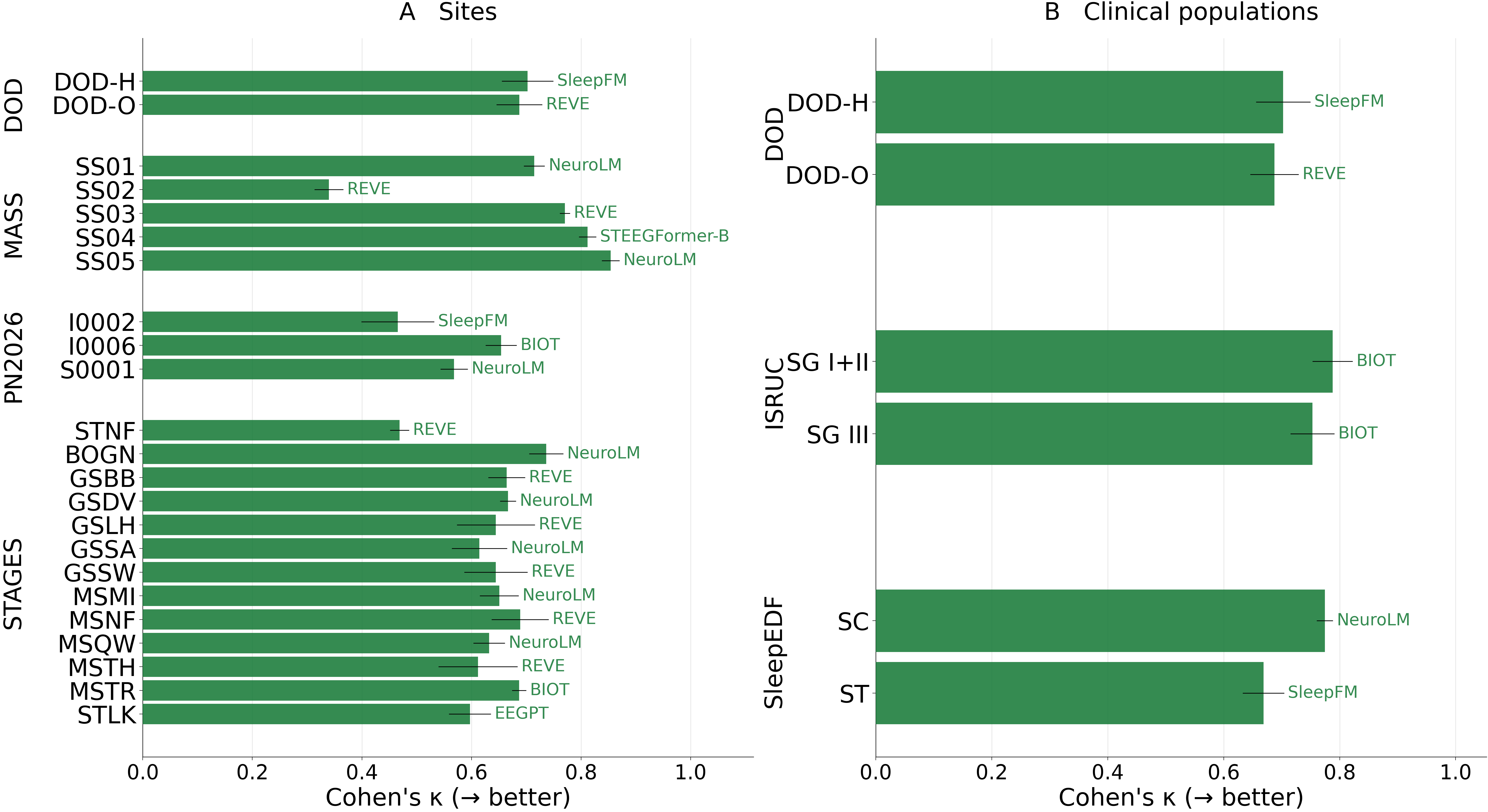}
    \caption{Cohen's $\kappa$ sleep staging agreement for best performing EEG-FM per subset. \textbf{(a)} \textbf{Subsets recorded at different sites differ significantly due to differing acquisition protocols.} \textbf{(b)} \textbf{Clinical cohorts are not systematically associated with lower performance.} DOD-O contains patients with OSA, DOD-H contains healthy controls. ISRUC Subgroup I and II contain patients with sleep disorders, Subgroup III contains healthy people. Parkinson's PD contains patients with Parkinson's disease, HOA contains healthy older adults. Sleep-EDF Cassette (SC) contains healthy patients, Telemetry (ST) contains medicated people.}
    \label{fig:SleepSubsets}
\end{figure}

Sleep staging performance was further evaluated across dataset subsets by training a separate linear probe for each subset (Figure \ref{fig:SleepSubsets}). Subsets recorded at different sites can show large variability, as acquisition protocols can differ in e.g., device and channels. Contrary to expectations, clinical populations, i.e., including patients with disorders or experimental interventions, were not consistently more difficult to stage than healthy populations.

\begin{figure}
    \centering
    \includegraphics[width=0.7\textwidth]{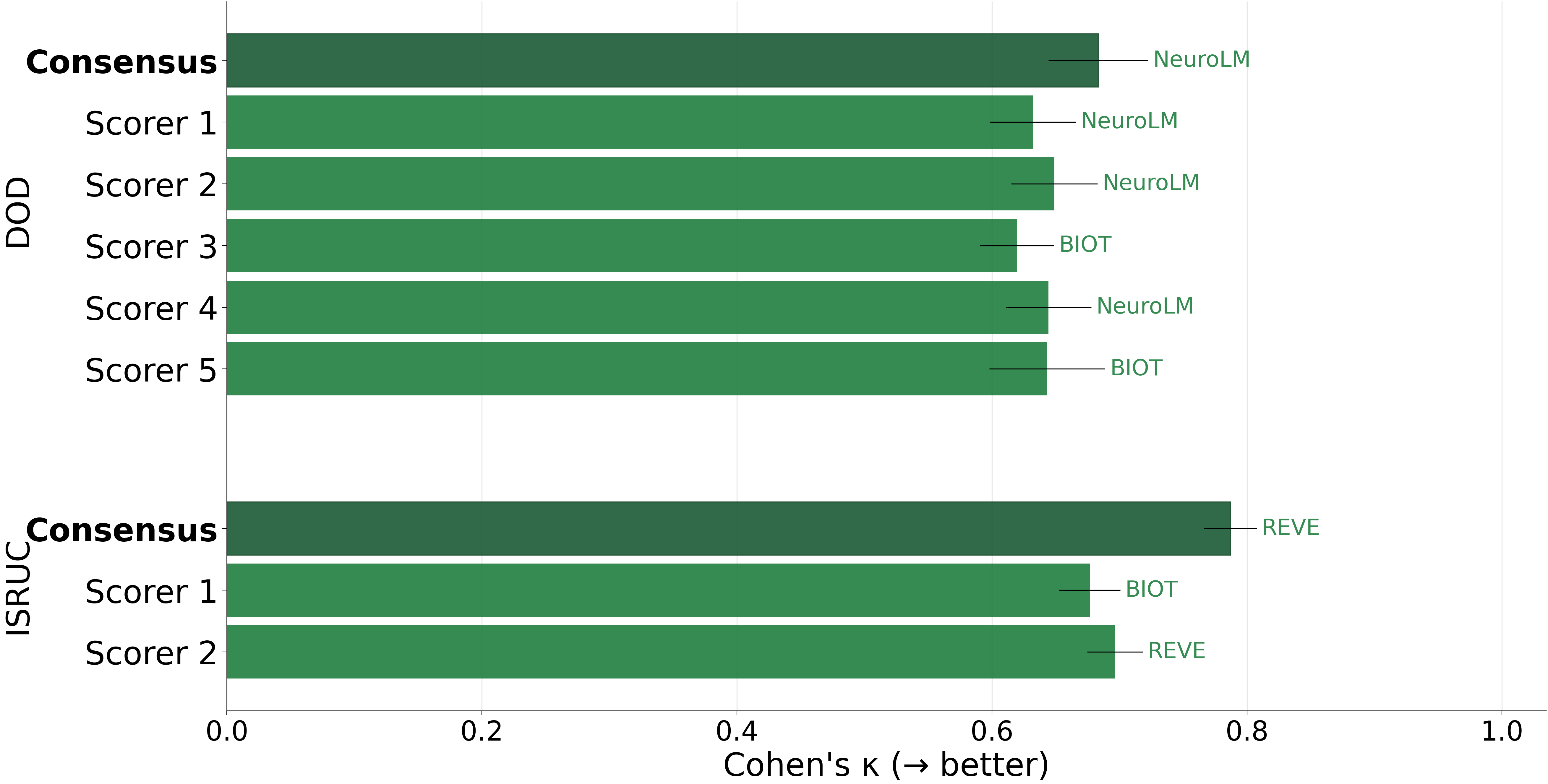}
    \caption{\textbf{Sleep staging agreement is highest with consensus scoring, as ambiguous epochs are excluded.} Cohen's $\kappa$ sleep staging agreement with ground-truth labels scored by different experts, and consensus scoring for best performing EEG-FM.}
    \label{fig:SleepScorers}
\end{figure}

In datasets with annotations from multiple experts, sleep staging agreement was consistently higher when evaluated against consensus scoring (Figure \ref{fig:SleepScorers}). This aligns with expectations, as consensus labels tend to exclude ambiguous epochs, which are also considered hard to stage automatically.

\begin{figure}
    \centering
    \includegraphics[width=\textwidth]{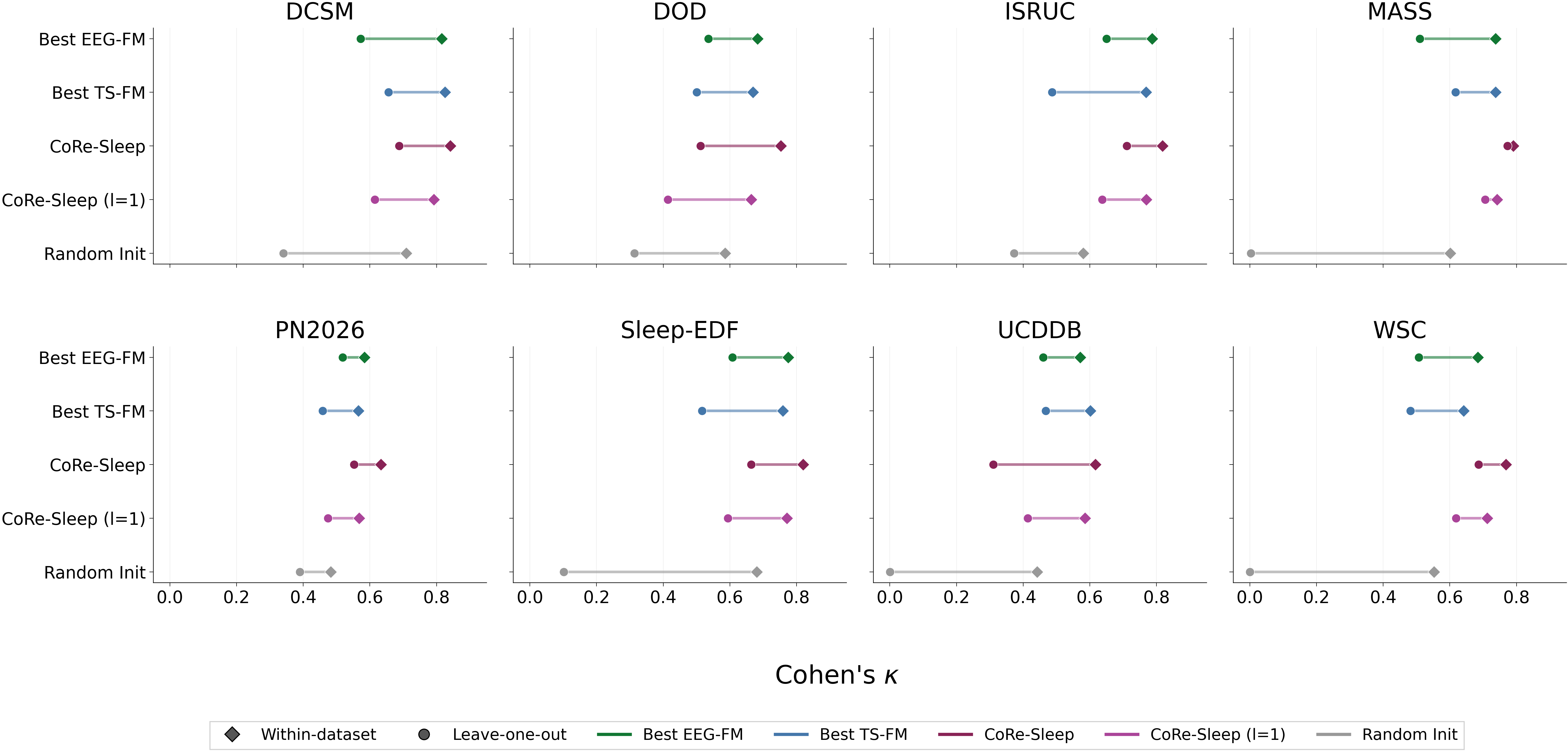}
    \caption{\textbf{Cross-dataset evaluation substantially degrades performance.} Cohen's $\kappa$ sleep staging agreement comparing within-dataset CV performance and leave-one-dataset-out performance, for best performing model per category, i.e., EEG-FMs, TS-FMs, and CoRe-Sleep, with and without temporal context (sequence length l=1), and a random-initialized reference.}
    \label{fig:SleepScorers_LODO}
\end{figure}

Lastly, we evaluate whether we can build a fully generalized model requiring no labeled data from the target dataset. We thus perform leave-one-dataset-out (LODO) linear probing, fitting the probe on the pooled remaining cohorts and testing on the held-out one (Figure \ref{fig:SleepScorers_LODO}). Performance drops substantially for every model, with the largest drops among EEG-FMs (REVE −0.35, NeuroLM −0.31 in $\kappa$) and supervised models degrading least. The family-level picture is reinforced: EEG-FMs do not outperform their supervised counterparts, and TS-FMs and EEG-FMs perform on par.

\subsubsection{Hypnogram features}
\begin{figure}
    \centering
    \includegraphics[width=\textwidth]{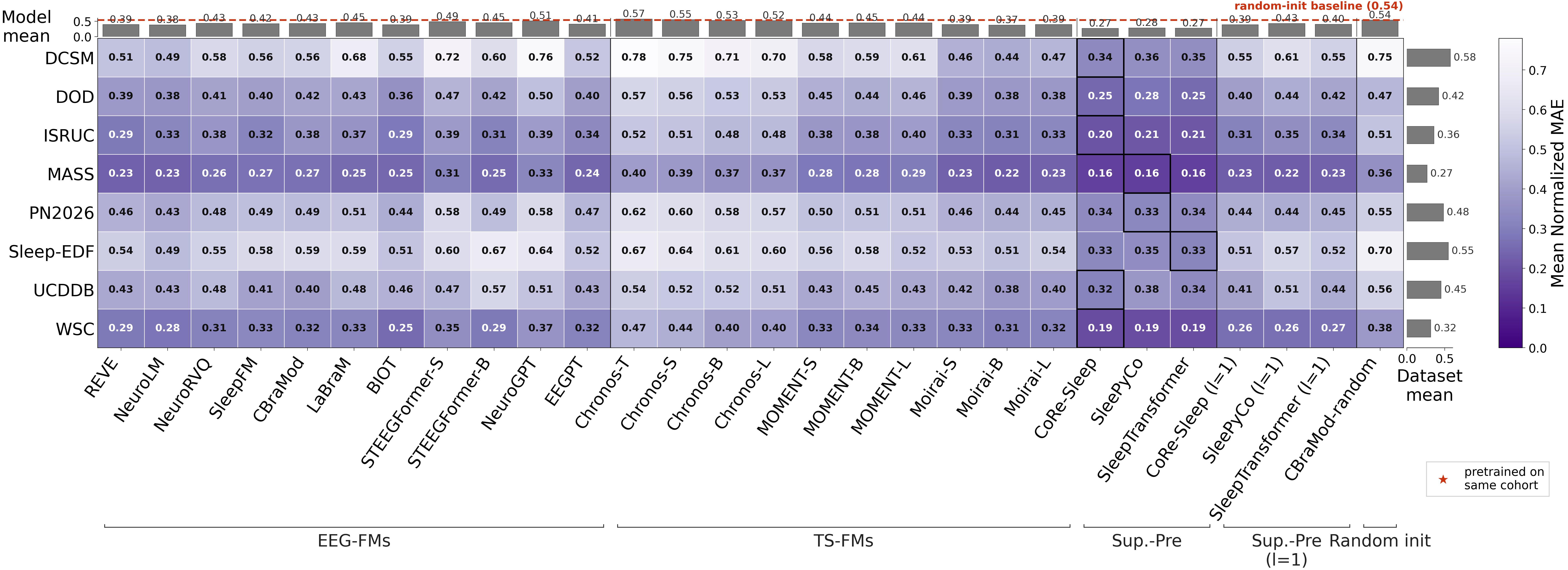}
    \caption{\textbf{Sleep staging and hypnogram feature performance are not consistently aligned across datasets.} Normalized Mean Absolute Error (MAE) between ground truth and predicted hypnogram features for all datasets and models per model category, i.e., EEG-FMs, TS-FMs, and Supervised-Pre models, with and without temporal context (sequence length l=1), and a random-initialized reference. Black borders mark the cohort best-performing model. Red stars flag pretraining/evaluation overlap. Grey bars show mean performance over dataset and models respectively, with the red dashed line indicating mean random-initialized baseline. Missing results are indicated with a cross.}
    \label{fig:SleepMetric1}
\end{figure}

Hypnogram-derived features used for downstream clinical diagnosis and analysis are presented in Figure \ref{fig:SleepMetric1}. Similar to sleep staging, substantial variability is observed across datasets. However, datasets with strong staging performance do not necessarily achieve the best performance on hypnogram features. In addition, supervised models benefit from incorporating inter-epoch temporal context through their sequence-based architecture.

\subsubsection{Event Detection}
Similar to sleep staging, supervised models outperform FMs in microarousal detection, suggesting that FMs do not adequately encode these events, although all model categories exceed the chance baseline set by prevalence. Performance is more limited for respiratory events and limb movements. As these are primarily defined using other modalities (e.g., respiratory signals and EMG), their representation in EEG embeddings is less essential. Although event detection focuses on identifying discrete events rather than performing longitudinal staging, the inter-epoch temporal context captured by supervised models still provides a modest benefit.

\begin{figure}
    \centering
    \includegraphics[width=\textwidth]{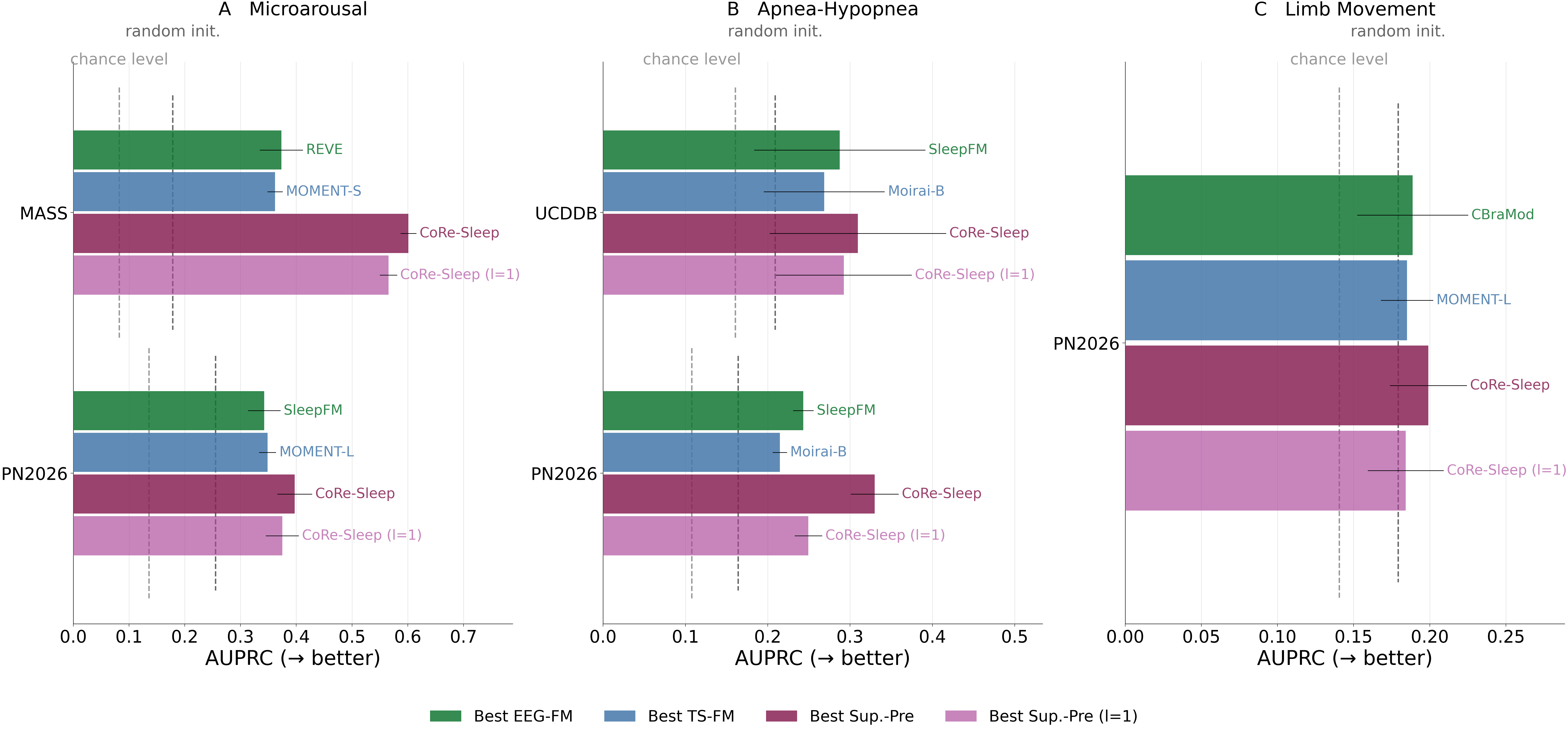}
    \caption{\textbf{FMs fail to adequately capture microarousals in their embeddings.} Area under the precision-recall curve (AUPRC) for sleep event detection for best performing model per category. Prevalence indicates chance baseline. \textbf{(a)} Microarousals with a minimal duration of 3 seconds. \textbf{(b)} Apnea and hypopnea events with a minimal duration of 10 seconds. \textbf{(c)} Periodic limb movement with a minimal duration of 0.5 seconds.
}
    \label{fig:SleepEvents}
\end{figure}

\subsubsection{Cognitive Impairment Diagnosis}
Recent studies have highlighted associations between sleep and cognitive impairment, and demonstrated that cognitive impairment can be detected in EEG recordings \cite{li_evaluating_2022, Busch2024-om_cognitive}. However, the models evaluated on cognitive impairment detection in the PhysioNet 2026 dataset achieved only modest performance (AUROC around 0.60), suggesting that the extracted representations do not sufficiently encode features relevant for reliable detection of cognitive impairment.

\begin{figure}
    \centering
    \includegraphics[width=0.5\textwidth]{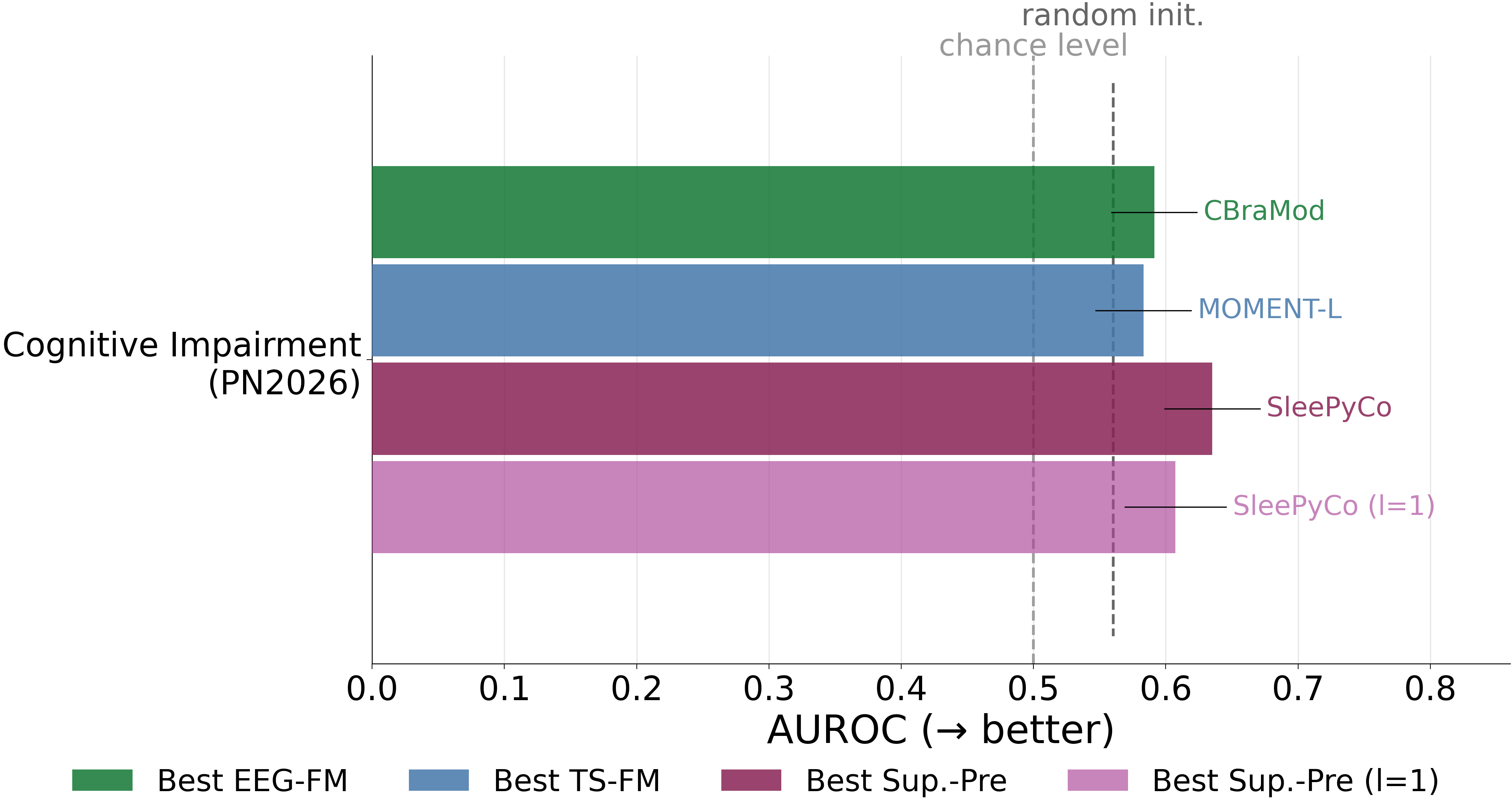}
    \caption{\textbf{Models insufficiently encode cognitive impairment for reliable detection.} Area under the receiver-operator curve (AUROC) for diagnosis of cognitive impairment in the PhysioNet 2026 dataset for best performing model per model category, i.e., EEG-FMs, TS-FMs, and Supervised-Pre models, with and without temporal context (sequence length l=1), and a random-initialized reference, is shown.}
    \label{fig:SleepCI}
\end{figure}



\subsection{Brain Age} \label{Appendix:Results-BrainAge}

\newcommand{\bares}[3]{#1\,{\fontsize{4.5}{5}\selectfont$\pm$ #2}\,{\fontsize{4.5}{5}\selectfont\textcolor{gray}{(#3)}}}
\newcommand{\babestres}[3]{\textbf{#1}\,{\fontsize{4.5}{5}\selectfont$\pm$ #2}\,{\fontsize{4.5}{5}\selectfont\textcolor{gray}{(#3)}}}
\newcommand{\basecres}[3]{\underline{#1}\,{\fontsize{4.5}{5}\selectfont$\pm$ #2}\,{\fontsize{4.5}{5}\selectfont\textcolor{gray}{(#3)}}}
\newcommand{\babase}[2]{#1\,{\fontsize{4.5}{5}\selectfont$\pm$ #2}}
\newcommand{\bamissing}{---}
\newcommand{\bagroup}[2]{\multicolumn{#1}{l}{\textcolor{gray}{\textbf{#2}}}\\}

\paragraph{Cohort- and model-family-level brain-age results.}
Detailed cohort- and model-family-level results are reported in Tables~\ref{tab:brain_age_full}, \ref{tab:brain_age_pearson_r}, \ref{tab:brain_age_guido}. Across most cohorts and metrics, the supervised feature-based age-prediction model remains the strongest reference. One exception occurs in cohorts with a restricted age range, such as MrOS and WSC, where its advantage is reduced, suggesting that cohort-specific age distributions partly modulate apparent performance differences. Within each model family, performance also varies across datasets and metrics.

\begin{table*}[ht]
\centering
\tiny
\setlength{\tabcolsep}{1.2pt}
\renewcommand{\arraystretch}{0.9}
\resizebox{\textwidth}{!}{%
\begin{tabular}{lcccccc}
\toprule
Model & PN2026 & WSC & ISRUC & SleepEDF SC & CFS & MrOS \\
\midrule
\bagroup{7}{EEG-FM}
BIOT & \bares{6.94}{0.60}{-0.72} & \bares{5.94}{0.21}{-0.50} & \bares{11.27}{1.25}{-2.51} & \basecres{11.98}{1.69}{-6.35} & \bares{9.38}{0.58}{-6.99} & \bares{3.99}{0.04}{-0.63} \\
CBraMod & \bares{7.57}{1.13}{-0.08} & \bares{5.88}{0.25}{-0.56} & \bares{11.78}{2.19}{-2.00} & \bares{15.55}{2.03}{-2.79} & \bares{9.53}{0.33}{-6.84} & \bares{4.22}{0.02}{-0.41} \\
LaBraM & \bares{7.47}{0.43}{-0.19} & \bares{5.81}{0.20}{-0.63} & \bares{10.64}{0.90}{-3.14} & \bares{15.26}{1.96}{-3.07} & \bares{9.05}{0.28}{-7.32} & \bares{4.21}{0.08}{-0.42} \\
NeuroLM-VQ & \bares{7.22}{0.57}{-0.44} & \bares{5.53}{0.18}{-0.91} & \bares{10.95}{1.22}{-2.83} & \bares{13.13}{1.88}{-5.21} & \bares{7.78}{0.43}{-8.59} & \bares{3.86}{0.10}{-0.76} \\
REVE & \basecres{6.72}{0.27}{-0.93} & \bares{5.83}{0.24}{-0.61} & \bares{11.24}{2.00}{-2.54} & \bares{14.33}{3.05}{-4.00} & \bares{8.81}{0.41}{-7.56} & \bares{3.85}{0.06}{-0.78} \\
SleepFM & \bares{7.28}{0.34}{-0.37} & \bares{5.78}{0.33}{-0.67} & \bares{9.40}{1.25}{-4.39} & \bares{12.73}{1.09}{-5.60} & \bares{8.81}{0.91}{-7.56} & \bares{3.95}{0.08}{-0.68} \\
EEGPT & \bares{7.54}{0.43}{-0.11} & \babestres{5.17}{0.20}{-1.27} & \basecres{8.99}{1.87}{-4.80} & \bares{14.31}{2.79}{-4.02} & \basecres{7.67}{0.58}{-8.70} & \bares{3.89}{0.11}{-0.74} \\
NeuroRVQ-EEG & \bares{7.06}{1.02}{-0.60} & \bares{5.34}{0.17}{-1.10} & \bares{11.94}{1.48}{-1.84} & \bares{12.34}{2.61}{-6.00} & \bares{8.12}{0.62}{-8.25} & \basecres{3.81}{0.10}{-0.81} \\
Neuro-GPT & \bares{7.23}{0.51}{-0.42} & \bares{6.06}{0.18}{-0.38} & \bares{12.60}{1.89}{-1.19} & \bares{14.56}{1.66}{-3.78} & \bares{9.65}{0.47}{-6.72} & \bares{4.20}{0.08}{-0.42} \\
STEEGFormer-S & \bares{7.48}{0.84}{-0.18} & \bares{5.67}{0.23}{-0.77} & \bares{11.79}{1.16}{-1.99} & \bares{15.23}{1.85}{-3.11} & \bares{8.36}{0.69}{-8.01} & \bares{3.93}{0.04}{-0.70} \\
STEEGFormer-B & \bares{7.23}{1.03}{-0.43} & \bares{5.56}{0.23}{-0.88} & \bares{10.82}{1.11}{-2.97} & \bares{14.24}{1.27}{-4.09} & \bares{8.26}{0.34}{-8.11} & \bares{3.86}{0.03}{-0.76} \\
\midrule
\bagroup{7}{Random-init}
CBraMod (random init) & \bares{7.07}{0.59}{-0.58} & \bares{5.83}{0.22}{-0.61} & \bares{11.54}{2.05}{-2.24} & \bares{12.30}{2.04}{-6.04} & \bares{11.23}{0.57}{-5.14} & \bares{4.60}{0.03}{-0.03} \\
\midrule
\bagroup{7}{TS-FMs}
Chronos-T5 (tiny) & \bares{7.68}{0.94}{+0.03} & \bares{5.88}{0.23}{-0.57} & \bares{10.91}{1.58}{-2.88} & \basecres{11.69}{0.90}{-6.64} & \bares{9.44}{0.64}{-6.93} & \bares{4.08}{0.06}{-0.55} \\
Chronos-T5 (small) & \bares{7.46}{0.73}{-0.19} & \bares{5.84}{0.25}{-0.60} & \bares{11.34}{1.46}{-2.44} & \bares{13.33}{1.43}{-5.01} & \bares{8.74}{0.37}{-7.63} & \bares{4.16}{0.04}{-0.47} \\
Chronos-T5 (base) & \bares{7.66}{0.80}{+0.00} & \bares{5.75}{0.23}{-0.69} & \bares{10.77}{1.27}{-3.01} & \bares{13.28}{2.81}{-5.06} & \bares{8.66}{0.72}{-7.71} & \bares{4.13}{0.03}{-0.50} \\
Chronos-T5 (large) & \bares{7.63}{0.78}{-0.03} & \bares{5.73}{0.24}{-0.72} & \bares{12.15}{1.97}{-1.63} & \bares{12.71}{1.85}{-5.63} & \bares{8.77}{0.52}{-7.59} & \bares{4.14}{0.01}{-0.49} \\
Moirai (small) & \bares{7.13}{0.69}{-0.53} & \bares{5.78}{0.30}{-0.67} & \basecres{9.13}{1.57}{-4.65} & \bares{14.04}{2.60}{-4.29} & \bares{8.89}{0.40}{-7.48} & \bares{4.07}{0.16}{-0.56} \\
Moirai (base) & \bares{7.10}{0.69}{-0.56} & \bares{5.75}{0.32}{-0.69} & \bares{10.22}{1.49}{-3.57} & \bares{14.03}{2.18}{-4.30} & \bares{8.02}{0.18}{-8.35} & \bares{4.17}{0.27}{-0.46} \\
Moirai (large) & \basecres{7.08}{0.69}{-0.58} & \bares{5.88}{0.32}{-0.57} & \bares{10.03}{1.57}{-3.75} & \bares{14.07}{3.18}{-4.26} & \bares{9.15}{0.46}{-7.22} & \bares{4.13}{0.05}{-0.50} \\
MOMENT (small) & \bares{7.11}{0.67}{-0.55} & \bares{5.79}{0.18}{-0.65} & \bares{10.27}{1.20}{-3.51} & \bares{14.30}{2.07}{-4.03} & \bares{8.31}{0.31}{-8.06} & \bares{3.99}{0.05}{-0.64} \\
MOMENT (base) & \bares{7.26}{0.59}{-0.40} & \basecres{5.70}{0.13}{-0.74} & \bares{10.20}{0.98}{-3.58} & \bares{12.62}{2.54}{-5.72} & \basecres{7.99}{0.57}{-8.38} & \babestres{3.71}{0.08}{-0.92} \\
MOMENT (large) & \bares{7.33}{0.63}{-0.32} & \bares{5.90}{0.19}{-0.55} & \bares{11.27}{1.93}{-2.52} & \bares{13.48}{2.26}{-4.86} & \bares{8.28}{0.27}{-8.09} & \bares{3.99}{0.04}{-0.64} \\
\midrule
\bagroup{7}{Supervised-Pre Sleep}
CoreSleep & \bares{7.15}{0.65}{-0.51} & \bares{5.43}{0.19}{-1.02} & \basecres{8.80}{1.42}{-4.98} & \basecres{10.39}{1.07}{-7.94} & \bares{8.17}{0.32}{-8.20} & \basecres{3.89}{0.09}{-0.74} \\
SleepTransformer & \bares{7.15}{0.46}{-0.51} & \basecres{5.43}{0.14}{-1.02} & \bares{10.30}{1.32}{-3.49} & \bares{11.75}{2.67}{-6.58} & \basecres{7.87}{0.30}{-8.50} & \bares{3.93}{0.06}{-0.70} \\
SleepyCo & \basecres{6.88}{0.67}{-0.78} & \bares{5.51}{0.20}{-0.94} & \bares{10.43}{1.58}{-3.35} & \bares{13.02}{2.51}{-5.31} & \bares{8.01}{0.48}{-8.35} & \bares{3.93}{0.05}{-0.70} \\
\midrule
\bagroup{7}{Supervised-Pre Age}
Sun et al. (2019) & \babestres{4.86}{0.48}{-2.79} & \bares{5.65}{0.20}{-0.79} & \babestres{5.54}{0.70}{-8.25} & \babestres{6.09}{0.89}{-12.25} & \babestres{6.22}{0.40}{-10.15} & \bares{4.98}{0.19}{+0.35} \\
\midrule
\bagroup{7}{Predict train-mean}
Predict train-mean & \babase{7.66}{0.82} & \babase{6.44}{0.19} & \babase{13.78}{1.01} & \babase{18.33}{0.40} & \babase{16.37}{0.59} & \babase{4.63}{0.02} \\
\bottomrule
\end{tabular}
}
\caption{Overall brain-age prediction performance. Cells report CV-test MAE in years, shown as mean $\pm$ standard deviation across five outer folds. Gray parentheses show the difference relative to the predict-train-mean baseline. The best non-baseline MAE per dataset is bolded, and the second-best is underlined.}
\label{tab:brain_age_full}
\end{table*}

.

\begin{table*}[ht]
\centering
\tiny
\setlength{\tabcolsep}{1.2pt}
\renewcommand{\arraystretch}{0.9}
\resizebox{\textwidth}{!}{%
\begin{tabular}{lcccccc}
\toprule
Model & PN2026 & WSC & ISRUC & SleepEDF SC & CFS & MrOS \\
\midrule
\bagroup{7}{EEG-FM}
BIOT & \babase{0.407}{0.126} & \babase{0.393}{0.037} & \babase{0.472}{0.112} & \babase{\underline{0.806}}{0.087} & \babase{0.782}{0.029} & \babase{0.473}{0.012} \\
CBraMod & \babase{0.294}{0.110} & \babase{0.407}{0.032} & \babase{0.477}{0.207} & \babase{0.566}{0.110} & \babase{0.779}{0.015} & \babase{0.395}{0.011} \\
LaBraM & \babase{0.257}{0.172} & \babase{0.409}{0.061} & \babase{0.552}{0.091} & \babase{0.587}{0.112} & \babase{0.803}{0.021} & \babase{0.380}{0.056} \\
NeuroLM-VQ & \babase{0.400}{0.108} & \babase{0.498}{0.045} & \babase{0.556}{0.079} & \babase{0.731}{0.109} & \babase{\underline{0.855}}{0.023} & \babase{0.488}{0.062} \\
REVE & \babase{\underline{0.458}}{0.090} & \babase{0.420}{0.060} & \babase{0.493}{0.131} & \babase{0.681}{0.123} & \babase{0.806}{0.019} & \babase{\underline{0.527}}{0.015} \\
SleepFM & \babase{0.356}{0.135} & \babase{0.431}{0.059} & \babase{0.696}{0.048} & \babase{0.726}{0.067} & \babase{0.815}{0.036} & \babase{0.489}{0.027} \\
EEGPT & \babase{0.252}{0.161} & \babase{\underline{0.646}}{0.019} & \babase{\underline{0.712}}{0.074} & \babase{0.609}{0.142} & \babase{0.855}{0.033} & \babase{0.418}{0.077} \\
NeuroRVQ-EEG & \babase{0.421}{0.114} & \babase{0.624}{0.022} & \babase{0.409}{0.125} & \babase{0.767}{0.079} & \babase{0.840}{0.038} & \babase{0.501}{0.108} \\
Neuro-GPT & \babase{0.287}{0.093} & \babase{0.491}{0.018} & \babase{0.364}{0.185} & \babase{0.637}{0.089} & \babase{0.773}{0.012} & \babase{0.381}{0.034} \\
STEEGFormer-S & \babase{0.264}{0.096} & \babase{0.562}{0.032} & \babase{0.459}{0.121} & \babase{0.605}{0.093} & \babase{0.833}{0.023} & \babase{0.468}{0.035} \\
STEEGFormer-B & \babase{0.339}{0.064} & \babase{0.541}{0.108} & \babase{0.543}{0.092} & \babase{0.687}{0.069} & \babase{0.843}{0.011} & \babase{0.454}{0.066} \\
\midrule
\bagroup{7}{Random-init}
CBraMod (random init) & \babase{0.389}{0.102} & \babase{0.431}{0.031} & \babase{0.419}{0.235} & \babase{0.756}{0.096} & \babase{0.686}{0.048} & \babase{0.126}{0.025} \\
\midrule
\bagroup{7}{TS-FMs}
Chronos-T5 (tiny) & \babase{0.313}{0.082} & \babase{0.423}{0.022} & \babase{0.617}{0.069} & \babase{\underline{0.783}}{0.055} & \babase{0.784}{0.035} & \babase{0.445}{0.022} \\
Chronos-T5 (small) & \babase{0.291}{0.105} & \babase{0.417}{0.042} & \babase{0.517}{0.134} & \babase{0.713}{0.082} & \babase{0.820}{0.013} & \babase{0.410}{0.037} \\
Chronos-T5 (base) & \babase{0.255}{0.105} & \babase{0.449}{0.029} & \babase{0.583}{0.102} & \babase{0.725}{0.066} & \babase{0.824}{0.031} & \babase{0.437}{0.020} \\
Chronos-T5 (large) & \babase{0.249}{0.101} & \babase{0.454}{0.022} & \babase{0.480}{0.152} & \babase{0.731}{0.051} & \babase{0.819}{0.017} & \babase{0.441}{0.015} \\
Moirai (small) & \babase{0.350}{0.175} & \babase{0.442}{0.041} & \babase{\underline{0.694}}{0.122} & \babase{0.642}{0.152} & \babase{0.809}{0.017} & \babase{0.377}{0.144} \\
Moirai (base) & \babase{0.377}{0.130} & \babase{0.454}{0.043} & \babase{0.578}{0.236} & \babase{0.643}{0.128} & \babase{0.848}{0.017} & \babase{0.359}{0.155} \\
Moirai (large) & \babase{0.355}{0.188} & \babase{0.417}{0.040} & \babase{0.611}{0.148} & \babase{0.633}{0.185} & \babase{0.793}{0.038} & \babase{0.392}{0.100} \\
MOMENT (small) & \babase{\underline{0.410}}{0.080} & \babase{0.453}{0.020} & \babase{0.626}{0.056} & \babase{0.642}{0.165} & \babase{0.844}{0.014} & \babase{0.455}{0.041} \\
MOMENT (base) & \babase{0.347}{0.106} & \babase{\underline{0.465}}{0.038} & \babase{0.645}{0.050} & \babase{0.725}{0.129} & \babase{\underline{0.848}}{0.032} & \babase{\underline{0.567}}{0.009} \\
MOMENT (large) & \babase{0.347}{0.107} & \babase{0.414}{0.025} & \babase{0.528}{0.114} & \babase{0.735}{0.082} & \babase{0.841}{0.014} & \babase{0.470}{0.034} \\
\midrule
\bagroup{7}{Supervised-Pre Sleep}
CoreSleep & \babase{0.353}{0.163} & \babase{0.529}{0.022} & \babase{\underline{0.710}}{0.069} & \babase{\underline{0.825}}{0.046} & \babase{0.843}{0.014} & \babase{\underline{0.516}}{0.021} \\
SleepTransformer & \babase{0.379}{0.169} & \babase{\underline{0.530}}{0.028} & \babase{0.604}{0.090} & \babase{0.759}{0.168} & \babase{\underline{0.856}}{0.015} & \babase{0.500}{0.015} \\
SleepyCo & \babase{\underline{0.432}}{0.132} & \babase{0.509}{0.022} & \babase{0.554}{0.157} & \babase{0.809}{0.033} & \babase{0.850}{0.013} & \babase{0.502}{0.022} \\
\midrule
\bagroup{7}{Supervised-Pre Age}
Sun et al. (2019) & \babase{\textbf{0.786}}{0.070} & \babase{\textbf{0.770}}{0.010} & \babase{\textbf{0.922}}{0.010} & \babase{\textbf{0.969}}{0.014} & \babase{\textbf{0.933}}{0.009} & \babase{\textbf{0.586}}{0.026} \\
\bottomrule
\end{tabular}
}
\caption{Overall brain-age prediction Pearson correlation between predicted and chronological age. Cells report cv-test Pearson $r$, shown as mean $\pm$ standard deviation across five outer folds. The best model per dataset is bolded, and the best model within each model family is underlined.}
\label{tab:brain_age_pearson_r}
\end{table*}

\begin{table*}[t]
\centering

\providecommand{\BAMergedCell}[2]{}
\renewcommand{\BAMergedCell}[2]{%
  \ensuremath{#1\,{\scriptstyle\pm\,#2}}%
}
\providecommand{\BAMergedGroup}[1]{}
\renewcommand{\BAMergedGroup}[1]{%
  \multicolumn{9}{@{}l}{\textbf{#1}}\\[1pt]%
}

\scriptsize
\setlength{\tabcolsep}{2.5pt}
\renewcommand{\arraystretch}{1.08}

\begin{adjustbox}{max width=\textwidth}
\begin{tabular}{
    @{}l
    cc@{\hspace{8pt}}
    cc@{\hspace{8pt}}
    cc@{\hspace{8pt}}
    cc@{}
}
\toprule
& \multicolumn{2}{c}{SHHS}
& \multicolumn{2}{c}{STAGES}
& \multicolumn{2}{c}{MESA}
& \multicolumn{2}{c}{HPAP} \\
\cmidrule(lr){2-3}
\cmidrule(lr){4-5}
\cmidrule(lr){6-7}
\cmidrule(lr){8-9}
Model
& MAE $\downarrow$ & $r$ $\uparrow$
& MAE $\downarrow$ & $r$ $\uparrow$
& MAE $\downarrow$ & $r$ $\uparrow$
& MAE $\downarrow$ & $r$ $\uparrow$ \\
\midrule

\BAMergedGroup{EEG-FM}
BIOT
& \BAMergedCell{7.53}{0.21} & \BAMergedCell{0.457}{0.127}
& \BAMergedCell{12.66}{0.68} & \BAMergedCell{0.047}{0.017}
& \BAMergedCell{7.40}{0.14} & \BAMergedCell{0.252}{0.027}
& \BAMergedCell{9.08}{1.03} & \BAMergedCell{0.397}{0.156} \\

CBraMod
& \BAMergedCell{7.53}{0.21} & \BAMergedCell{0.546}{0.022}
& \BAMergedCell{10.69}{0.28} & \BAMergedCell{0.458}{0.060}
& \BAMergedCell{7.15}{0.13} & \BAMergedCell{0.338}{0.029}
& \BAMergedCell{9.24}{0.90} & \BAMergedCell{0.332}{0.131} \\

LaBraM
& \BAMergedCell{7.00}{0.11} & \BAMergedCell{0.615}{0.013}
& \BAMergedCell{10.42}{0.29} & \BAMergedCell{0.493}{0.057}
& \BAMergedCell{6.99}{0.04} & \BAMergedCell{0.393}{0.023}
& \BAMergedCell{9.18}{1.10} & \BAMergedCell{0.387}{0.133} \\

NeuroLM-VQ
& \BAMergedCell{\underline{6.11}}{0.11} & \BAMergedCell{\underline{0.713}}{0.008}
& \BAMergedCell{\underline{8.45}}{0.53} & \BAMergedCell{\underline{0.656}}{0.056}
& \BAMergedCell{\underline{6.15}}{0.17} & \BAMergedCell{\underline{0.563}}{0.022}
& \BAMergedCell{9.28}{2.37} & \BAMergedCell{0.404}{0.158} \\

REVE
& \BAMergedCell{6.70}{0.16} & \BAMergedCell{0.655}{0.014}
& \BAMergedCell{9.77}{0.55} & \BAMergedCell{0.565}{0.067}
& \BAMergedCell{6.81}{0.08} & \BAMergedCell{0.428}{0.018}
& \BAMergedCell{8.91}{1.20} & \BAMergedCell{0.420}{0.125} \\

SleepFM
& \BAMergedCell{6.53}{0.10} & \BAMergedCell{0.675}{0.007}
& \BAMergedCell{10.06}{0.23} & \BAMergedCell{0.542}{0.042}
& \BAMergedCell{6.42}{0.08} & \BAMergedCell{0.509}{0.013}
& \BAMergedCell{8.64}{1.06} & \BAMergedCell{0.424}{0.086} \\

EEGPT
& \BAMergedCell{6.40}{0.13} & \BAMergedCell{0.670}{0.038}
& \BAMergedCell{9.22}{0.16} & \BAMergedCell{0.611}{0.036}
& \BAMergedCell{6.67}{0.28} & \BAMergedCell{0.463}{0.058}
& \BAMergedCell{9.15}{0.49} & \BAMergedCell{0.392}{0.088} \\

NeuroRVQ-EEG
& \BAMergedCell{6.21}{0.14} & \BAMergedCell{0.703}{0.012}
& \BAMergedCell{10.32}{0.44} & \BAMergedCell{0.517}{0.066}
& \BAMergedCell{6.77}{0.20} & \BAMergedCell{0.436}{0.023}
& \BAMergedCell{\underline{8.31}}{0.43} & \BAMergedCell{\underline{0.500}}{0.152} \\

Neuro-GPT
& \BAMergedCell{7.87}{0.15} & \BAMergedCell{0.482}{0.027}
& \BAMergedCell{11.77}{0.29} & \BAMergedCell{0.296}{0.054}
& \BAMergedCell{7.65}{0.09} & \BAMergedCell{0.136}{0.043}
& \BAMergedCell{9.11}{0.97} & \BAMergedCell{0.358}{0.120} \\

STEEGFormer-S
& \BAMergedCell{6.99}{0.13} & \BAMergedCell{0.622}{0.014}
& \BAMergedCell{11.23}{0.32} & \BAMergedCell{0.387}{0.028}
& \BAMergedCell{7.62}{0.08} & \BAMergedCell{0.153}{0.034}
& \BAMergedCell{8.88}{1.05} & \BAMergedCell{0.436}{0.052} \\

STEEGFormer-B
& \BAMergedCell{7.00}{0.15} & \BAMergedCell{0.617}{0.010}
& \BAMergedCell{10.81}{0.27} & \BAMergedCell{0.434}{0.046}
& \BAMergedCell{7.50}{0.07} & \BAMergedCell{0.210}{0.036}
& \BAMergedCell{8.96}{0.82} & \BAMergedCell{0.406}{0.018} \\

\midrule
\BAMergedGroup{Random-init}
CBraMod (random init)
& \BAMergedCell{7.37}{0.10} & \BAMergedCell{0.565}{0.018}
& \BAMergedCell{11.54}{0.24} & \BAMergedCell{0.339}{0.032}
& \BAMergedCell{6.98}{0.19} & \BAMergedCell{0.395}{0.036}
& \BAMergedCell{8.98}{1.24} & \BAMergedCell{0.420}{0.152} \\

\midrule
\BAMergedGroup{TS-FMs}
Chronos-T5 (tiny)
& \BAMergedCell{6.84}{0.08} & \BAMergedCell{0.628}{0.008}
& \BAMergedCell{10.78}{0.33} & \BAMergedCell{0.456}{0.044}
& \BAMergedCell{6.71}{0.11} & \BAMergedCell{0.460}{0.021}
& \BAMergedCell{9.28}{0.94} & \BAMergedCell{0.342}{0.075} \\

Chronos-T5 (small)
& \BAMergedCell{6.66}{0.10} & \BAMergedCell{0.654}{0.014}
& \BAMergedCell{10.49}{0.20} & \BAMergedCell{0.483}{0.035}
& \BAMergedCell{6.55}{0.12} & \BAMergedCell{0.485}{0.026}
& \BAMergedCell{9.13}{0.88} & \BAMergedCell{0.379}{0.077} \\

Chronos-T5 (base)
& \BAMergedCell{6.22}{0.11} & \BAMergedCell{0.693}{0.030}
& \BAMergedCell{10.47}{0.30} & \BAMergedCell{0.492}{0.053}
& \BAMergedCell{6.53}{0.05} & \BAMergedCell{0.494}{0.021}
& \BAMergedCell{8.92}{0.61} & \BAMergedCell{0.390}{0.102} \\

Chronos-T5 (large)
& \BAMergedCell{6.21}{0.17} & \BAMergedCell{0.686}{0.041}
& \BAMergedCell{10.06}{0.26} & \BAMergedCell{0.541}{0.020}
& \BAMergedCell{6.43}{0.17} & \BAMergedCell{0.515}{0.019}
& \BAMergedCell{8.90}{0.65} & \BAMergedCell{0.386}{0.097} \\

Moirai (small)
& \BAMergedCell{6.68}{0.22} & \BAMergedCell{0.619}{0.061}
& \BAMergedCell{10.01}{0.13} & \BAMergedCell{0.541}{0.040}
& \BAMergedCell{6.70}{0.10} & \BAMergedCell{0.463}{0.035}
& \BAMergedCell{8.90}{0.71} & \BAMergedCell{0.446}{0.092} \\

Moirai (base)
& \BAMergedCell{6.31}{0.23} & \BAMergedCell{0.632}{0.147}
& \BAMergedCell{9.46}{0.26} & \BAMergedCell{0.597}{0.041}
& \BAMergedCell{6.28}{0.08} & \BAMergedCell{0.541}{0.027}
& \BAMergedCell{8.81}{0.63} & \BAMergedCell{0.449}{0.099} \\

Moirai (large)
& \BAMergedCell{6.52}{0.15} & \BAMergedCell{0.647}{0.041}
& \BAMergedCell{9.94}{0.17} & \BAMergedCell{0.538}{0.056}
& \BAMergedCell{6.55}{0.10} & \BAMergedCell{0.484}{0.029}
& \BAMergedCell{8.76}{0.78} & \BAMergedCell{0.455}{0.086} \\

MOMENT (small)
& \BAMergedCell{6.18}{0.13} & \BAMergedCell{0.699}{0.025}
& \BAMergedCell{9.05}{0.28} & \BAMergedCell{0.634}{0.022}
& \BAMergedCell{6.16}{0.13} & \BAMergedCell{0.564}{0.023}
& \BAMergedCell{\underline{8.42}}{0.72} & \BAMergedCell{\underline{0.471}}{0.060} \\

MOMENT (base)
& \BAMergedCell{\underline{6.00}}{0.14} & \BAMergedCell{\underline{0.724}}{0.020}
& \BAMergedCell{\underline{9.05}}{0.13} & \BAMergedCell{\underline{0.637}}{0.022}
& \BAMergedCell{\underline{6.01}}{0.10} & \BAMergedCell{\underline{0.589}}{0.012}
& \BAMergedCell{8.95}{0.86} & \BAMergedCell{0.388}{0.064} \\

MOMENT (large)
& \BAMergedCell{6.11}{0.17} & \BAMergedCell{0.689}{0.054}
& \BAMergedCell{9.51}{0.39} & \BAMergedCell{0.584}{0.034}
& \BAMergedCell{6.26}{0.06} & \BAMergedCell{0.540}{0.026}
& \BAMergedCell{9.14}{0.97} & \BAMergedCell{0.400}{0.065} \\

\midrule
\BAMergedGroup{Supervised-Pre Sleep}
CoreSleep
& \BAMergedCell{6.05}{0.10} & \BAMergedCell{0.732}{0.008}
& \BAMergedCell{8.05}{0.24} & \BAMergedCell{0.720}{0.023}
& \BAMergedCell{6.29}{0.13} & \BAMergedCell{0.551}{0.029}
& \BAMergedCell{\underline{7.85}}{1.17} & \BAMergedCell{\underline{0.584}}{0.153} \\

SleepTransformer
& \BAMergedCell{\underline{6.00}}{0.08} & \BAMergedCell{\underline{0.737}}{0.007}
& \BAMergedCell{\underline{7.83}}{0.24} & \BAMergedCell{\underline{0.745}}{0.026}
& \BAMergedCell{\underline{6.17}}{0.18} & \BAMergedCell{\underline{0.563}}{0.029}
& \BAMergedCell{8.33}{1.25} & \BAMergedCell{0.491}{0.139} \\

SleepyCo
& \BAMergedCell{6.07}{0.09} & \BAMergedCell{0.727}{0.004}
& \BAMergedCell{8.28}{0.53} & \BAMergedCell{0.700}{0.054}
& \BAMergedCell{6.35}{0.21} & \BAMergedCell{0.524}{0.037}
& \BAMergedCell{7.92}{1.17} & \BAMergedCell{0.560}{0.155} \\

\midrule
\BAMergedGroup{Supervised-Pre Age}
Sun et al. (2019)
& \BAMergedCell{\textbf{4.72}}{0.11} & \BAMergedCell{\textbf{0.861}}{0.008}
& \BAMergedCell{\textbf{7.06}}{0.13} & \BAMergedCell{\textbf{0.870}}{0.017}
& \BAMergedCell{\textbf{5.96}}{0.17} & \BAMergedCell{\textbf{0.729}}{0.017}
& \BAMergedCell{\textbf{6.66}}{0.71} & \BAMergedCell{\textbf{0.810}}{0.062} \\

\midrule
\BAMergedGroup{Predict train-mean}
Predict train-mean
& \BAMergedCell{9.31}{0.08} & --
& \BAMergedCell{12.08}{0.31} & --
& \BAMergedCell{7.76}{0.08} & --
& \BAMergedCell{9.77}{1.00} & -- \\

\bottomrule
\end{tabular}
\end{adjustbox}

\caption{%
Brain-age prediction performance on the remaining cohorts:
SHHS, STAGES, MESA, and HPAP.
For each cohort, we report CV-test MAE in years (lower is better)
and Pearson correlation $r$ between predicted and chronological age
(higher is better), as mean $\pm$ standard deviation across five outer folds.
The best non-baseline mean for each cohort and metric is bolded;
the best means within model families containing multiple models
are underlined.
Dashes indicate that within-fold Pearson $r$ is undefined for the
constant predict-train-mean baseline.
}
\label{tab:brain_age_guido_combined}

\label{tab:brain_age_guido}
\label{tab:brain_age_guido_pearson_r}
\end{table*}

\paragraph{Effect of linear-probe regularization.}
The effect of probe regularization was examined on a representative subset of cohorts and models (Table \ref{tab:ridge_ols_sensitivity}). We compared the  Ridge regression probe with an unregularized ordinary least squares (OLS) alternative. Regularization consistently improved stability, particularly in smaller cohorts. This analysis also indicates that the reported linear-probing results are partly shaped by the choice of regularization

\begin{table*}[t]
\centering
\caption{
\textbf{Ridge regularization stabilizes linear probing, especially in smaller cohorts.}
Baseline MAE denotes the dataset-specific dummy error obtained by predicting the training-set mean age. $\Delta$ denotes OLS MAE minus Ridge MAE; positive values indicate that Ridge performs better. Dataset-level values are aggregated across evaluated checkpoints within each cohort, whereas model family-level values are aggregated across all model--dataset evaluations within each model family.
}
\label{tab:ridge_ols_sensitivity}
\resizebox{\textwidth}{!}{
\begin{tabular}{lrrrrr}
\toprule
\multicolumn{6}{l}{\textbf{A. Dataset-level summary}} \\
\midrule
Dataset 
& $n_{\mathrm{train}}$ 
& Baseline MAE 
& Ridge MAE 
& OLS MAE 
& $\Delta$ \\
\midrule
WSC 
& 2055 
& 6.44 
& $5.71 \pm 0.18$ 
& $5.89 \pm 0.36$ 
& $+0.17$ \\
CFS 
& 584 
& 16.37 
& $8.55 \pm 0.62$ 
& $12.04 \pm 3.51$ 
& $+3.49$ \\
PhysioNet 2026 
& 169 
& 7.66 
& $7.23 \pm 0.27$ 
& $18.79 \pm 13.75$ 
& $+11.56$ \\
ISRUC 
& 99 
& 13.78 
& $10.51 \pm 0.86$ 
& $19.69 \pm 5.13$ 
& $+9.18$ \\
SleepEDF-SC 
& 62 
& 18.33 
& $13.19 \pm 1.54$ 
& $18.10 \pm 4.31$ 
& $+4.92$ \\
\midrule
\multicolumn{6}{l}{\textbf{B. Model-family summary}} \\
\midrule
Family 
& \multicolumn{2}{c}{}
& Ridge MAE 
& OLS MAE 
& $\Delta$ \\
\midrule
EEG-FM 
& \multicolumn{2}{c}{}
& $9.32 \pm 2.97$ 
& $15.81 \pm 10.31$ 
& $+6.49$ \\
Supervised-Pre Sleep 
& \multicolumn{2}{c}{}
& $8.24 \pm 2.11$ 
& $13.27 \pm 7.50$ 
& $+5.02$ \\
TS-FMs 
& \multicolumn{2}{c}{}
& $9.00 \pm 2.74$ 
& $14.17 \pm 4.86$ 
& $+5.17$ \\
\bottomrule
\end{tabular}
}
\end{table*}

\begin{figure}[ht]
    \centering
    \includegraphics[width=0.99\textwidth]{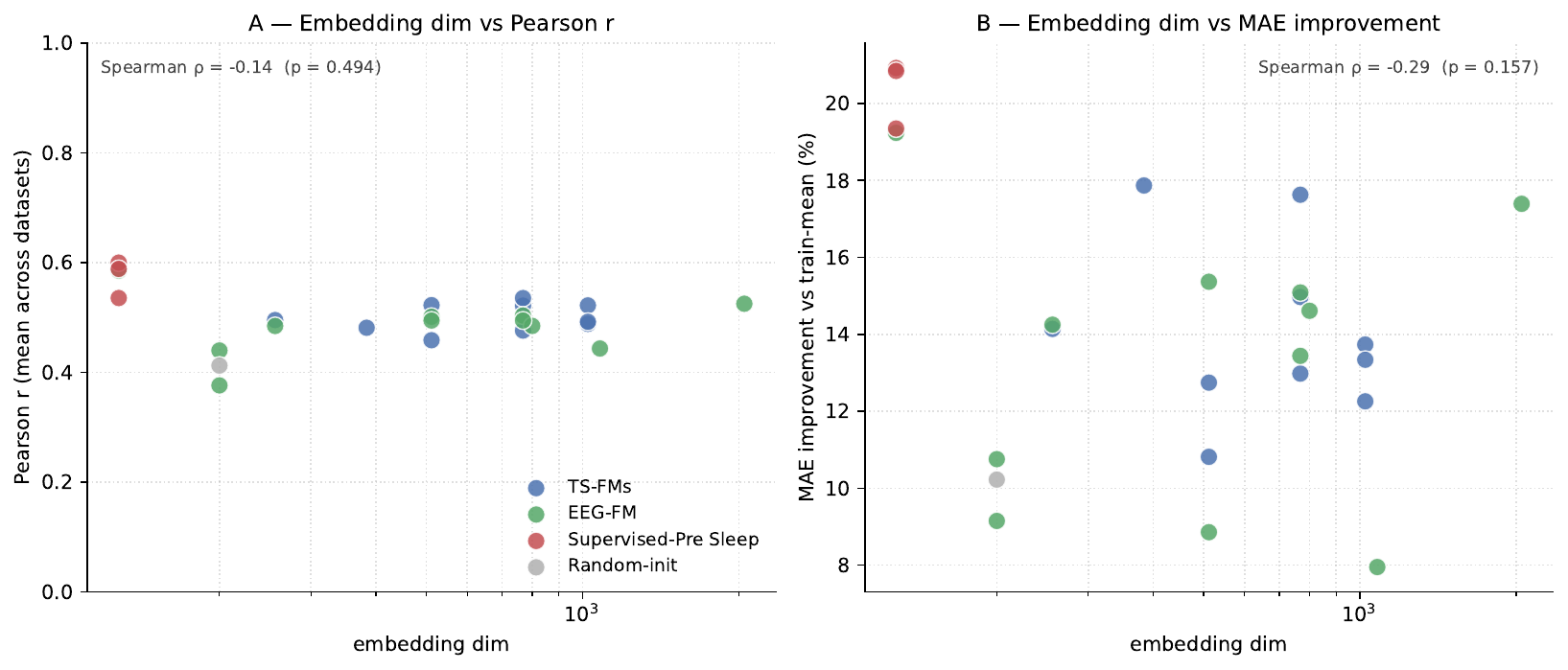}
 \caption{                                                   
  \textbf{Embedding dimensionality does not correlate with brain-age performance.}                           
  Each marker is one frozen-encoder checkpoint,   
  with metrics fold-averaged across the seven main-pool      
  brain-age cohorts (PN2026, WSC, ISRUC, SleepEDF SC, 
  CFS, MrOS). Marker color denotes encoder family.
  \textbf{(a)} Embedding dim versus mean Pearson $r$ between  
  predicted and chronological age. \textbf{(b)} Embedding dim
  versus mean MAE improvement over a predict-cohort-mean      
  baseline (\%). The relationship is essentially flat in (a)
  (Spearman $\rho = -0.14$, $p = 0.49$) and weakly negative in
   (b) ($\rho = -0.29$, $p = 0.16$).
  }
  \label{fig:emb_dim_result}
\end{figure}

\clearpage
\paragraph{Relationship between embedding dimensionality and brain-age performance.}
We next examined whether brain-age performance scales with embedding dimensionality across the evaluated frozen-encoder checkpoints. Across 25 models, embedding dimension showed no significant association with the Pearson correlation metric and only a weak, non-significant negative association with MAE improvement over the train-mean baseline ($\rho=-0.29$, $p=0.16$; Figure \ref{fig:emb_dim_result}). Thus, larger embedding spaces did not translate into systematically better age prediction.

This lack of dimensional scaling is most apparent for supervised sleep-stage models, which use comparatively compact 128-dimensional representations yet achieve some of the strongest average MAE improvements. In contrast, higher-dimensional EEG and time-series foundation models are competitive but not consistently superior. These results suggest that the information encoded by the representation, rather than the dimensionality of the embedding space, is the primary determinant of brain-age regression performance.

\begin{figure}[ht]
    \centering
\includegraphics[width=0.99\textwidth]{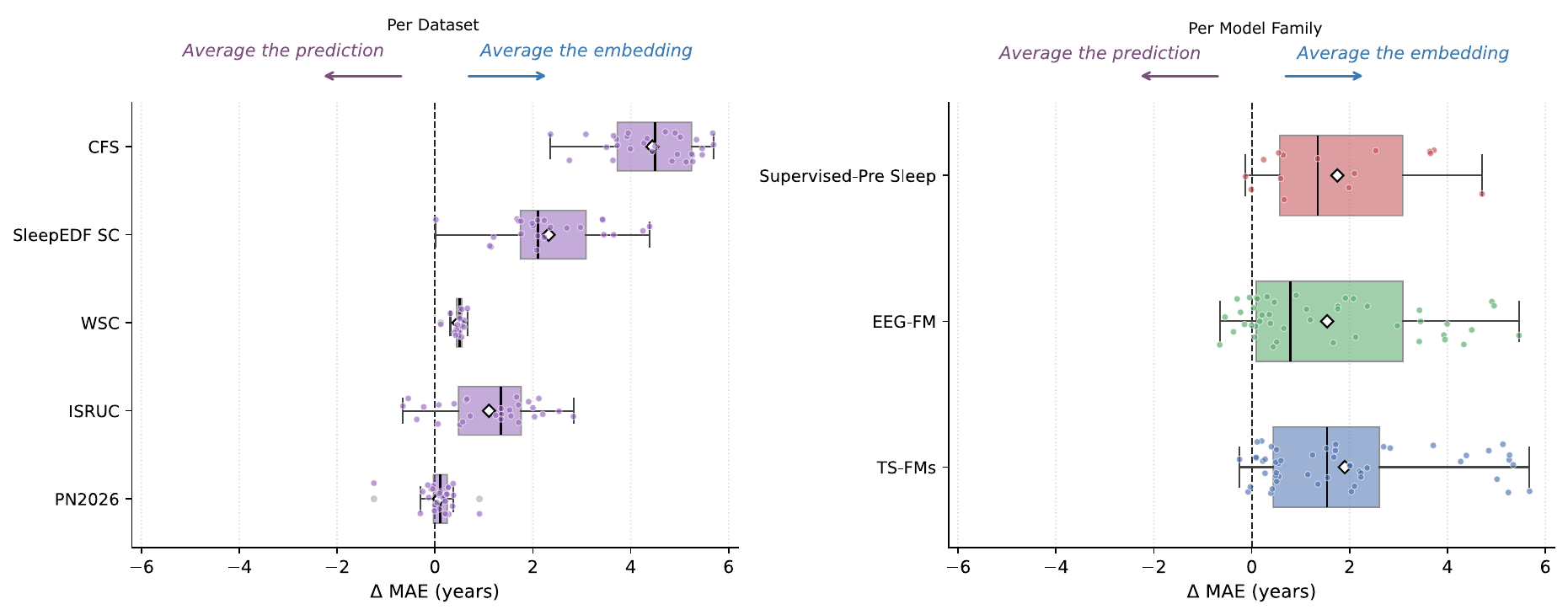}
 \caption{                                                   
  \textbf{Aggregating per subject tends to outperform aggregating  
  per epoch.}                                               
  Each marker is one (model, cohort) pair, fold-averaged      
  across the 5  folds;  The horizontal axis shows
  $\Delta\,\text{MAE} = \text{MAE}_{\text{epoch}} -           
  \text{MAE}_{\text{subject}}$; positive values indicate that
  subject-level Ridge (mean-pool a recording's epoch          
  embeddings into one subject vector before fitting
  to age beats epoch-level Ridge fit per 30s
  epoch, then average predictions per subject).
  \textbf{(left)} Per-cohort distributions, ordered by
  $n_{\text{outer\_train}}$ ascending;
  }
  \label{fig:subject_vs_epoch}
\end{figure}

\paragraph{Comparison of subject-level and epoch-level aggregation.}
Finally, we compared two aggregation strategies for using epoch-level embeddings. In the subject-level pipeline, epoch embeddings were averaged within each recording before fitting Ridge regression from the resulting subject-level vector to age. In the epoch-level pipeline, Ridge regression was fitted to individual 30-second epoch embeddings, and predicted ages were averaged within each recording. This comparison was performed on a subset of models and datasets. The subject-level pipeline generally outperformed the epoch-level predict-then-average strategy, supporting mean-pooled subject embeddings as the main approach for the reported brain-age analyses (Figure \ref{fig:subject_vs_epoch}).

\paragraph{Cross-Dataset Generalization}

\begin{figure}[ht]
    \centering
    \includegraphics[width=0.99\textwidth]{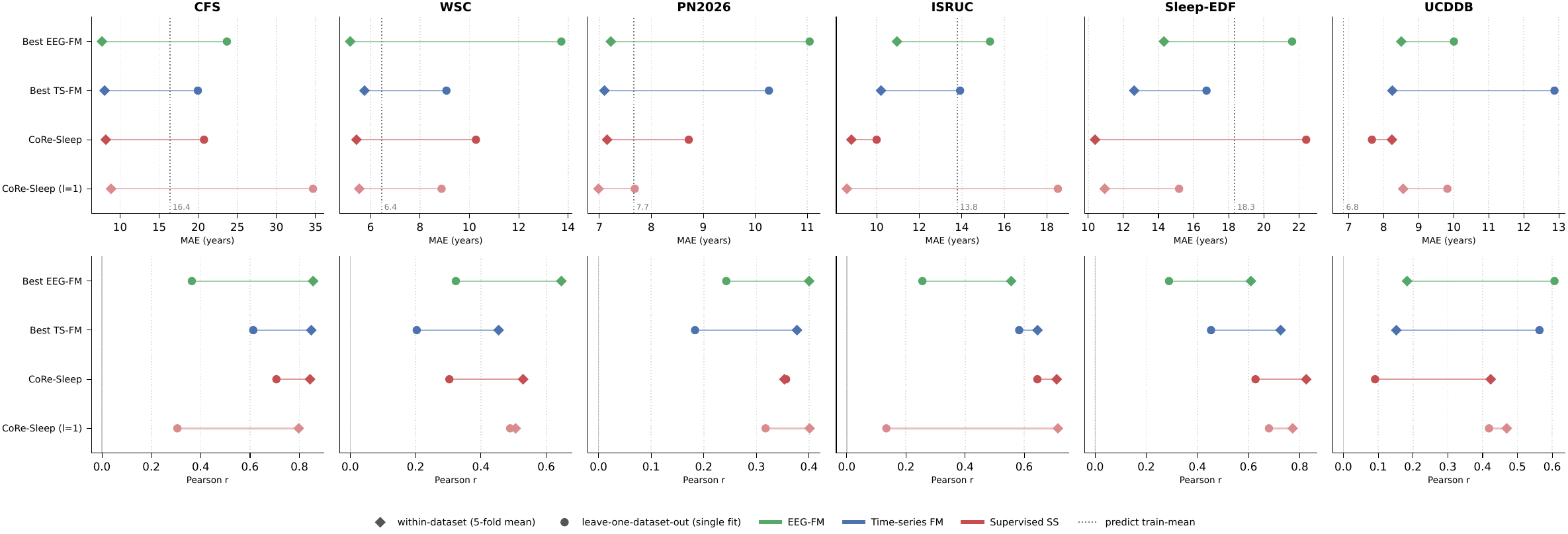}
    \caption{
    \textbf{Leave-one-dataset-out (LODO) brain-age prediction.}
    Each panel corresponds to one held-out cohort, with MAE shown in the top row and Pearson $r$ in the bottom row. Within-dataset performance is compared with LODO performance, where the probe is trained on the pooled remaining five cohorts and evaluated on the held-out cohort without labeled target data. Marker color denotes model family: EEG-FM (green), time-series FM (blue), and SHHS-supervised models (red; lighter red for the $l=1$ variant). For EEG-FMs and TS-FMs, the best-performing model for each cohort is shown based on LODO MAE. The dotted vertical line marks the predict-train-mean baseline. UCDDB is the only cohort in which LODO transfer improves correlation, consistent with its small sample size and restricted age range relative to the pooled training data.
    }
    \label{fig:lodo_brain_age}
\end{figure}

Finally, we assess whether age can be predicted in a fully dataset-general setting, without using labeled data from the target cohort. We therefore perform leave-one-dataset-out (LODO) ridge probing, training on the pooled remaining cohorts and evaluating on the held-out dataset. This setting is particularly challenging because cohorts differ not only in recording domain but also in their age distributions, with standard deviations ranging from 8.5 years in WSC to 22.0 years in Sleep-EDF. Pooling thus changes both the input distribution and the regression target. Performance declines substantially under LODO, with median MAE increasing by 3.2–13.9 years and Pearson (r) decreasing by 0.11–0.34. Although the supervised models show the smallest MAE increases, there is otherwise little evidence for a consistent family-level ordering. (Figure \ref{fig:lodo_brain_age}).

Overall, the loss in performance is large enough to limit practical utility: only a few runs outperform the trivial baseline of predicting the held-out cohort's mean age. The extent of degradation appears to depend more strongly on the age distribution of the held-out cohort than on model family. Cohorts with broader age ranges show the largest reductions in correlation, whereas UCDDB, which is both small and age-restricted (SD 9.4; 25 subjects), is the only cohort to improve under LODO (\(\Delta r = +0.10\)). overall, only 3 of 42 LODO fits outperform the trivial baseline of predicting the held-out cohort's mean age, suggesting that cross-dataset age prediction is limited less by a particular model family than by mismatch in the target age distribution and the amount of target-specific data available.

\begin{figure}[h]
    \centering
\includegraphics[width=0.99\textwidth]{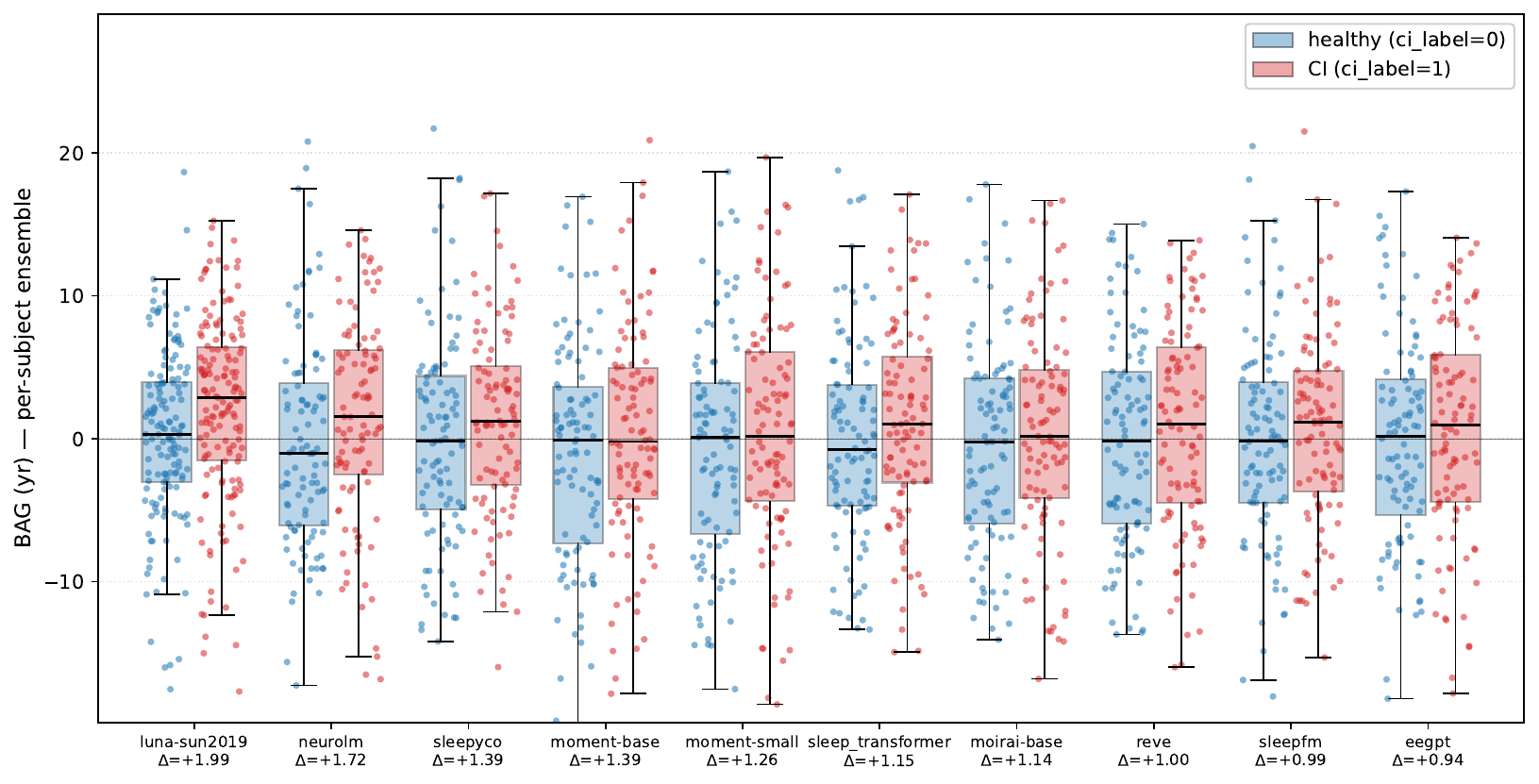}
 \caption{
\textbf{Brain-age-gap (BAG) distributions on the PN2026 Healthy/Cognitive Impairment holdout.}
BAG was computed per subject as predicted age minus chronological age using the 5-fold ensemble prediction. We show the 10 models with the largest positive mean BAG difference, ordered by
$\Delta\mathrm{BAG} =
\overline{\mathrm{BAG}}_{\mathrm{CI}} -
\overline{\mathrm{BAG}}_{\mathrm{H}}$.
For each model, box plots show the BAG distributions of 100 age-matched healthy subjects and 100 cognitively impaired (CI) subjects; dots indicate individual subjects. The horizontal line marks BAG $=0$. Values below model names report $\Delta\mathrm{BAG}$ in years, where positive values indicate older-appearing EEG in the CI group.
}
\label{fig:dbag_per_subject_top10}
\end{figure}

\begin{table*}[t]
\centering
\tiny
\setlength{\tabcolsep}{3.5pt}
\renewcommand{\arraystretch}{1.10}

\begin{adjustbox}{max width=\textwidth}
\begin{tabular}{
    @{}ll
    cc@{\hspace{9pt}}
    cc@{\hspace{15pt}}
    cc@{\hspace{9pt}}
    cc@{}
}
\toprule
& & \multicolumn{4}{c}{All models ($k=26$)}
  & \multicolumn{4}{c}{EEG-FM only ($k=11$)} \\
\cmidrule(lr){3-6}
\cmidrule(lr){7-10}
& & \multicolumn{2}{c}{MAE $\downarrow$}
  & \multicolumn{2}{c}{Pearson $r$ $\uparrow$}
  & \multicolumn{2}{c}{MAE $\downarrow$}
  & \multicolumn{2}{c}{Pearson $r$ $\uparrow$} \\
\cmidrule(lr){3-4}
\cmidrule(lr){5-6}
\cmidrule(lr){7-8}
\cmidrule(lr){9-10}
Model & Family
& Wins & $p$ & Wins & $p$
& Wins & $p$ & Wins & $p$ \\
\midrule

Sun et al. (2019) & Supervised-Pre Age
& \textbf{5/7} & $\boldsymbol{<\!10^{-3}}$
& \textbf{7/7} & $\boldsymbol{<\!10^{-3}}$
& -- & --
& -- & -- \\

EEGPT & EEG-FM
& \textbf{3/7} & \textbf{0.038}
& 2/7 & 0.189
& 3/7 & 0.292
& 4/7 & 0.093 \\

NeuroRVQ-EEG & EEG-FM
& 2/7 & 0.189
& 1/7 & 0.576
& \textbf{5/7} & \textbf{0.019}
& 4/7 & 0.093 \\

REVE & EEG-FM
& 2/7 & 0.189
& 2/7 & 0.189
& 3/7 & 0.292
& 2/7 & 0.610 \\

CoreSleep & Supervised-Pre Sleep
& 2/7 & 0.189
& 2/7 & 0.189
& -- & --
& -- & -- \\

SleepTransformer & Supervised-Pre Sleep
& 2/7 & 0.189
& 2/7 & 0.189
& -- & --
& -- & -- \\

SleepFM & EEG-FM
& 1/7 & 0.576
& 1/7 & 0.576
& 3/7 & 0.292
& 3/7 & 0.292 \\

NeuroLM-VQ & EEG-FM
& 1/7 & 0.576
& 1/7 & 0.576
& 2/7 & 0.610
& 3/7 & 0.292 \\

Chronos-T5 (tiny) & TS-FMs
& 1/7 & 0.576
& 0/7 & 1.000
& -- & --
& -- & -- \\

MOMENT (base) & TS-FMs
& 1/7 & 0.576
& 1/7 & 0.576
& -- & --
& -- & -- \\

SleepyCo & Supervised-Pre Sleep
& 1/7 & 0.576
& 2/7 & 0.189
& -- & --
& -- & -- \\

BIOT & EEG-FM
& 0/7 & 1.000
& 0/7 & 1.000
& 2/7 & 0.610
& 2/7 & 0.610 \\

STEEGFormer-B & EEG-FM
& 0/7 & 1.000
& 0/7 & 1.000
& 2/7 & 0.610
& 1/7 & 0.892 \\

LaBraM & EEG-FM
& 0/7 & 1.000
& 0/7 & 1.000
& 1/7 & 0.892
& 0/7 & 1.000 \\

CBraMod & EEG-FM
& 0/7 & 1.000
& 0/7 & 1.000
& 0/7 & 1.000
& 0/7 & 1.000 \\

Neuro-GPT & EEG-FM
& 0/7 & 1.000
& 0/7 & 1.000
& 0/7 & 1.000
& 0/7 & 1.000 \\

STEEGFormer-S & EEG-FM
& 0/7 & 1.000
& 0/7 & 1.000
& 0/7 & 1.000
& 2/7 & 0.610 \\

CBraMod (random init) & Random-init
& 0/7 & 1.000
& 0/7 & 1.000
& -- & --
& -- & -- \\

Chronos-T5 (base) & TS-FMs
& 0/7 & 1.000
& 0/7 & 1.000
& -- & --
& -- & -- \\

Chronos-T5 (large) & TS-FMs
& 0/7 & 1.000
& 0/7 & 1.000
& -- & --
& -- & -- \\

Chronos-T5 (small) & TS-FMs
& 0/7 & 1.000
& 0/7 & 1.000
& -- & --
& -- & -- \\

MOMENT (large) & TS-FMs
& 0/7 & 1.000
& 0/7 & 1.000
& -- & --
& -- & -- \\

MOMENT (small) & TS-FMs
& 0/7 & 1.000
& 0/7 & 1.000
& -- & --
& -- & -- \\

Moirai (base) & TS-FMs
& 0/7 & 1.000
& 0/7 & 1.000
& -- & --
& -- & -- \\

Moirai (large) & TS-FMs
& 0/7 & 1.000
& 0/7 & 1.000
& -- & --
& -- & -- \\

Moirai (small) & TS-FMs
& 0/7 & 1.000
& 0/7 & 1.000
& -- & --
& -- & -- \\

\bottomrule
\end{tabular}
\end{adjustbox}

\caption{%
\textbf{Binomial top-3 win tests across the seven main-pool
brain-age cohorts.}
Wins count the cohorts on which a model ranks among the top three
by CV-test MAE (lower is better) or Pearson $r$ between predicted
and chronological age (higher is better).
Tests are conducted separately for each metric and model pool.
The exact one-sided binomial test uses $N=7$ and $p_0=3/k$:
$3/26\approx0.115$ for all models and $3/11\approx0.273$
for EEG foundation models only.
Bold wins and $p$-values identify individual tests with
unadjusted $p<0.05$.
Dashes indicate that a model is not in the EEG-FM-only pool.
}
\label{tab:top3_win_test_combined}

\label{tab:top3_win_test_mae}
\label{stat_test_MAE}
\label{tab:top3_win_test_pearson_r}
\label{stat_test_r}
\end{table*}
\clearpage
\newpage
\subsection{BCI}\label{Appendix:BCI_Res}

\begin{figure*}[t]
    \centering
    \includegraphics[width=\textwidth]{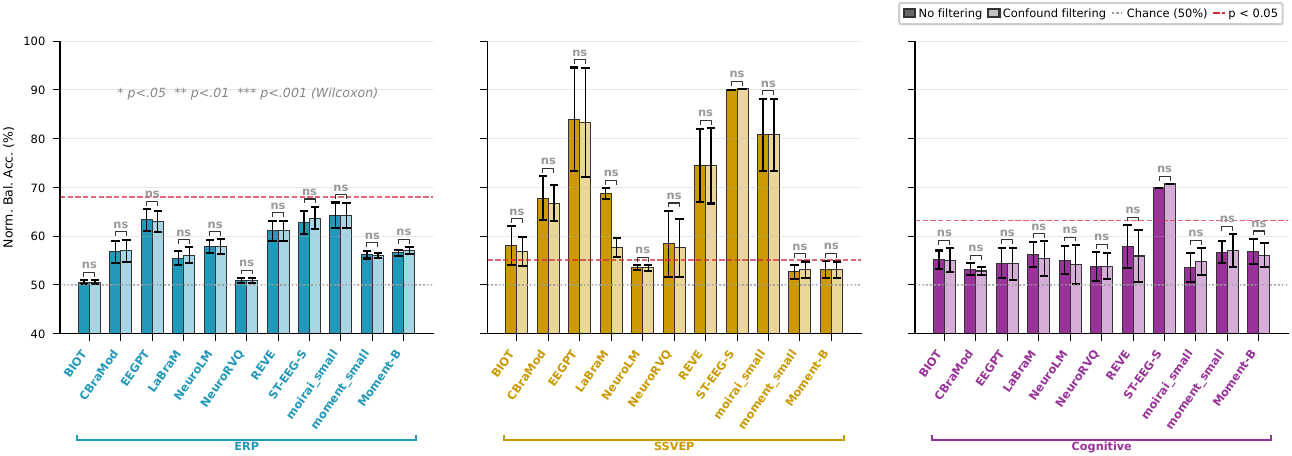}
    \caption{%
        \textbf{Per-paradigm comparison of frozen EEG foundation models under
        No filtering vs.\ Confound filtering for ERP,
        SSVEP, and Cognitive BCI datasets.}
        Bars show the mean normalized balanced accuracy (\%) across datasets
        within each paradigm; error bars denote the standard error of the mean.
        Asterisks indicate significant paired differences between tracks
        (Wilcoxon signed-rank test; $*$\,$p<.05$, $**$\,$p<.01$,
        $***$\,$p<.001$; ``ns'' = not significant).
        The dashed red line marks the aggregated significance threshold
        ($p<0.05$, binomial test); the dotted gray line marks chance level
        (50\%).
        All models use per-patch embeddings except NeuroLM (mean-pool) and
        REVE (mean-pool fallback where per-patch is unavailable).
        Overall, adopting a more task-specific confound filtering strategy
        does not lead to substantial performance changes compared to using
        the filtering configurations originally suggested by the models.
        Across paradigms, No filter and Confound filter results remain largely
        comparable, with most pairwise differences failing to reach
        statistical significance.
        \textbf{ERP:} Most models achieve above-chance performance, with
        EEGPT and ST-EEG-S reaching the highest accuracy ($\sim$65\%).
        Confound filtering produces only marginal variations in performance,
        and none of the observed differences are statistically significant,
        suggesting that ERP decoding primarily relies on robust neural
        representations rather than recording-related confounds.
        \textbf{SSVEP:} EEGPT achieves the strongest performance
        ($\sim$84\% in No filtering), followed by ST-EEG-S ($\sim$78\%), whereas
        NeuroLM and REVE remain near chance level. Although filtering
        effects remain limited overall, SSVEP is the only paradigm showing
        performance consistently above the aggregated significance threshold,
        albeit with substantial inter-dataset variability reflected by the
        large error bars.
        \textbf{Cognitive:} All models remain near or below the aggregated
        significance threshold, with REVE reaching the highest accuracy
        ($\sim$59\%). Differences between filtering strategies are again not
        statistically significant, indicating that current frozen EEG
        foundation models struggle to capture the subtler cognitive-state
        distinctions represented in these datasets.
}
    \label{fig:bci_figure2}
\end{figure*}

\paragraph{Confound filtering on other BCI paragigms does not lead to significant changes in performance.}
The comparison between no filtering and confound filtering shows that adopting a more task-specific filtering strategy does not substantially alter the performance of frozen EEG foundation models relative to the filtering configurations originally suggested by each model. Here, confound filtering refers to paradigm-specific preprocessing designed to retain task-relevant neural activity while reducing potential recording-related artifacts: ERP datasets were filtered using a 0.5-40,Hz band-pass filter, SSVEP datasets were only high-pass filtered without low-pass filtering in order to preserve stimulus harmonics, and Cognitive datasets were filtered within the 4-40,Hz range. As seen in Figure~\ref{fig:bci_figure2}, across ERP, SSVEP, and Cognitive paradigms, the observed differences between filtering strategies are generally small and fail to reach statistical significance. Moreover, with the exception of the SSVEP paradigm, most results remain below the aggregated statistical significance threshold, indicating that current frozen EEG foundation models still struggle to consistently extract robust task-relevant representations across paradigms. These findings suggest that model architecture and pretraining strategy currently have a stronger impact on downstream decoding performance than the specific filtering configuration applied during evaluation.

\begin{figure}[t]
    \centering
    \includegraphics[width=0.98\linewidth]{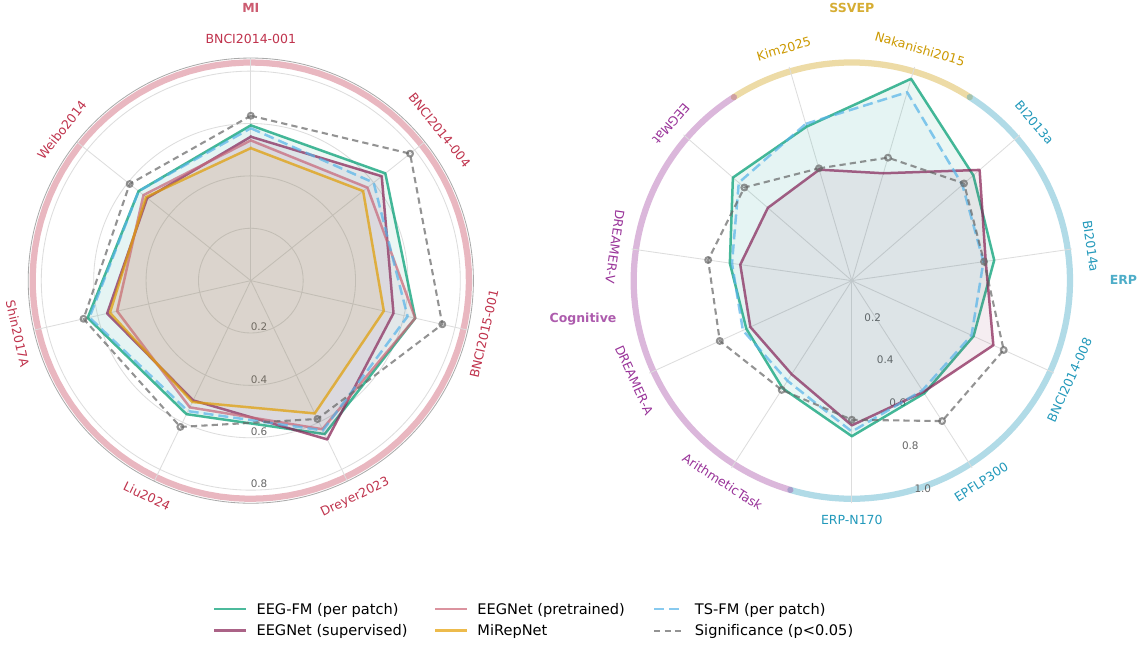}
    \caption{
    Normalized balanced accuracy of foundation model families under task-specific confound filtering across four BCI paradigms.
    Each spoke represents a dataset, and the radial axis maps chance-level performance to 0.5 and perfect classification to 1.0,
    enabling comparisons across paradigms with different class counts and difficulty levels.
    \textbf{Left:} Motor Imagery (MI) datasets comparing EEG foundation models (EEG-FM, per-patch pooling),
    time-series foundation models (TS-FM, mean pooling), pretrained supervised EEGNet models, and MIRepNet against the
    aggregated binomial significance threshold ($p<0.05$).
    \textbf{Right:} SSVEP, ERP, and Cognitive datasets comparing EEG-FM and TS-FM families.
    Colored arcs indicate paradigm groupings.
    For each family, the best-performing model across all members is shown per dataset.
    }
    \label{fig:bci_radar_trackD}
\end{figure}

\paragraph{Best performing model families per paradigm}
Figure~\ref{fig:bci_radar_trackD} provides a per-dataset comparison of the best-performing model families under task-specific confound filtering using radar plots of normalized balanced accuracy. Each spoke corresponds to an individual dataset, while the radial axis maps chance-level performance to 0.5 and perfect classification to 1.0, allowing direct comparison across paradigms with different numbers of classes and intrinsic task difficulty. For each family, we report the best-performing model across all members on each dataset, highlighting the strongest achievable performance of EEG foundation models (EEG-FM), time-series foundation models (TS-FM), and supervised baselines under a unified frozen-evaluation framework.

For the MI paradigm, we additionally compare against pretrained supervised models obtained by ~\citep{GuePapTan23a}, which is why EEGNet (pretrained)~\citep{lawhern2018eegnet} is only included in this section. To ensure fair comparison with EEG-FMs and TS-FMs, embeddings were extracted from the pretrained supervised models using the same frozen-encoder protocol employed for the foundation models, followed by linear probing using a leave-one-subject-out (LOSO) evaluation strategy. The same evaluation pipeline was applied under task-specific confound filtering, where the linear classification layer was adapted to account for the reduced temporal dimensionality introduced by the three-second filtered segments. For MIRepNet~\citep{liu2025mirepnet}, we similarly froze the pretrained encoder and extracted embeddings before linear probing. MIRepNet was originally trained as a paradigm-specific motor imagery foundation model using a hybrid pretraining strategy combining supervised MI classification with self-supervised masked token reconstruction, together with MI-specific preprocessing including 8--30\,Hz filtering, temporal alignment, and channel-template normalization designed to capture sensorimotor rhythm dynamics associated with ERD/ERS phenomena in motor imagery tasks. Although MIRepNet follows a supervised pretraining philosophy conceptually related to EEGNet-style approaches, it additionally incorporates self-supervised representation learning and large-scale MI-specific pretraining across multiple datasets. Nevertheless, despite these specialized training strategies, neither EEGNet nor MIRepNet outperformed the best EEG foundation models in the frozen linear-probing setting, and their normalized performance generally remained close to chance level, suggesting that supervised specialization alone does not necessarily yield more transferable EEG representations under strict cross-subject frozen evaluation protocols.

\begin{figure*}[t]
\centering

\begin{subfigure}[t]{0.48\textwidth}
    \centering
    \includegraphics[width=\linewidth]{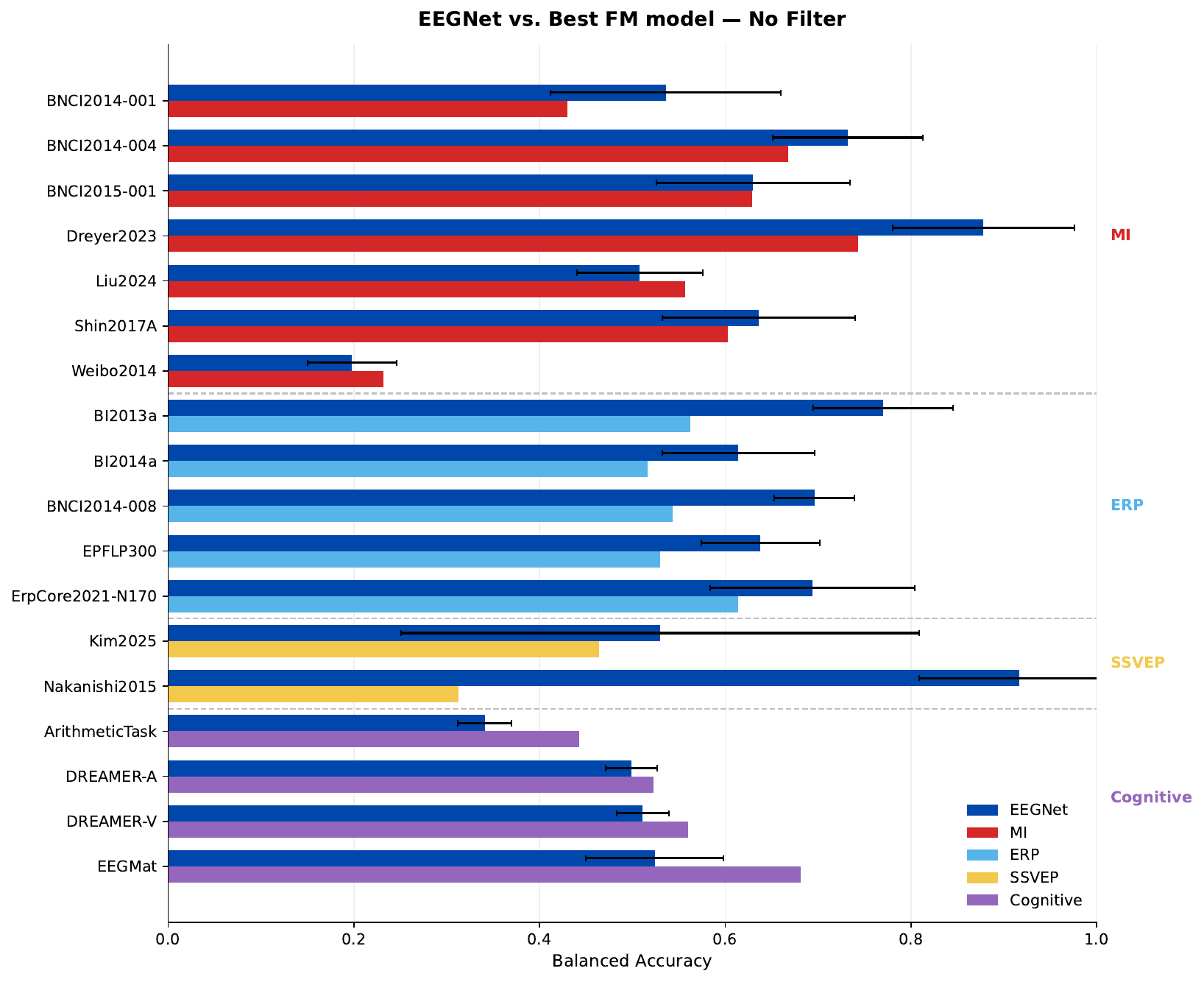}
    \caption{No Filter}
    \label{fig:eegnet-fm-no-filter}
\end{subfigure}
\hfill
\begin{subfigure}[t]{0.48\textwidth}
    \centering
    \includegraphics[width=\linewidth]{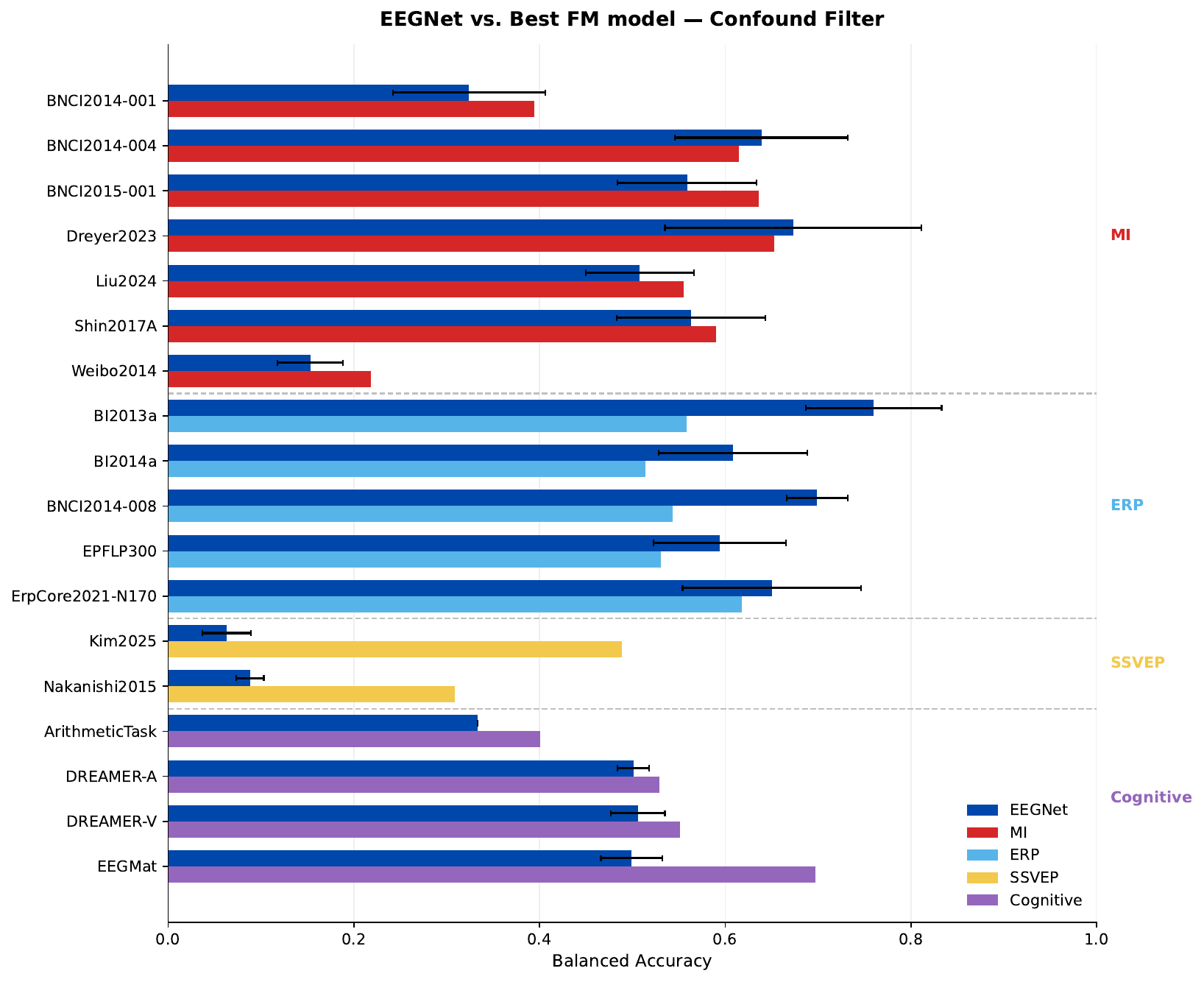}
    \caption{Confound Filter}
    \label{fig:eegnet-fm-confound-filter}
\end{subfigure}
\caption{\textbf{Comparison of foundation models with end-to-end supervised EEGNet for BCI decoding.}
Balanced accuracy of the best-performing FM for each dataset is compared with EEGNet trained end-to-end using leave-one-subject-out (LOSO) evaluation under \textbf{(a)} No Filter and \textbf{(b)} Confound Filter conditions. EEGNet is shown in dark blue, while the best FM is colored according to the BCI paradigm: motor imagery (MI) in red, event-related potentials (ERP) in light blue, steady-state visually evoked potentials (SSVEP) in yellow, and cognitive tasks in purple. Error bars indicate the standard deviation of EEGNet performance across held-out subjects. Datasets are grouped by paradigm.}
\label{fig:eegnet-vs-best-fm}
\end{figure*}

\subsection{Comparison with End-to-End Supervised EEGNet}
\label{app:bci-eegnet-comparison}

To further contextualize the performance of foundation models for BCI decoding, we additionally compare against EEGNet, a widely used supervised architecture for EEG classification. For each dataset, EEGNet was trained end-to-end using the LOSO evaluation strategy. The evaluation was performed independently under both the No Filter and Confound Filter settings, providing a dataset-specific fully supervised reference against which to compare the representations learned by the pretrained FMs. Figure~\ref{fig:eegnet-vs-best-fm} reports the resulting balanced accuracy together with that of the best-performing FM on each dataset.

Under the No Filter setting, EEGNet outperforms the best-performing FM on 13 of the 18 datasets. The largest differences are observed for Nakanishi2015, where EEGNet reaches $0.917$ compared with $0.313$ for EEGPT, BI2013a ($0.770$ vs.\ $0.562$ for STEEGFormer-Large), and Dreyer2023 ($0.878$ vs.\ $0.743$ for EEGPT). Nevertheless, the best FMs outperform the end-to-end supervised baseline on several datasets, with STEEGFormer-Large improving over EEGNet on EEGMat ($0.681$ vs.\ $0.524$), BIOT on ArithmeticTask ($0.443$ vs.\ $0.341$), REVE on DREAMER-V ($0.560$ vs.\ $0.511$), and EEGPT on Liu2024 ($0.557$ vs.\ $0.508$). Across the remaining datasets, the best-performing FMs are EEGPT for BI2014a, ErpCore2021-N170, Kim2025, Nakanishi2015, and Shin2017A; NeuroLM for BNCI2014-001, BNCI2014-008, and EPFLP300; NeuroGPT for BNCI2014-004 and BNCI2015-001; STEEGFormer-Base for DREAMER-A; and STEEGFormer-Large for Weibo2014.

After applying the Confound Filter, the comparison becomes substantially more balanced, with the best-performing FM outperforming EEGNet on 9 of the 18 datasets. Particularly large advantages are observed for Kim2025, where BIOT reaches $0.489$ compared with $0.063$ for EEGNet; Nakanishi2015, where EEGPT reaches $0.309$ compared with $0.088$; and EEGMat, where STEEGFormer-Large reaches $0.697$ compared with $0.499$. EEGNet nevertheless retains a clear advantage on BI2013a ($0.760$ vs.\ $0.558$ for STEEGFormer-Large), together with smaller advantages across several other datasets. Under this setting, the best-performing FM is STEEGFormer-Large for ArithmeticTask, BI2013a, EEGMat, Shin2017A, and Weibo2014; EEGPT for BI2014a, BNCI2014-001, BNCI2014-004, Dreyer2023, ErpCore2021-N170, and Nakanishi2015; NeuroLM for BNCI2014-008 and EPFLP300; STEEGFormer-Base for BNCI2015-001; LaBraM for DREAMER-A; REVE for DREAMER-V; BIOT for Kim2025; and Chronos-T5-Large for Liu2024.

Overall, the LOSO comparison shows that end-to-end dataset-specific supervised training remains a strong reference for BCI decoding. However, the advantage of EEGNet is markedly reduced under confound-controlled evaluation, where the best-performing FM matches or exceeds the supervised baseline on half of the evaluated datasets. Importantly, no single FM consistently dominates across datasets or BCI paradigms, with the best-performing model instead varying substantially across tasks. These results complement the primary FM evaluation by showing that pretrained representations can remain competitive with a conventional supervised architecture trained end-to-end specifically for each target dataset.

  \begin{figure*}[ht]
    \centering                                                              
    \includegraphics[width=\textwidth]{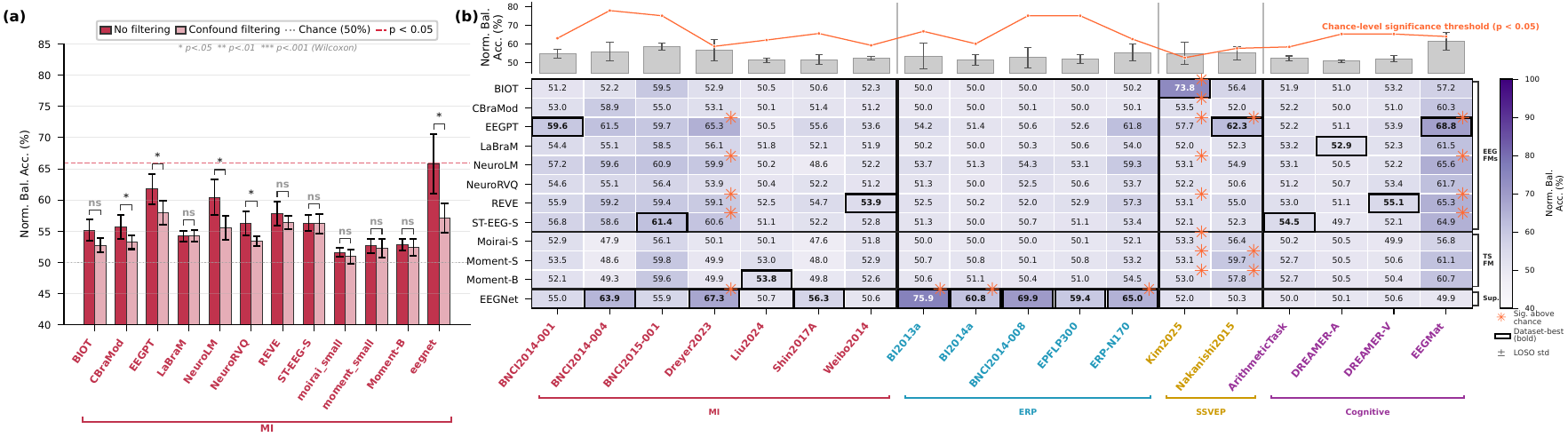}
    \caption{\textbf{BCI: MI performance is strongly driven by confounds} Linear probing performance of frozen EEG and time-series
        foundation models across 17 BCI datasets spanning four paradigms (MI,
        ERP, SSVEP, Cognitive) for mean pool. Metrics are
        reported as normalized balanced accuracy (\%), where 50\% equals chance
        regardless of the number of classes. The per-dataset significance
        threshold ($p<0.05$, binomial test \cite{BetterThanChance}) is shown as a dashed red line or
        orange markers.
        \textbf{(a)}~Normalized balanced accuracy on Motor Imagery datasets
        comparing embeddings extracted without confound filtering and with confound filtering 
        bars for 8 EEG-FMs, 3 TS-FMs and EEGNet supervised model. Error bars denote the standard
        error of the mean across MI datasets (Wilcoxon signed-rank test; $*$\,$p<.05$,
        $**$\,$p<.01$, $***$\,$p<.001$). Confound filtering significantly affect performance, suggesting that the evaluated models
        rely primarily on artifacts.
        \textbf{(b)}~Per-dataset heatmap for 8 EEG-FMs and 3 TS-FM evaluated on confound filtered data. Cell color encodes normalized balanced accuracy
        (40-100\%; darker~$=$~higher); each cell reports the score (bold for dataset-best) with LOSO standard deviation ($\pm$). Orange stars
        (\textcolor{orange}{$\ast$}) mark results significantly above chance. Note that due to the varying number of classes and the low sample counts, the significance level is much higher than chance level and varies across datasets.}    
  \label{fig:BCI_mean}                                     
  \end{figure*}  
 
\subsection{Token aggregation strategy}
We further investigate how to aggregate the token-level representations produced by the foundation models for downstream BCI decoding. Specifically, we compare \emph{mean pooling}, where the representations of all output embeddings are averaged into a single feature vector (Fig.~\ref{fig:BCI_mean}), against \emph{token flattening}, where the individual token representations are concatenated and jointly provided to the downstream classifier (Fig.~\ref{fig:BCI}). Overall, token flattening consistently provides stronger downstream performance than mean pooling, with balanced accuracy increasing on average from $45.9\%$ to $53.0\%$, corresponding to an absolute improvement of $7.0$ percentage points.

The benefit of preserving individual token representations varies substantially across BCI paradigms. The largest improvement is observed for SSVEP, where token flattening increases balanced accuracy by $22.5$ percentage points on average. Considerable improvements are also observed for MI and ERP, with average gains of $6.4$ and $6.3$ percentage points, respectively, whereas the improvement for Cognitive tasks is considerably smaller at $0.9$ percentage points. One possible explanation for these differences is the temporal and spatial localization of the task-relevant information. For paradigms characterized by shorter trial durations and finer temporal structure, the discriminative information may be concentrated within relatively narrow time intervals and specific brain regions. Averaging representations across tokens may therefore dilute these localized patterns, whereas retaining individual token representations gives the downstream classifier greater flexibility to identify and exploit information associated with specific time-related and spatial components. This explains the larger benefit of token flattening observed for SSVEP, MI, and ERP than for cognitive tasks.

The improvement is observed under both evaluation settings, although it is larger without confound filtering, where token flattening increases balanced accuracy by $8.5$ percentage points on average, compared with $5.6$ percentage points under the Confound Filter. Overall, these results suggest that averaging token-level representations can discard information that remains useful for downstream BCI decoding. In contrast, preserving the individual token representations through flattening allows the downstream classifier to exploit information distributed across tokens, resulting in a substantially stronger readout strategy for the evaluated FMs.

\begin{table}[t]
\centering
\small
\caption{\textbf{Binomial top-3 win test across 17 BCI datasets (DREAMER counted twice for two distinct tasks) for per patch token aggregation.} For each model, the exact one-sided binomial test asks whether the number of datasets on which the model ranks in the top~3 (by normalized balanced accuracy) exceeds the chance rate $\bar{p}_0 = 3/k$. The pool contains $k=12$ models (8 EEG-FMs + 3 TS-FMs + EEGNet) evaluated under confound filtering with per-patch aggregation. Bold rows are significant at $\alpha=0.05$.}
\label{tab:top3_win_test_trackD_perpatch}
\begin{tabular}{l l c c}
\toprule
Model & Family & Wins & $p$ \\
\multicolumn{4}{l}{\footnotesize $N=18$ datasets; $k=12$ models; $\bar{p}_0 = 3/12 = 0.250$. Bold = significant at $\alpha=0.05$.} \\
\midrule
\textbf{ST-EEG-S}    & \textbf{EEG-FM} & \textbf{15/18} & \textbf{0.000} \\
REVE             & EEG-FM          & 8/18      & 0.057         \\
EEGPT            & EEG-FM          & 7/18      & 0.139         \\
EEGNet           & Supervised      & 6/18      & 0.283         \\
Moment-S         & TS-FM           & 5/18      & 0.481         \\
Moirai-S         & TS-FM           & 4/18      & 0.694         \\
Moment-B         & TS-FM           & 4/18      & 0.694         \\
BIOT             & EEG-FM          & 3/18      & 0.865         \\
NeuroLM          & EEG-FM          & 0/18      & 1.000         \\
NeuroRVQ         & EEG-FM          & 0/18      & 1.000         \\
CBraMod          & EEG-FM          & 0/18      & 1.000         \\
LaBraM           & EEG-FM          & 0/18      & 1.000         \\
\bottomrule
\end{tabular}
\end{table}

\begin{table}[t]
\centering
\small
\caption{\textbf{Binomial top-3 win test across 17 BCI datasets for mean pooling.} For each model, the exact one-sided binomial test asks whether the number of datasets on which the model ranks in the top~3 (by normalized balanced accuracy) exceeds the chance rate $\bar{p}_0 = 3/k$. The pool contains $k=12$ models (8 EEG-FMs + 3 TS-FMs + EEGNet) evaluated under confound filtering with mean pool aggregation. Bold rows are significant at $\alpha=0.05$.}
\label{tab:top3_win_test_trackD_meanpool}
\begin{tabular}{l l c c}
\toprule
Model & Family & Wins & $p$ \\
\multicolumn{4}{l}{\footnotesize $N=18$ datasets; $k=12$ models; $\bar{p}_0 = 3/12 = 0.250$. Bold = significant at $\alpha=0.05$.} \\
\midrule
\textbf{EEGPT}    & \textbf{EEG-FM} & \textbf{13/18} & \textbf{0.000} \\
\textbf{NeuroLM}    & \textbf{EEG-FM} & \textbf{10/18} & \textbf{0.005} \\
EEGNet           & Supervised      & 8/18      & 0.057         \\
REVE             & EEG-FM          & 7/18      & 0.139         \\
Moment-S         & TS-FM           & 4/18      & 0.694         \\
ST-EEG-S         & EEG-FM          & 4/18      & 0.694         \\
LaBraM           & EEG-FM          & 2/18      & 0.961         \\
NeuroRVQ         & EEG-FM          & 2/18      & 0.961         \\
Moment-B         & TS-FM           & 2/18      & 0.961         \\
BIOT             & EEG-FM          & 1/18      & 0.994         \\
CBraMod          & EEG-FM          & 1/18      & 0.994         \\
Moirai-S         & TS-FM           & 0/18      & 1.000         \\
\bottomrule
\end{tabular}
\end{table}

\paragraph{Statistical analysis.}
To identify which foundation models reliably rank among the best across BCI tasks, we apply a one-sided exact binomial test to each model's top-3 finish count.
For each of the $N=17$ datasets, we rank the $k=7$ models by normalized balanced accuracy and record whether a model places in the top~3.
Under the null hypothesis, each model has a $p_0 = 3/7 \approx 0.43$ probability of finishing in the top~3 on any given dataset by chance.
We then test whether the observed number of top-3 finishes significantly exceeds this rate using \texttt{scipy.stats.binomtest} (one-sided, $\alpha=0.05$).
Two models pass the significance threshold: \textbf{ST-EEG-S} (14/18 datasets, $p=0.003$) and \textbf{REVE} (12/18, $p=0.036$), indicating that both consistently outperform the other foundation models across paradigms.
EEGPT narrowly misses significance (11/18, $p=0.093$), while the remaining models, ST-EEG-S, LaBraM, CBraMod, and Moment-B do not finish in the top~3 more often than expected by chance.

\end{document}